\documentclass[12pt]{article}
\usepackage[latin1]{inputenc}
\usepackage{amsmath}
\usepackage{amsfonts}
\usepackage{amssymb}
\usepackage{graphicx}
\usepackage{setspace}
\usepackage[left=1.00in, right=1.00in, top=1.00in, bottom=1.00in]{geometry}

\usepackage{chngcntr}
\counterwithin{figure}{section}
\counterwithin{table}{section}
\counterwithin{equation}{section}

\usepackage{multirow}
\usepackage{array}
\usepackage{graphicx}

\newcommand\MyBox[2]{
	\fbox{\lower0.75cm
		\vbox to 1.2cm{\vfill
			\hbox to 1.2cm{\hfil\parbox{1.2cm}{#1\\#2}\hfil}
			\vfil}%
	}%
}

\usepackage[round]{natbib}
\usepackage{float}
\usepackage{caption}
\usepackage{subcaption}
\usepackage{wrapfig}
\usepackage{sidecap}

\begin{document}

	\title{Modern Dimension Reduction}

	\author{Philip D. Waggoner \\ University of Chicago \\ \texttt{pdwaggoner@uchicago.edu}}
	\date{ }
	
\maketitle

\thispagestyle{empty}

\begin{center}
	\textit{Forthcoming, Cambridge University Press}\footnote[0]{\textbf{Please let me know if you plan to cite this book in published work.}}
\end{center}

	\vspace{.3 in}
	
\begin{abstract}
	\noindent Data are not only ubiquitous in society, but are increasingly complex both in size and dimensionality. Dimension reduction offers researchers and scholars the ability to make such complex, high dimensional data spaces simpler and more manageable. This Element offers readers a suite of modern unsupervised dimension reduction techniques along with hundreds of lines of R code, to efficiently represent the original high dimensional data space in a simplified, lower dimensional subspace. Launching from the earliest dimension reduction technique principal components analysis and using real social science data, I introduce and walk readers through application of the following techniques: locally linear embedding, t-distributed stochastic neighbor embedding (t-SNE), uniform manifold approximation and projection, self-organizing maps, and deep autoencoders. The result is a well-stocked toolbox of unsupervised algorithms for tackling the complexities of high dimensional data so common in modern society. All code is publicly accessible on Github.
\end{abstract}

\clearpage

\tableofcontents

\clearpage

\listoffigures

\clearpage

\listoftables

\clearpage

\section*{Acknowledgements}

I am indebted to the my colleagues and students in the computational social science program at the University of Chicago. The UChicago environment has both stimulated and challenged me at a number of turns, which has undoubtedly impacted this manuscript for good, whether in conversations with colleagues, writing code for class, picking up new texts that inspire me, or simply sitting and thinking in my office. \\

Further, the series editors Mike Alvarez and Neal Beck have been immensely encouraging and supportive throughout the entire process. As with my previous Element, a special thanks is owed to Neal Beck, who has been a constant source of challenge and support. His friendship and insight have not only made this a notably stronger manuscript, but have also provided me immense personal benefit. \\

I am forever thankful for my precious three daughters and to my wife, Becky, who continues to deepen her love and support for me as my best friend and closest companion. 

\clearpage

\section*{Replication \& Code}

All code (DOI: 10.5281/zenodo.4594352) replicating all numerical and visual results are available at \textbf{**Link to Cambridge Repository xxxxx.zip file**}, and can be openly accessed online via Github, \texttt{https://github.com/pdwaggoner/dimension-reduction-CUP}.

\clearpage

\pagenumbering{arabic}

\section{Introduction}

The modern era of research is more concerned with collecting and learning from data than ever before. Whether for influencing decision-making, deepening an understanding of ``process'' broadly defined, or some other task, excitement and preoccupation with data are based in the rapid and massive \textit{production} of data. Correspondingly, we are witnessing rapid development of new methods for efficiently learning from these new data.

Accomplishing this central task of learning from data typically takes shape in a \textit{supervised} way, as forecasting and predictions are of high importance. Yet, supervised machine learning, which is concerned with predicting some labeled value based on learned patterns from a tuned model, in such a rapidly changing landscape can be tricky. For example, some labels may not exist (e.g., measurement error), the target outcome may be unclear, a priori expectations on outcome patterns may be nonexistent, or the input features may not be thoroughly or properly motivated. In such cases, the nature of ``supervision'' becomes unclear.

Alternatively, \textit{unsupervised} machine learning offers a different approach to the task of learning from data. Unsupervised learning allows data to essentially speak for itself, where no ground truth or expected outcomes condition the modeling process, nor does predicting some labeled value. Rather, unsupervised learning is primarily concerned with uncovering latent, non-random structure in data. By uncovering this structure, a deeper understanding of the data, and potentially how it was produced, are possible. Guided by our main task of learning from data, an unsupervised framework eases the theoretical burden of researchers relying on theoretical innovation or some assumed underlying data generating process (which could be incorrect), to instead defer to the data. In a word, at present we are interested in \textit{exploring} data. 

Importantly, while data are allowed to speak more freely in an unsupervised setting than in a constrained supervised setting, unsupervised methods are not without assumptions as well as choices to be made. Assumptions of different approaches to reducing dimensionality can result in different views of the data, as demonstrated and discussed across all techniques covered in this Element. Similarly, the choices researchers make throughout the process can also impact the patterns and results that emerge from different versions of different algorithms. As such, great care should be taken along with justification of choices made along the way to place results into a proper exploratory framework. To this end, a theme throughout the Element is encouraging researchers to try out and compare across several versions of algorithms, as well as set up tuning grids to search across combinations of hyperparameters where it makes sense. And at a higher level, it is useful to remember that unsupervised exploration of data as presented and discussed in this Element is rooted in an effort to parse signal from noise, which is extremely common in high dimensional data.

There are two main approaches to unsupervised machine learning: clustering and dimension reduction. In my previous Element, I covered this first realm of clustering \citep{waggoner2020unsupervised}. Clustering searches some data space for natural groupings and patterns, and then seeks to partition the data space in a way that reflects the underlying structural similarity. Dimension reduction, on the other hand, though still concerned with recovering structure in data, is instead interested in doing so by creating a simpler version of the more complex original version of the full data space. As a result, dimension reduction represents the structure of the original data in a clearer, more digestible, and usually simpler way. By combining these two Elements on clustering and dimension reduction, researchers and practitioners are afforded a firm understanding of a modern approach to unsupervised machine learning for addressing a host of social science problems and questions.

Though straightforward to define at a high level, unsupervised machine learning includes \textit{many} techniques and algorithms aimed at learning from data as there are many ways to conceptualize structure. Though some methods are relatively simple to build and understand, many unsupervised methods can quickly become complex, both in algorithm construction as well as in implementation and interpretation.

The goal of this Element, then, is to offer a framework for understanding and applying dimension reduction in a modern context. In service of this goal, I detail many algorithms, all of which differ in how they treat and process data, and thus how they conceptualize structure. Motivated by a common goal (learn from data) and situated in a common framework (unsupervised machine learning), the diverse suite of methods covered in this Element offers a representative picture of the rich diversity of the unsupervised dimension reduction landscape.

While the task is clear enough, it is important to remember that data are rarely, if ever, simple and tidy. Though an unsupervised approach to our central task of learning from data, if properly executed, allows for natural structure to emerge, the complexities of data make this intuitively-simple task often complex in application. That is, as data complexity deepens, so too does the process of method selection, implementation, and interpretation of the patterns and structure we uncover from the data. The reality of increased data complexity complicating the modeling process is especially true in a social science setting, where subjects are often people or institutions (\textit{occupied} by people), who are inherently complex and messy. Thus, building efficient unsupervised learners and then meaningfully interpreting patterns in a social science context are particularly challenging tasks in modern applications. Yet, a deeper, unified understanding of dimension reduction can lead to well-motivated, intentional, and justified decisions made throughout the modeling process. As a result, social scientists can overcome the hurdles of data complexity in pursuit of the central goal of learning from data in an unsupervised way.

\subsection{Defining the Title}

To put meat on the bones of the purpose of this Element and thus the value of dimension reduction in data analysis, I spend a few paragraphs unpacking the Element's title, \textit{Modern Dimension Reduction}.

First, \textit{what are dimensions?} Dimensions are variables or features of data. Mathematically, these are column vectors in some data matrix that, when increased, also increase the complexities among a set of features existing in a common data space.\footnote{Importantly, data complexity is often defined by both \textit{size} and \textit{dimensionality}. Where size refers to the volume of data in a single space and often dubbed ``big data,'' this Element focuses on the dimensionality part of complex data in light of the scope of methods covered. Yet it is important to highlight many of these techniques, especially UMAP and autoencoders, easily adapt to big data settings.} So, a \textit{high} dimensional data set consists of many features with measured values across some set of observations (row vectors). This, or any data space can be summarized as $\mathbf{X}$, which signifies an $N \times P$ data matrix with $n \in \{1,\dots, N\}$ observations (rows) and $p \in \{1,\dots, P\}$ features (columns). Of note, different fields use different terms for $p$, e.g., ``variables'', ``predictors''/``regressors'' (in a supervised setting), ``inputs,'' ``features,'' and so on. Throughout this Element, I usually refer to $p$ as \textit{features}, because this is a more descriptive name for these components of the data space, as the measured values record specific \textit{features} of the observations, $n$, that exist in $\mathbf{X}$.

Of note, many dimension reduction algorithms can take mixed data types (e.g., continuous and categorical). But these algorithms tend to perform best with continuous features rather than categorical features that have discrete levels or classes, as categorical features lack variance or nuance that is useful to understand latent structure. That is, if a feature has values of 0 or 1, then these are orthogonal, discrete differences. The lack of nuance in the scale limits the value of dimension reduction as we will see. There may be room to alter the feature construction (e.g., one-hot encoding). Yet, the value of dimension reduction as covered in this Element is most beneficial when using continuous, numeric features. We will come back to this point and also the role of scaling input features in the following sections.

Now, \textit{what is dimension reduction?} Dimension reduction is primarily concerned with taking a high dimensional data space, which typically means $p > 4$, and making a simpler version of it, which typically means $p = 2$. And a visual version of $p = 2$ might be a scatterplot, with $X$ and $Y$ axes capturing the reduced first and second dimensions. We address visualization of dimension reduction results throughout the Element. But the main idea with dimension reduction is to embed or represent the high dimensional \textit{full data space} on a lower dimensional \textit{subspace}. Why ``represent data'' in the first place? Taken as a whole, the high dimensional, original data space is \textit{uninterpretable} by humans such that we cannot readily understand a visual rendering of data in more than four dimensions.

The process of moving from a high dimensional space to a low dimensional subspace substantively results in making data being more interpretable, understandable, and digestible. This value of dimension reduction is why it is sometimes referred to as ``low dimensional embedding,'' ``mapping,'' ``lower dimensional projection,'' or ``data representation.'' The task of dimension, then, is to learn prominent patterns in the higher dimensional setting, and then project these patterns onto a lower dimensional subspace. The lower dimensional space acts as a summary of the full space based on the patterns naturally existing across all features, $\mathbf{X}$. Importantly, there are many aspects of this definition of dimension reduction that likely make social scientists uncomfortable. Namely, by summarizing anything, whether data, a film, or a baseball game, some information is necessarily discarded or left out. So too in dimension reduction, which often and \textit{intentionally} throws away some data for the benefit of a simpler look at the full, complex original data space. Using the learned information to move from the higher to the lower dimensional setting requires a choice of discarding some of the information deemed non-substantive, which is up to each researcher. For example, principal components analysis (PCA) is one of the oldest dimension reduction techniques, and it searches for a lower dimensional version of the data space that maximizes the total variance. The initial principal component is calculated for the direction along which the data vary most. Subsequent components are orthogonal to the preceding components, such that unique variance is captured in subsequent component calculations. The ultimate choice in most PCA applications, then, is to decide on the number of components that do a good (enough) job of characterizing the data. The number of selected components should be less than $p$, as technically up to $p$ components can be found in any data space, $\mathbf{X}$. Though unpacked later in the Element, the point at present is that the choice of selecting a subset of components from among all calculated components requires the researcher to select some of the data and discard other parts of the data. This choice is equivalent to saying, ``I am OK moving forward explaining most of the data, but not all of it, for the benefit of simpler, yet still informative data.''

Finally, though the task of projecting a high dimensional data space onto a lower dimensional subspace is a common one, there are \textit{many} ways to conceptualize patterns in data, and thus many methods for learning these patterns. For example, we might be interested in reducing dimensionality based on maximizing similarity across features. Though similarity can be conceptualized and measured in many ways, e.g., correlation, covariance, or spatial distance, such a goal would put us in the world of PCA, as previously discussed. Alternatively, if we suspect an aggregation of many small neighborhoods of data produce a simplified version of the higher dimensional space, then we might be in the world of uniform manifold approximation and projection (UMAP). As discussed later, the goal of UMAP is to first learn the shape and contours of the manifold (a geometric shape defining a set of data) underlying the high dimensional data space. Once learned, UMAP translates the learned manifold to a lower dimensional version of it, based on the learned distances between observations distributed across the manifold. UMAP has the added benefit of giving a \textit{reproducible} solution that efficiently balances global and local structure defined by the learned manifold. The reproducibility aspect of UMAP is an immensely powerful extension of another dimension reduction technique, t-distributed stochastic neighbor embedding (t-SNE), which is also covered later in the Element. As such, our conceptualization of similarity, and thus how we treat data during modeling, will place us in vastly different realms for reducing dimensionality. Despite the many flavors and differences across techniques, though, the task remains constant: to learn the structure underlying the high dimensional data, and then produce a lower dimensional, simpler version of the full, complex data.

As (big) data becomes increasingly complex with demand for sophisticated computational skills also growing, dimension reduction is a critical skill all quantitative researchers should know and practice. As one of the core approaches to unsupervised machine learning, dimension reduction is extremely helpful for making these widely occurring, complex data spaces more interpretable and manageable. Whether using dimension reduction as a part of a broader research program through feature extraction, or on its own to learn natural patterns in data allowing latent structure to emerge in an intuitive light, its value is no less diminished.

\subsection{Running Example: 2019 American National Election Pilot Study}

As a running example throughout the Element, I use the 2019 American National Election Pilot Study data \citep{american2019anes}.\footnote{The American National Election Studies (www.electionstudies.org). These materials are based on work supported by the National Science Foundation under grant numbers SES 1444721, 2014-2017, the University of Michigan, and Stanford University.} The ANES is a large, opt-in survey including complete responses from over 3,000 respondents.

Though many rich features are included in the data (e.g., political preferences, demographics, etc.), I focus on the battery of feeling thermometers, which measure respondents' preferences on a host of topics. Respondents are asked about how they feel toward some person or concept, and then asked to record that feeling on a scale from 0 to 100, with 0 being extremely cold toward the concept and 100 being extremely warm toward the concept. Though the question wording has evolved over several iterations of the ANES, the 2019 pilot study used in this Element includes consistent question wording for all feeling thermometers, \textit{How would you rate [\texttt{topic}]?}

There are 35 feeling thermometers in the 2019 ANES Pilot survey ranging from political candidates (e.g., Sanders, Trump, Biden, Harris, etc.) to social issues and people groups (e.g., transgender, Asians, Muslims, journalists, etc.) and even institutions (e.g., immigration and customs, NATO, the UN, etc.) and countries (e.g., Mexico, France, Israel, etc.). The value of these thermometers for present purposes is many are likely collinear with each other pointing to common variation across features, and equally contribute to the complexity of the ANES data space. The assumption here, then, is these feeling thermometers should project onto some lower dimensional, subspace on the basis of similarity (however defined). The simpler subspace, then, can be used to understand natural patterns in the American electorate. Of note, I thoroughly explore correlation across features at the outset of Section 3, prior to the treatment of PCA.

In an effort to deepen an understanding of the recovered patterns, I color data points (respondents) in most visualizations based on stated party affiliations. Taken with the solutions comprising only a battery of \textit{apolitical} feeling thermometers and no feature for party affiliation, my assumption, which also serves as a naive expectation throughout, is that the structure underlying responses to these thermometers should take shape in a partisan way. On average, I expect Democrats to be grouped together and distinct from Non-Democrats who should also be grouped together. It is worth reiterating that \textit{no} party affiliation feature is included in the fit of any algorithms. Party affiliation is only used to contextualize findings and add clarity to the recovered latent structure.

Social scientists do not typically work with extremely high dimensional data in the traditional sense with hundreds or even thousands of dimensions. Yet, recall that substantively interpreting visual patterns in any dimension greater than four (or really greater than three in most cases) is a nearly impossible task. Though not a traditionally ``big data'' application, the inclusion of 35-dimensional data in this Element still allows for demonstration of dimension reduction's value in social science applications.

For reference, question wording along with the ANES labels for each feeling thermometer are listed in Table \ref{table:fts}. The shorthand labels in the second column will be used throughout the Element.

\begin{table}[!h] \centering 

	\caption{Question Wording for Feeling Thermometers} 

	\label{table:fts} 

	\begin{tabular}{l|c}

		\hline

		\textit{Question} & \textit{Shorthand Label} \\

		\hline

		\textit{How would you rate Donald Trump?} & Trump \\

		\textit{How would you rate Barack Obama?} & Obama \\

		\textit{How would you rate Joe Biden?} & Biden \\

		\textit{How would you rate Elizabeth Warren?} & Warren \\

		\textit{How would you rate Bernie Sanders?} & Sanders \\

		\textit{How would you rate Pete Buttigieg?} & Buttigieg \\

		\textit{How would you rate Kamala Harris?} & Harris \\

		\textit{How would you rate blacks?} & Black \\

		\textit{How would you rate whites?} & White \\

		\textit{How would you rate Hispanics?} & Hispanic \\

		\textit{How would you rate Asians?} & Asian \\

		\textit{How would you rate Muslims?} & Muslim \\

		\textit{How would you rate illegal immigrants?} & Illegal \\

		\textit{How would you rate immigrants?} & Immigrants \\

		\textit{How would you rate legal immigrants?} & Legal Immigrants \\

		\textit{How would you rate journalists?} & Journalists \\

		\textit{How would you rate NATO?} & NATO \\

		\textit{How would you rate United Nations (UN)?} & UN \\

		\textit{How would you rate Immigration and Customs Enforcement (ICE)?} & ICE \\

		\textit{How would you rate National Rifle Association (NRA)?} & NRA \\

		\textit{How would you rate China?} & China \\

		\textit{How would you rate North Korea?} & North Korea \\

		\textit{How would you rate Mexico?} & Mexico \\

		\textit{How would you rate Saudi Arabia?} & Saudi Arabia \\

		\textit{How would you rate Ukraine?} & Ukraine \\

		\textit{How would you rate Iran?} & Iran \\

		\textit{How would you rate Great Britain?} & Britain \\

		\textit{How would you rate Germany?} & Germany \\

		\textit{How would you rate Japan?} & Japan \\

		\textit{How would you rate Israel?} & Israel \\

		\textit{How would you rate France?} & France \\

		\textit{How would you rate Canada?} & Canada \\

		\textit{How would you rate Turkey?} & Turkey \\

		\textit{How would you rate Russia?} & Russia \\

		\textit{How would you rate Palestine?} & Palestine \\

		\hline

	\end{tabular}

\end{table}

\clearpage

\subsection{The Methods}

This Element is focused on introducing readers to the intuition of dimension reduction and its value in applied data analysis and computational modeling contexts. To this end, along with introduction of hundreds of lines of R code to guide application and engagement with practical dimension reduction, I cover six methods.

Principal components analysis (PCA) is a fitting method with which to begin any treatment of dimension reduction, whether formal or applied. PCA is one of the earliest approaches to explicitly reducing dimensionality and complexity of a data space, as opposed to informally reducing dimensionality through two-dimensional visualization, which has been around for hundreds of years in various forms. PCA, as tacitly mentioned above, is concerned with finding a new, lower dimensional subspace based on maximizing the natural variance that exists across the full set of inputs, $\mathbf{X}$. PCA computes a new set of components based on a linear combination of weighted features values, which are often considered as ``prototypes'' that capture unique variance in higher dimensions. Observations load onto each component, which are orthogonal to all other components. The loading of observations onto components results in a new set of measured values for observations across a new set of features. These so-called component scores can be extracted and used as new features in subsequent analyses or can be used for analytical purposes in their own rite.

Next, building on the linear construction of PCA, locally linear embedding (LLE) is a newer technique based on a similar intuition. LLE linearly combines weighted feature values to learn a latent structure as in PCA, but it does so in a neighbor-based way that is more overtly rooted in manifold learning. LLE searches small neighborhoods in the full data space assuming data are distributed across a latent manifold, and then calculates weights for observations existing around the candidate observation, $\forall i \in N$. Higher weights are given to closer observations, and lower weights are given to distant observations. Then, at this point, the construction is similar to PCA, where distance-based weights are multiplied by raw feature values and are then summed to give a lower dimensional picture of the underlying structure. LLE learns structure locally and uses the local structure to reproduce the global structure, whereas PCA is interested in producing a summary of global structure based on maximal variance.

After PCA and LLE, we abandon linear combinations of weighted feature values, but build on the manifold aspect introduced with LLE. The focus next is on explicitly nonlinear approaches to searching some data space for an efficient, lower dimensional version of the full space. Specifically, we cover t-distributed stochastic neighbor embedding (t-SNE) and uniform manifold approximation and projection (UMAP). These approaches are neighbor-based like LLE, but they differ in how they approximate the lower dimensional version of the data. Namely, they are both interested in minimizing some cost function (Kullback-Leibler divergence in t-SNE and cross-entropy in UMAP; discussed much more later) to reduce dimensionality but based on balancing both local and global structure learned from the high dimensional space. Then, they attempt to replicate the learned same structure in a lower dimensional setting. Though feature extraction is possible with these techniques, they are much more widely used for visualization in typically two dimensions. I cover these techniques focusing also on their visual value for learning and representing the dimension reduction solution.

In the final substantive section, we once again shift gears, and focus on neural network-based approaches to dimension reduction. These covered techniques are unsupervised in that we have no expected outcome nor are we working with labeled data. Instead, we are interested in emulating the way the human brain learns by firing neurons to give some output based on raw inputs. This process, as well as it's connection to computational modeling are discussed at length in the respective section. The methods are, first, self-organizing maps (SOM), and then autoencoders. SOM have been around for several years, and autoencoders are, by comparison, more recent innovations on older techniques, specifically the restricted Boltzmann machine. SOM are feedforward neural networks with no hidden layer, and autoencoders are feedforward neural networks, but with an output layer the size of the input layer. These terms will be fully unpacked later, but for now, the core idea with autoencoders is information loss is forced through a process called \textit{encoding}. The encoded version of the high dimensional data acts as the ``hidden layer.'' Then, the task of the autoencoder is to \textit{decode} this layer in an attempt to reproduce the original high dimensional input space, but only on the basis of the simplified encoded layer. The difference between the output and input layers is called reconstruction error, and this captures the quality of the autoencoder. It is important to note the autoencoders can be either shallow or deep, where the latter version puts us into an explicitly \textit{deep learning} realm. Thus, we will cover what it means to move from shallow to deep, and associated benefits and drawbacks to both.

Also, where it makes sense, important associated innovations or related concepts will be addressed in an effort to situate readers more firmly in this world of modern, unsupervised dimension reduction. For example, when covering LLE, t-SNE, and UMAP, I will place them in the broader subfield of manifold learning, which is a branch of machine learning dedicated to deriving some latent manifold, assuming one exists. Or, for example, it is impossible to understand the final section on SOM and autoencoders without introducing a basic neural network architecture, which we will do for the task of dimension reduction.

Of note, there are \textit{many} other approaches to dimension reduction that are not covered in this Element, such as multidimensional scaling, ISOMAP, and factor analysis. Though the main reason for exclusion of other techniques is a limited amount of space, some of these methods can also suffer from pernicious problems such as computational inefficiency (ISOMAP compared to UMAP) or a poor rendering of the high dimensional space (factor analysis compared to deep autoencoders). In short, though biased by my preferences at some level, I have made every effort to justify selection of these methods on the basis of computational efficiency, creativity in dealing with high dimensional data complexity, generalizability of the solution, as well as linkages across methods, all of which can be linked back to principal components analysis (PCA), as we will see at a later point. 

\subsection{The Methods \textit{Not} Covered}

Though we cover a lot of ground in this Element, it is impossible to cover \textit{all} approaches to dimension reduction. Of note, I do not address multidimensional scaling (MDS), other approaches to space scaling (e.g., optimal classification, NOMINATE, etc.), factor analysis, item response theory (IRT) models, non-negative matrix factorization (NMF), or Bayesian flavors of these and many other methods. The reason for this is largely due to insufficient space, but it is also due to substantive goals. Paired with my recent Element on clustering \citep{waggoner2020unsupervised}, I am interested in widening the social scientist's \textit{unsupervised machine learning} toolbox. And further, many of the methods not covered in this Element are either widely used and taught in quantitative social science applications or are covered eleswhere in other excellent texts, such as \citet{armstrong2014analyzing}.

Also, related to presenting an unsupervised machine learning approach to dimension reduction, many of the methods not covered in this Element's approach dimension reduction from a very different perspective, which is often more assumption-laden. For example, factor analysis starts with assuming latent factors are Gaussian to allow for asserting conditional independence across the factor structure. The goal then is to recover this \textit{latent factor structure that results in the observed input space}. That is, factor analysis assumes there is unobserved structure that causes the observed structure comprised of input features, e.g.,

\begin{equation}
X = b_{1}\text{Component}_1.
\end{equation}

In this way, factor analysis is arguably more interested in measurement rather than explicit dimension reduction, though the line between measurement and dimension reduction can quickly become gray. 

Yet, principal components analysis (PCA) on the other hand stops short of making any assumptions, distributional or otherwise, related to the ``latent factor structure.'' Instead, PCA is interested in learning a reduced version of the data space comprised of components that are combinations of weighted features, e.g.,

\begin{equation}
\text{Component} = b_1X_1.
\label{eq:pcaeq}
\end{equation}

The goal of PCA, as described in greater depth later in the Element, is to find the optimal weights (e.g., $b_1$ in equation \ref{eq:pcaeq}), which are defined by maximizing variance on the basis of what we get to observe, which again is the set of unlabeled input features. Thus, while factor analysis and PCA are both interested in learning latent structure at some level, these methods consider and search for this structure in fundamentally different ways, with the former being a more assumption-heavy approach. Though omission of certain methods in this Element is largely a practical concern, it remains a second-order goal to push the reader to think about dimension reduction from a perspective that may be both new and (hopefully) challenging.

\subsection{Tidy Programming in R}

In this Element, wherever possible, I will leverage the tidy approach to programming in R \citep{wickham2019welcome}, which is gaining ground in the social sciences \citep{kennedyR}. Essentially, the reason for this and at the core of the tidyverse, is the notion of consistent syntax and stacked functions. Streamlining key functions across multiple packages with a common syntax gives way to writing efficient and reproducible code. Of note, the ``stacking'' of functions previous mentioned occurs via a key operator: the pipe, \texttt{\%$>$\%}, which is read ``then...''. The pipe allows for multiple functions to be passed to each other, which allows for writing large chunks of code to be run simultaneously, as demonstrated in the following section.

Though readers need not be expert programmers, some familiarity with R, or other open-source object-oriented languages (e.g., Python or Julia) will significantly ease reading and implementation of techniques covered throughout. Where possible, I do my best to annotate and explain code construction and choices. But I leave it to readers to take what they will from the code written for this Element.

Readers are also encouraged to load the 2019 ANES Pilot Study data with the relevant code in the Github repository, which as written requires installation of the \texttt{tidyverse} and \texttt{here} packages. For the latter package to work, readers should ensure that the script for running the code in this Element is in a subfolder called, \texttt{Data}. Otherwise, the function will not work. I highly recommend inspecting the help documentation for the \texttt{here} package to learn more about it's immense value in large scale projects of this sort.

\subsection{Cleaning the Data}

Rather than jump into a technical, though applied work of this sort with clean data, part of my goal with this Element is to demonstrate an effective workflow for analysis from start to finish. I strongly encourage readers to think very carefully about collection and cleaning of data, as much as they would about modeling and inferences on the back end. Indeed, data, especially in the social sciences, are rarely clean and complete. Even if complete, data are rarely in the shape required for specific research projects. In a word, getting data in good shape is called \textit{preprocessing}. The first major step in preprocessing is to clean and organize the data in line with specific project goals. The second major step, covered in the following subsection, is addressing missing data. In some circles the second step is called \textit{feature engineering}. The code used to preprocess the ANES data used throughout, along with additional code for techniques not covered in this Element, are included in the complementary repositories on Github.

There are two main approaches to cleaning data in R: base R operations or the tidy approach. The base R approach is typically considered manual, e.g., selecting features using indexing (e.g., using $\texttt{data\_set}[ \hspace{.5em}, -3]$ to omit the third feature from the object \texttt{data\_set}). A tidy approach to data management on the other hand, uses consistent syntax with ``human-readable'' code to result in a clean, single chunk of code comprised of many piped functions, which streamlines the research process.

For present purposes, we need to first restrict the space to contain the features we care about. There are 900 features in the full ANES data set. So, to make this more manageable and avoid the arduous task of manually selecting features for each model, the first step is to \texttt{select()} the relevant features for modeling, which includes the 35 feeling thermometers and an indicator for party affiliation. This subset of the full feature space is stored in a new dataset called \texttt{anes\_raw}. This is a convenience that will ease plotting both raw features as well as model objects later in the Element.

With the subset in hand, we need to recode missing values to be a value of \texttt{NA}. This is a critical step to clarify missing cases in the data, which will be built upon in the following subsection on feature engineering.

Finally, we take a \texttt{glimpse()} of our cleaned data subset to ensure everything looks as it should. The \texttt{glimpse()} function from the tidyverse displays feature names and the first few values, along with displaying the dimensions of the data matrix, $\mathbf{X}$. We have 3,165 observations along with the 35 feeling thermometers and a feature for party affiliation, which will be used for coloring the plots throughout the Element. Summary statistics for the feeling thermometers are presented in Table \ref{table:fts_ss}.

\begin{table}[!h] \centering 

	\caption{Feeling Thermometer Summary Statistics: Raw Features} 

	\label{table:fts_ss} 

	\begin{tabular}{l|r|r|r|r|r|r|r|r}

		\hline

		\textit{Feature} & \textit{Complete} & \textit{Mean} & \textit{SD} & \textit{Min} & \textit{Q2} & \textit{Median} & \textit{Q3} & \textit{Max} \\

		\hline

		Trump & 98.9\% & 43.87 & 41.43 & 0 & 2 & 38 & 91 & 100 \\

		Obama & 99.4\% & 53.53 & 37.54 & 0 & 11 & 59 & 91 & 100 \\

		Biden & 98.4\% & 42.15 & 33.44 & 0 & 7 & 42 & 70 & 100 \\

		Warren & 95.8\% & 40.51 & 34.36 & 0 & 4 & 41 & 71 & 100 \\

		Sanders & 98.6\% & 42.15 & 34.88 & 0 & 5 & 42 & 73 & 100 \\

		Buttigieg & 90.1\% & 38.66 & 30.40 & 0 & 7 & 41 & 61 & 100 \\

		Harris & 93.3\% & 35.30 & 30.18 & 0 & 4 & 35 & 58 & 100 \\

		Black & 99.8\% & 70.86 & 23.78 & 0 & 51 & 73 & 91 & 100 \\

		White & 99.8\% & 70.91 & 22.82 & 0 & 51 & 73 & 90 & 100 \\

		Hispanic & 99.9\% & 69.00 & 24.20 & 0 & 50 & 71 & 90 & 100 \\

		Asian & 99.8\% & 69.30 & 23.54 & 0 & 51 & 71 & 90 & 100 \\

		Muslim & 99.9\% & 51.90 & 29.67 & 0 & 31 & 51 & 75 & 100 \\

		Illegal & 99.9\% & 42.13 & 32.01 & 0 & 9 & 45 & 68 & 100 \\

		Immigrants & 50.1\% & 68.33 & 26.71 & 0 & 51 & 72 & 91 & 100 \\

		Legal Immigrants & 49.8\% & 73.07 & 24.32 & 0 & 55 & 79 & 93 & 100 \\

		Journalists & 100\% & 49.87 & 31.57 & 0 & 21 & 50 & 77 & 100 \\

		NATO & 95.3\% & 54.95 & 27.40 & 0 & 41 & 53 & 75 & 100 \\

		UN & 97.9\% & 50.59 & 30.59 & 0 & 25 & 51 & 75 & 100 \\

		ICE & 98.2\% & 52.22 & 33.32 & 0 & 23 & 51 & 84 & 100 \\

		NRA & 98.1\% & 45.87 & 36.54 & 0 & 6 & 49 & 82 & 100 \\

		China & 100\% & 37.93 & 24.92 & 0 & 17 & 40 & 52 & 100 \\

		North Korea & 99.9\% & 20.78 & 23.16 & 0 & 2 & 11 & 35 & 100 \\

		Mexico & 99.9\% & 53.52 & 26.50 & 0 & 38 & 52 & 72 & 100 \\

		Saudi Arabia & 99.9\% & 32.08 & 24.18 & 0 & 10 & 31 & 50 & 100 \\

		Ukraine & 99.8\% & 49.08 & 24.37 & 0 & 35 & 50 & 64 & 100 \\

		Iran & 99.9\% & 26.56 & 24.79 & 0 & 4 & 20 & 47 & 100 \\

		Britain & 100\% & 69.14 & 23.69 & 0 & 52 & 72 & 89 & 100 \\

		Germany & 100\% & 60.45 & 24.822 & 0 & 49 & 60 & 80 & 100 \\

		Japan & 100\% & 64.79 & 23.87 & 0 & 50 & 67 & 84 & 100 \\

		Israel & 99.9\% & 59.64 & 29.1 & 0 & 43 & 58 & 86 & 100 \\

		France & 99.9\% & 58.58 & 25.07 & 0 & 47 & 59 & 78 & 100 \\

		Canada & 100\% & 72.42 & 23.95 & 0 & 56 & 78 & 92 & 100 \\

		Turkey & 99.8\% & 37.69 & 23.81 & 0 & 18 & 41 & 51 & 100 \\

		Russia & 99.9\% & 31.43 & 24.98 & 0 & 8 & 30 & 50 & 100 \\

		Palestine & 99.7\% & 40.13 & 27.44 & 0 & 13 & 47 & 55 & 100 \\

		\hline

	\end{tabular}

\end{table}

\clearpage

\subsection{Addressing Missing Data}

Most of the algorithms covered in this Element require complete data to give a reliable lower dimensional version of the data. In fact, some algorithms have no mechanism to handle missing data, whereas others simply drop these cases if they are passed to the algorithm. Listwise, or ``brute force'' deletion can be a risky strategy and problematic for generalizations drawn from patterns. It is highly discouraged to simply drop (or delete) data, as data are extremely valuable, especially in a social science context where it is costly and time consuming to repeatedly return to a population and draw new samples. As such, I handle missing data with imputation, which means to replace the instance of \texttt{NA} with some calculated value that is assumed to be a plausible substitution for that which the value likely would have been, had it been recorded. Beyond losing all or most of our data (which alone is sufficient to bypass the deletion strategy), as previously mentioned a key reason to impute rather than delete is because many of the algorithms require complete data to run properly, or at least produce reliable results. To the latter point, some algorithms may work with incomplete data, opting to essentially ignore these cases rather than stop. But if the goal is to understand how some complex higher dimensional data space maps onto a lower dimensional version of the original complex space, then it does not make sense to reduce dimensionality of a partially populated data space.

Before imputing data, I recommend first getting to know the data better to understand the contours of missingness. Such a step will help diagnose the cause of the missingness, if any, as well as the precise location of missingness in the full data space. Though omitted from the text, I leverage a series of visualizations built with the \texttt{naniar} package and present the code on Github. Inspecting the patterns of missingness allow for more targeted strategies for dealing with missingness. Once visually explored, but before getting to the imputation strategy, it is important to consider the \textit{cause} of missingness in light of the \textit{patterns} of missingness. \citet{rubin1976inference} defined three major causes: missing not at random (MNAR; a systematic issue in recording or measuring data resulted in missingness), missing at random (MAR; missingness is random, but is not an equal probability for each observation), and missing completely at random (MCAR; missingness is equally random for all observations). Though often difficult to defend one of these underlying causes of missingness in the data, researchers should take care in at least thinking about this underlying driver before proceeding to imputation or any strategy aimed at dealing with the missingness.

Assuming MCAR, I proceed with imputation by taking advantage of the \texttt{recipes} package, which is has many powerful functions for data preprocessing of this sort. Specifically, I impute missing values using the k-Nearest Neighbors algorithm, or kNN. kNN is a simple supervised learning algorithm that, as leveraged in the current case, imputes based on a subset of observations surrounding a data point. First, kNN defines a small neighborhood of size $k$ around a candidate case with a missing value. Averages are then taken based on the values of those existing in the smaller neighborhood. The neighborhood average is the imputed value for the missing case. Put differently, kNN is used to first reduce the search space. Then, from among the reduced group of observations, the average of the non-missing cases in the neighborhood serve as the new, imputed value for the missing case. The procedure follows these steps for all missing cases across all features. This approach is considered \textit{multiple} imputation in that there is a statistical procedure resulting in a computed value, rather than the more common \textit{single} imputation approach that might be on the basis of taking the same value from the most recent complete case along each feature. kNN for imputation has the added benefit of deciding on which features to create the neighborhood. For example, should we define a neighborhood on the basis of people-based feeling thermometers? Or institution-based thermometers? Or all features? For this application, I base imputation on all features. The summary statistics for the full, imputed ANES data set based on our kNN recipe is presented in Table \ref{table:imputed_ss}. We can see from the table, first and foremost, that all completion rates are 100\%, suggesting we have successfully imputed the missing cases. 

\begin{table}[!h] \centering 

	\caption{Feeling Thermometer Summary Statistics: Imputed Features} 

	\label{table:imputed_ss} 

	\begin{tabular}{l|r|r|r|r|r|r|r|r}

		\hline

		\textit{Feature} & \textit{Complete} & \textit{Mean} & \textit{SD} & \textit{Min} & \textit{Q2} & \textit{Median} & \textit{Q3} & \textit{Max} \\

		\hline

		Trump & 100\% & 43.72 & 41.31 & 0 & 2.0 & 37.0 & 90 & 100 \\

		Obama & 100\% & 53.48 & 37.48 & 0 & 11.0 & 59.0 & 91 & 100 \\

		Biden & 100\% & 42.26 & 33.26 & 0 & 8.0 & 43.0 & 70 & 100 \\

		Warren & 100\% & 40.74 & 33.95 & 0 & 4.0 & 41.0 & 70 & 100 \\

		Sanders & 100\% & 42.26 & 34.74 & 0 & 5.0 & 43.0 & 73 & 100 \\

		Buttigieg & 100\% & 38.97 & 29.67 & 0 & 9.0 & 42.0 & 60 & 100 \\

		Harris & 100\% & 35.57 & 29.68 & 0 & 5.0 & 36.0 & 57 & 100 \\

		Black & 100\% & 70.86 & 23.77 & 0 & 51.0 & 73.0 & 91 & 100 \\

		White & 100\% & 70.91 & 22.80 & 0 & 51.0 & 73.0 & 90 & 100 \\

		Hispanic & 100\% & 69.00 & 24.18 & 0 & 50.0 & 71.0 & 90 & 100 \\

		Asian & 100\% & 69.30 & 23.52 & 0 & 51.0 & 71.0 & 90 & 100 \\

		Muslim & 100\% & 51.90 & 29.66 & 0 & 31.0 & 51.0 & 75 & 100 \\

		Illegal & 100\% & 42.14 & 32.00 & 0 & 9.0 & 45.0 & 68 & 100 \\

		Immigrants & 100\% & 67.52 & 23.69 & 0 & 51.0 & 70.0 & 87 & 100 \\

		Legal Immigrants & 100\% & 72.67 & 21.43 & 0 & 58.4 & 76.2 & 90 & 100 \\

		Journalists & 100\% & 49.87 & 31.57 & 0 & 21.0 & 50.0 & 77 & 100 \\

		NATO & 100\% & 54.85 & 27.09 & 0 & 41.0 & 53.0 & 75 & 100 \\

		UN & 100\% & 50.63 & 30.40 & 0 & 25.6 & 51.0 & 75 & 100 \\

		ICE & 100\% & 52.06 & 33.12 & 0 & 24.0 & 51.0 & 84 & 100 \\

		NRA & 100\% & 45.72 & 36.34 & 0 & 6.0 & 49.0 & 81 & 100 \\

		China & 100\% & 37.93 & 24.92 & 0 & 17.0 & 40.0 & 52 & 100 \\

		North Korea & 100\% & 20.78 & 23.15 & 0 & 2.0 & 11.0 & 35 & 100 \\

		Mexico & 100\% & 53.52 & 26.50 & 0 & 38.0 & 52.0 & 72 & 100 \\

		Saudi Arabia & 100\% & 32.09 & 24.17 & 0 & 10.0 & 31.0 & 50 & 100 \\

		Ukraine & 100\% & 49.07 & 24.35 & 0 & 35.0 & 50.0 & 64 & 100 \\

		Iran & 100\% & 26.55 & 24.78 & 0 & 4.0 & 20.0 & 47 & 100 \\

		Britain & 100\% & 69.14 & 23.69 & 0 & 52.0 & 72.0 & 89 & 100\\

		Germany & 100\% & 60.45 & 24.82 & 0 & 49.0 & 60.0 & 80 & 100 \\

		Japan & 100\% & 64.79 & 23.87 & 0 & 50.0 & 67.0 & 84 & 100 \\

		Israel & 100\% & 59.63 & 29.09 & 0 & 43.0 & 58.0 & 86 & 100 \\

		France & 100\% & 58.58 & 25.07 & 0 & 47.0 & 59.0 & 78 & 100 \\

		Canada & 100\% & 72.42 & 23.95 & 0 & 56.0 & 78.0 & 92 & 100\\

		Turkey & 100\% & 37.68 & 23.79 & 0 & 18.0 & 41.0 & 51 & 100 \\

		Russia & 100\% & 31.42 & 24.98 & 0 & 8.0 & 30.0 & 50 & 100 \\

		Palestine & 100\% & 40.13 & 27.42 & 0 & 13.0 & 47.0 & 55 & 100 \\

		Democrat & 100\% & 0.42 & 0.49 & 0 & 0.0 & 0.0 & 1 & 1\\

		\hline

	\end{tabular}

\end{table}

\clearpage

To inspect the quality of our solution and to avoid repeatedly comparing back and forth with the raw ANES summary statistics in Table \ref{table:fts_ss}, we can take a closer, individual look at some of the particularly problematic features. For these three features with the most missing values, we are looking for substantively similar summary statistics for the raw version of the feature compared to the imputed version. Take a look at the comparisons for the \texttt{Immigrants}, \texttt{Legal Immigrants}, and \texttt{Buttigieg} features presented in Table \ref{table:compare_ss}.

\begin{table}[!h] \centering 

	\caption{Comparing Summaries for the Three ``Most Missing'' Features} 

	\label{table:compare_ss} 

	\begin{tabular}{l|r|r|r|r|r|r|r|r}

		\hline

		\textit{Feature} & \textit{Complete} & \textit{Mean} & \textit{SD} & \textit{Min} & \textit{Q2} & \textit{Median} & \textit{Q3} & \textit{Max} \\

		\hline

		Immigrants (\textit{raw}) & 50.1\% & 68.33 & 26.71 & 0 & 51 & 72 & 91 & 100 \\

		Immigrants (\textit{imputed}) & 100\% & 67.52 & 23.69 & 0 & 51.0 & 70.0 & 87 & 100 \\

		\hline

		Legal Immigrants (\textit{raw}) & 49.8\% & 73.07 & 24.32 & 0 & 55 & 79 & 93 & 100 \\

		Legal Immigrants (\textit{imputed}) & 100\% & 72.67 & 21.43 & 0 & 58.4 & 76.2 & 90 & 100 \\

		\hline

		Buttigieg (\textit{raw}) & 90.1\% & 38.66 & 30.40 & 0 & 7 & 41 & 61 & 100 \\

		Buttigieg (\textit{imputed}) & 100\% & 38.97 & 29.67 & 0 & 9.0 & 42.0 & 60 & 100 \\

		\hline

	\end{tabular}

\end{table}

In general, when the missingness is greater as for the \texttt{Immigrants} and \texttt{Legal Immigrants} features, we will do a slightly worse job of matching the summaries for the raw (non-imputed) version as we simply have less data to learn from in the full data space. Yet, even still, with completion rates of 50.1\% and 49.8\% for \texttt{Immigrants} and \texttt{Legal Immigrants} respectively, which equates to about half of the values missing, the summary statistics are remarkably close to the original, raw values. This suggests that the other respondents in the neighborhood of the missing cases were quite helpful (and similar along other dimensions) in imputing these missing values. Regarding the third feature, \texttt{Buttigieg}, the summaries were much more closely aligned with the original, raw summaries. Substantively, the similarity across patterns of summaries suggests that respondents are revealing relatively consistent signals about preferences across these feeling thermometers, regardless of the issues. The redundancy of signaled information across the feeling thermometers allows for an ideal scenario to leverage dimension reduction to rid our data space of redundancies, and instead retain the most unique, useful information and patterns underlying the data space.

\clearpage

\section{A Classic Approach to Dimension Reduction} 

The main idea with dimension reduction is to reduce complexity (that is, \textit{dimensionality} for present purposes) of a data space to create a lower dimensional representation of that original space. Such a task makes data more accessible, patterns more intuitive, and thereby eases to task of detecting and interpreting natural structure. 

Perhaps the most commonly taught dimension reduction technique is principal components analysis (PCA), which is a linear approach to dimension reduction that reduces complexity by maximizing variance across the data space. The starting place with PCA is assuming \textit{variance} is the best way to conceptualize structure in a data space. As such, variance and specifically \textit{shared} variance will be a key theme in covering this classic approach to dimension reduction. 

Applying PCA requires some information to be willfully thrown out. This might make some feel uncomfortable, but our main goal of dimension reduction in the first place is to \textit{simplify} a complex higher dimensional space. This could be as simple as moving from 2 dimensions to 1, or as complex as moving from 5000 dimensions to 2. Regardless of the complexity of the problem, by reducing dimensionality of the data space, information is by definition intentionally being discarded as we decide what the lower dimensional configuration should look like. As such, when dimension reduction is the goal, this ``downside'' of discarding information for the sake of a simpler representation is in reality usually not considered a downside at all, because we are assuming the simplified space retains enough of the information from the higher dimensional space to allow for an understanding of the \textit{structure} of the space. Put differently, and more bluntly, dimension reduction in these terms focuses on the most interesting parts of the data, and gets rid of the less interesting parts. In so doing, we are able draw conclusions about \textit{latent} structure. In terms of PCA, then, this structure is defined by shared variance across all features. 

\subsection{Why PCA?}

Before diving into PCA, why does it make sense to leverage PCA for dimension reduction? As referenced in different ways to this point, PCA makes a high dimensional space less complex, and thus more interpretable. With this overt benefit, come several additional benefits such as visualization, understanding patterns underlying the data, and ``feature extraction'' for creating new features to be used for other tasks, e.g., prediction.

Further, PCA helps with diagnosing and bypassing limitations caused by multicollinearity across the feature space. As hinted at in the previous section, the feeling thermometer space in the ANES data is likely characterized by a sizeable amount of overlap in responses to the battery of issues. For example, feelings toward Joe Biden, Barack Obama, and Elizabeth Warren are likely picking up on commonality in preferences of respondents, as these people are all Democrats who ran for the U.S. presidency, and who have commanded active media attention. Similarly, feelings toward the U.S. Immigration and Customs Enforcement Agency (ICE) and feelings toward immigration likely correlate as well, either negatively or positively, as these are addressing different aspects of a common underlying concept: immigration. The expectation with features that correlate then, is that they also explain similar variance in preferences of people who responded to the ANES survey. Substantively, many of the features in the full feeling thermometer space might be more clearly and parsimoniously characterized when we focus on uncovering the underlying, shared patterns (e.g., immigration), which is ultimately recorded in a reduced version of the data space that is picking up on these shared aspects of respondents' preferences. 

Further, and closely related, dimension reduction helps guard against the \textit{curse of dimensionality}, which asserts that as the dimensionality of the space increases, true similarities and differences across the dimensions/features becomes less clear. To guard against obscuring real similarities and real differences that naturally exist in the data in higher dimensions, dimension reduction, which simplifies this complex space by \textit{removing dimensions}, is an extremely useful tool.

To make these ideas, and thus the benefits of dimension reduction come alive, I shift to show correlations across all feeling thermometer features to help us clarify this assumption of likely shared variance and correlation across features. Importantly, correlation and variance are distinct statistical concepts, which are unpacked in the formal definition of PCA below. But to this point, they are used interchangeably to draw out the substantive value of PCA for dimension reduction. Another way to put this is to begin by diagnosing the space to get a clue as to whether structure and \textit{similarity} naturally exist in the feeling thermometer space.

\begin{figure}[h!]
	\centering
	\includegraphics[scale = 0.5]{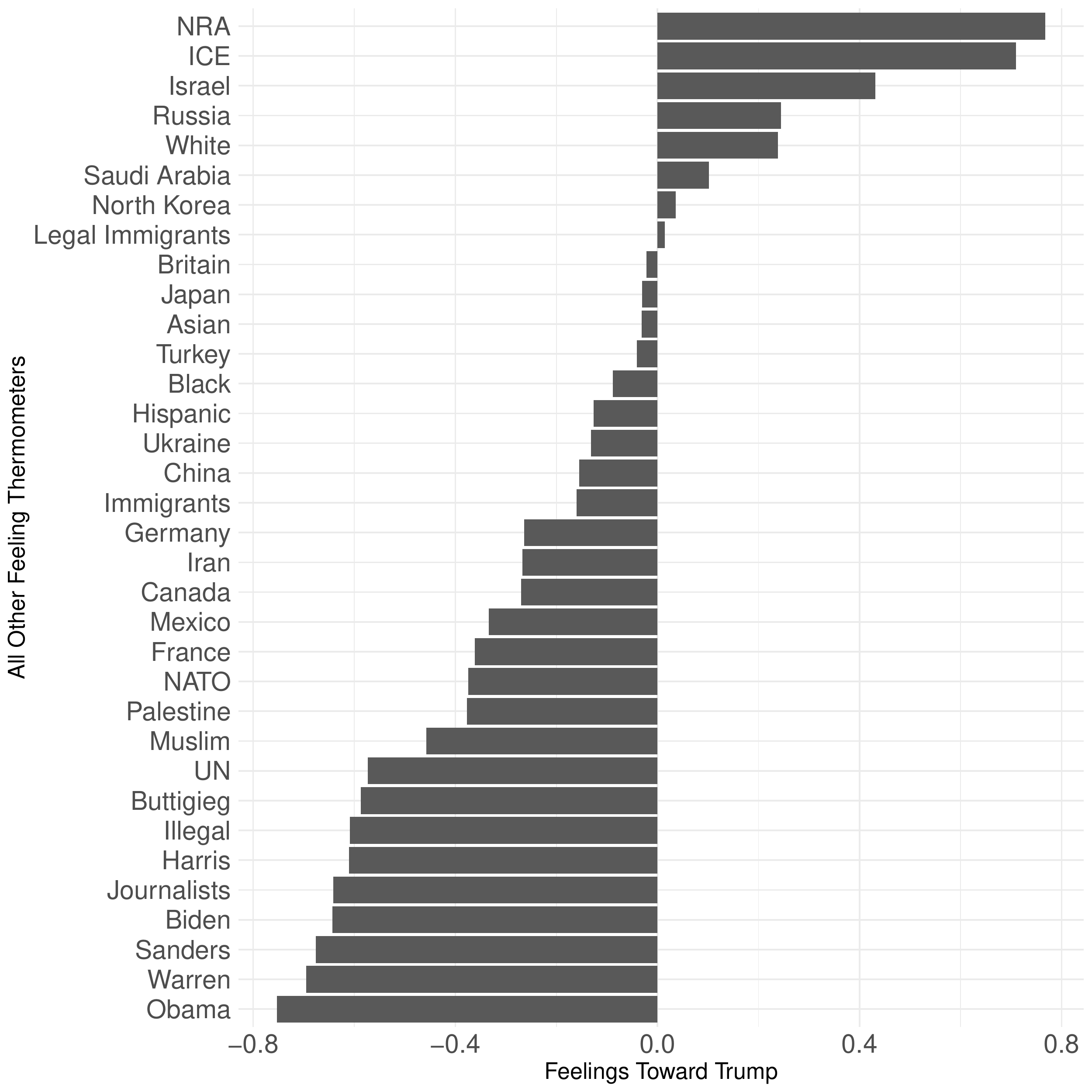}
	\caption{Correlation between Features and Trump}
	\label{figure:corr1}
\end{figure}

In this demonstration, several view of correlations across all feeling thermometers are provided. First, I compare all features to feelings toward Donald Trump. The justification for doing so is purely descriptive, where Trump (being a Republican president) and frequently in the news, is a political figure allowing us to check some base expectations. For example, we might expect feelings toward Barack Obama (presidential predecessor and member of the opposite party) would be strongly negatively correlated with feelings toward Trump. To do so, we load the cleaned version of the ANES data from the previous section, and then work primarily with the \texttt{corrr} package from the tidyverse. Importantly, as \texttt{corrr} is apart of the tidyverse, we can directly pipe plotting functions giving a clear rendering of the correlations in Figure \ref{figure:corr1}.

Examine the correlations between all features and feelings toward Trump in Figure \ref{figure:corr1}. Indeed, as expected, several features correspond with an intuitive, base set of expectations. For example, feelings toward Barack Obama are indeed most strongly and negatively correlated with feelings toward Donald Trump. And feelings toward the National Rifle Association (NRA) most strongly and positively correlated with feelings toward Trump. 

\begin{figure}[h!]
	\centering
	\includegraphics[scale = 0.5]{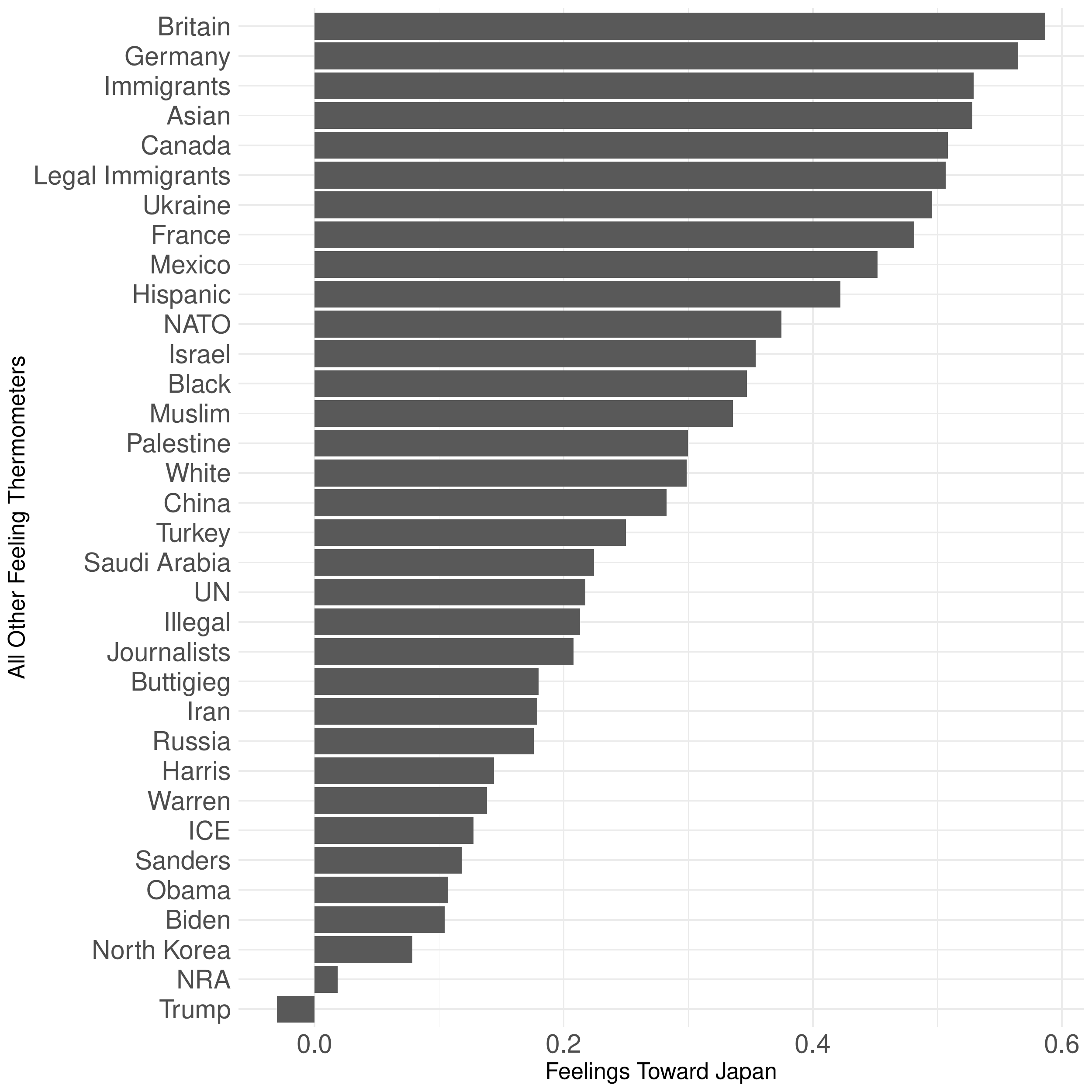}
	\caption{Correlation between Features and Japan}
	\label{figure:corr2}
\end{figure}

Whereas feelings toward Donald Trump took shape in a largely intuitive way given the salience of Trump's presidency and the media attention he commands, we turn now to a different case, and check correlations in another light by exploring across all features in relation to feelings toward Japan, where expectations of correlation patterns are perhaps less obvious. See the results in Figure \ref{figure:corr2}. 

Notably, correlations between all features and Japan are relatively strong and positive, with the exception of feelings toward Trump. Though interesting and confusing, possible reasons for exactly why this is the case are beyond the scope of present purposes. Rather, I am interested in demonstrating that there are clearly strong correlations across the data space, some intuitive (Trump) and others less so (Japan). 

We continue this exploration of correlations naturally in the feature space, which will deepen an answer to our motivating question on why it is useful to pursue PCA to simplify this space. Rather than explore correlations between a single feature and all others, we can instead view a network representation of the correlations that naturally exist in feature space, which provides a fuller, more nuanced picture of degrees and directions of correlations across \textit{all} features. To do so, we still rely on core tidyverse packages, \texttt{corrr} and \texttt{ggplot2}. But for this case, we leverage a different function, \texttt{network\_plot()}, which gives the network. Further, we use the \texttt{amerika} package to color the network \citep{amerika}, as well as many of the visualizations used throughout the Element. 

\begin{figure}[h!]
	\centering
	\includegraphics[scale = 0.5]{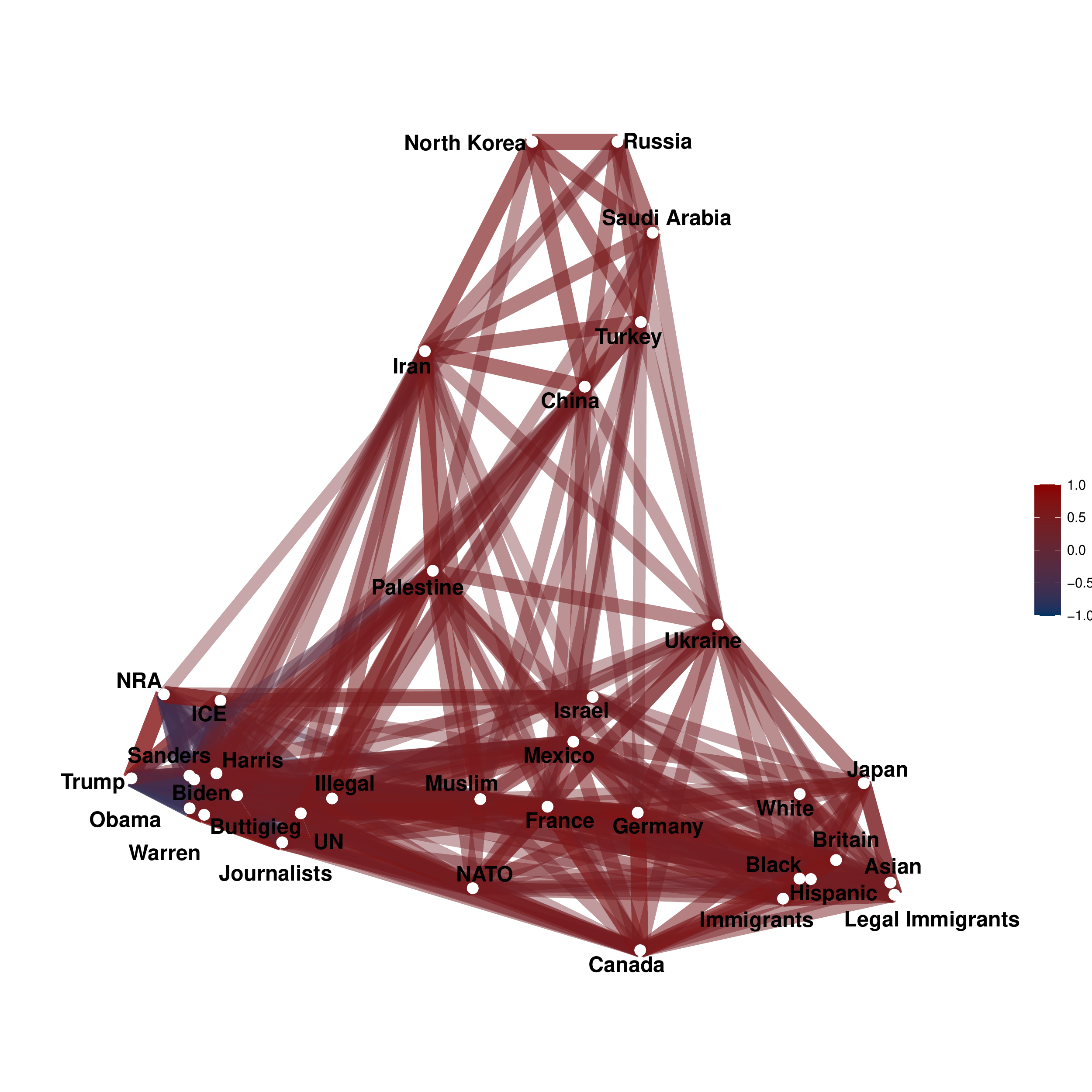}
	\caption{Network Representation of Feature Correlations}
	\label{figure:corrnet}
\end{figure}

A few useful trends emerge from the network configuration of correlations in the full features space shown in Figure \ref{figure:corrnet}. First, there is strong correlation (darker shades) across much of the plot, implying strong correlations across the \textit{full} high dimensional data space. Related, a clear structure seems to characterize this space in a way we might expect, whether an ``expert'' or not in American politics. That is, feelings toward countries are largely grouped together in the upper part of the plot, feelings toward issues are largely grouped together in the lower right of the plot, and feelings toward people tend to be grouped together on average in the lower left of the plot in Figure \ref{figure:corrnet}. These groupings emerge from the construction of the \texttt{network\_plot()} function, which groups based on the strength of correlations within a subgroup of features. The goal, really, is to push these groups apart from one another, again on the basis of natural structure, to essentially exaggerate differences between features on the basis of the strength of the correlations. This plot is similar to the concept of modularity in graph theory, where stronger ties within a module/cluster as well as sparser connections between modules or clusters implies latent structure in a common space. Similarly, some latent structure based on clear collinearity across features is present in this space. The results across Figures \ref{figure:corr1}, \ref{figure:corr2}, and \ref{figure:corrnet}, then, offer sound motivation to move forward with dimension reduction. 

\clearpage

\subsection{What is PCA Doing?}

We have a clear sense that the feeling thermometer feature space is indeed highly correlated. With this information, we might conclude that it makes sense to progress with PCA to draw out and isolate dimensions of greatest variance underlying the data. Yet, before formalizing PCA, we must first address what it is doing, or precisely how it handles the task of dimension reduction. I prefer to start in words, and then use equations to clarify.

\begin{figure}[h!]
	\centering
	\parbox{7cm}{
		\includegraphics[width=7cm]{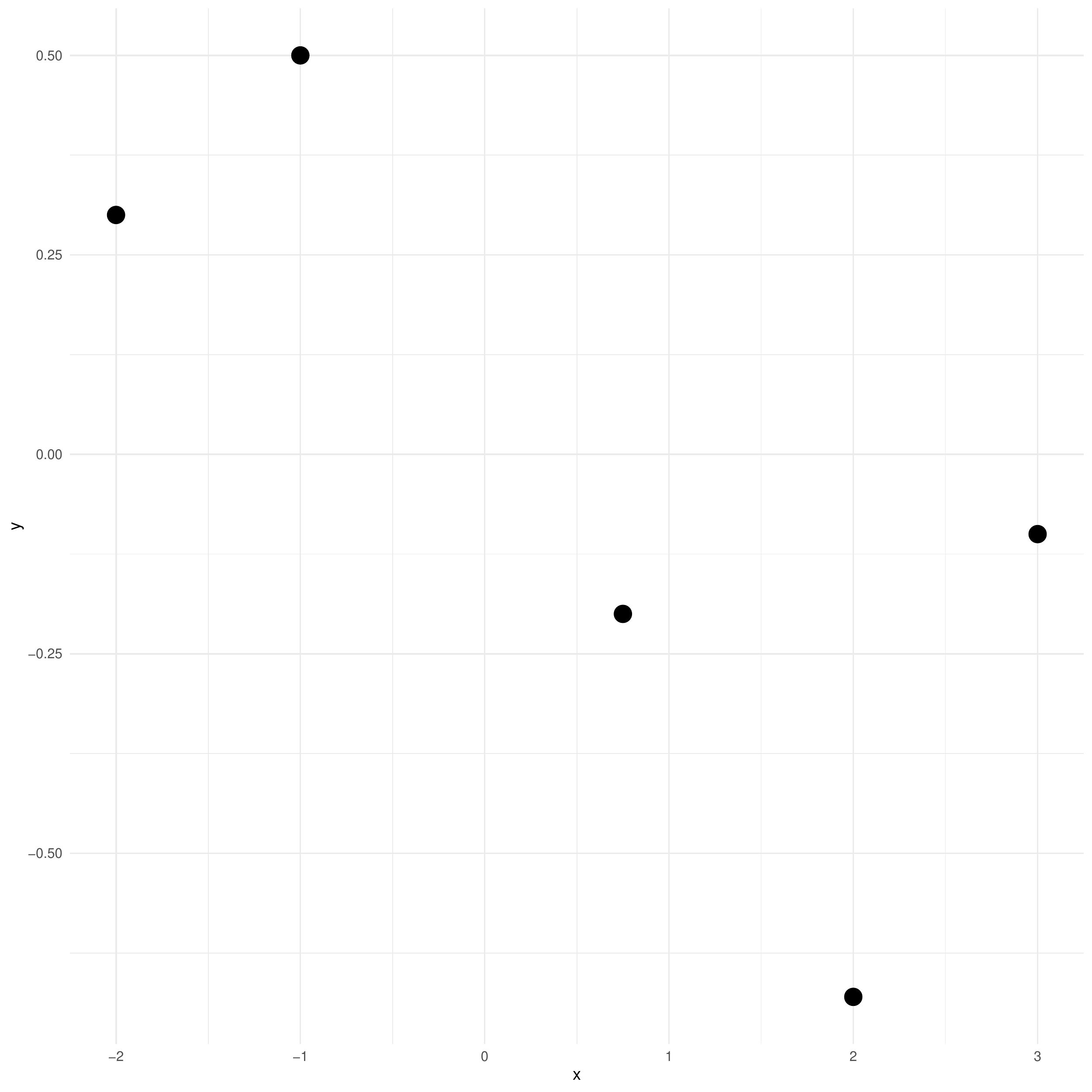}
		\caption{Five Data Points}
		\label{figure:five}
	}
	\qquad
	\begin{minipage}{7cm}
		\includegraphics[width=7cm]{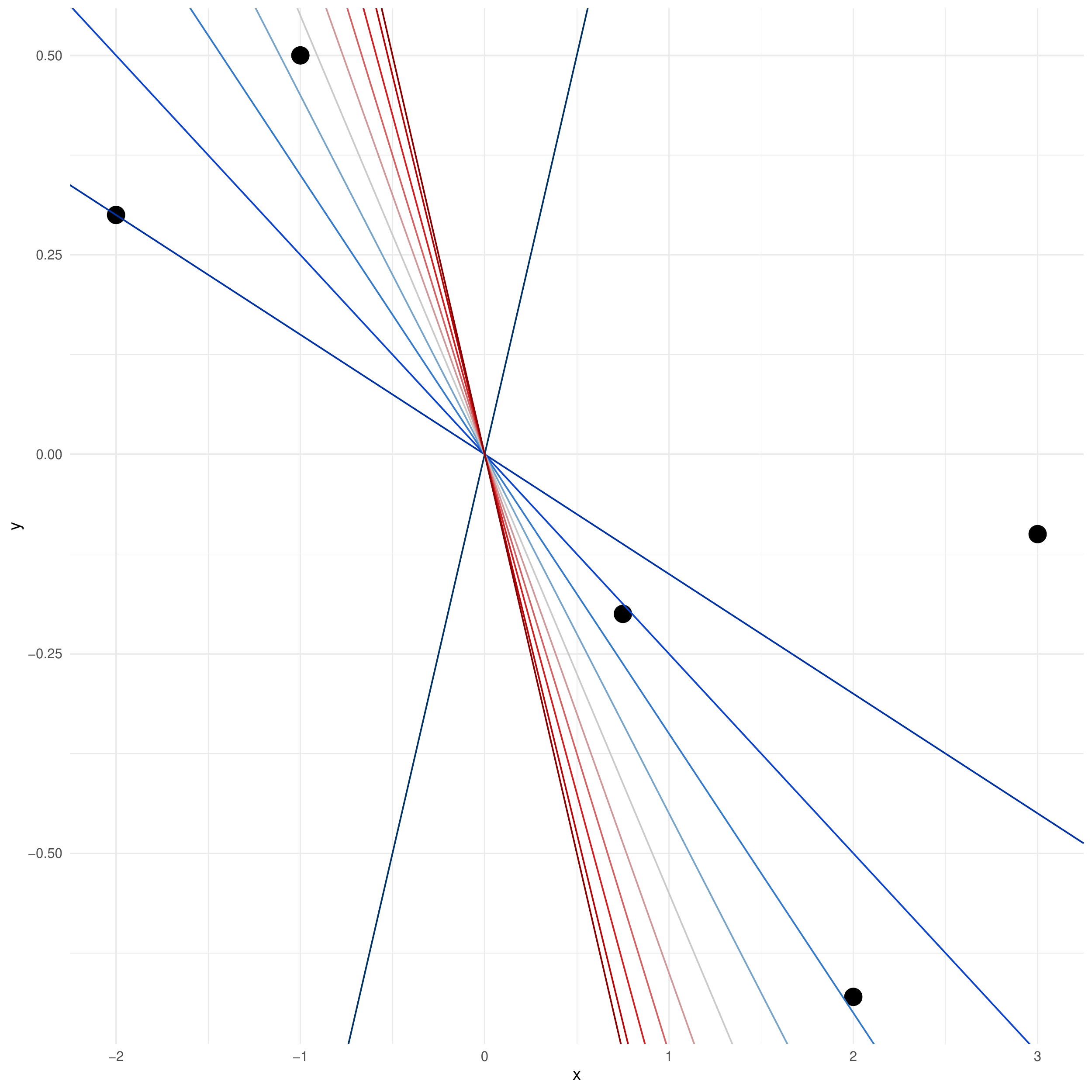}
		\caption{Summary Options}
		\label{figure:options}
	\end{minipage}
\end{figure}

When approaching a task of summarizing data for the purpose of making it simpler, which is at the heart of dimension reduction, there are many ways to go about this. For example, we might overlay some line to summarize a bunch of observations. There are many possible lines that \textit{could} exist as a summary, with some better than others. Consider the hypothetical case of five data points in Figure \ref{figure:five}, along with some candidate summary options in Figure \ref{figure:options}. 

\begin{figure}[h!]
	\centering
	\includegraphics[scale = 0.5]{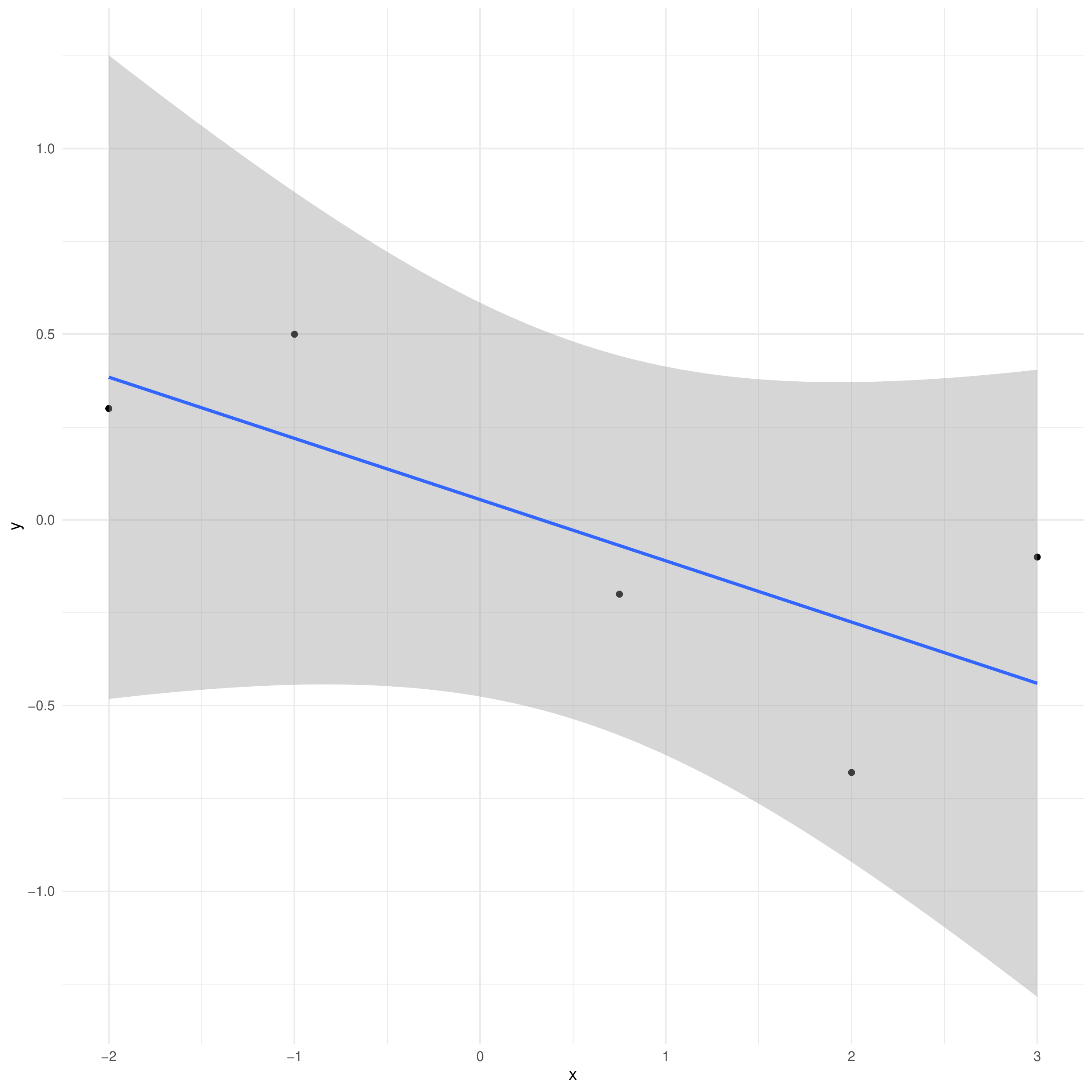}
	\caption{Linear Regression Approach to Summary}
	\label{figure:lm}
\end{figure}

Such an incremental approach to honing in on an optimal summary of data shown in Figure \ref{figure:options} not only inefficient, but also makes it impossible to hone in on an \textit{optimal} option absent a definition of \textit{optimality}. Simple linear regression handles this problem in a parametric way, by estimating parameters (an intercept and a slope) that minimize the sum of squared residuals, on the basis of which we can represent this summary with a line through the data, with some uncertainty of course, given the process of parameter \textit{estimation}. See this approach in Figure \ref{figure:lm}. 

Rather, PCA, which is technically a special case of linear regression, but with no estimated parameters and thus no intercept, approaches this problem from a different perspective with a different goal in mind. In PCA, the focus is on summarizing a higher dimensional data space by focusing on \textit{maximizing the total variance} of that space. As variance is the focus, thereby giving us a definition of \textit{optimality}, PCA is initialized to look for the direction in the data along with the full data vary most. Once that direction is found, PCA computes and places a summary line called a \textit{principal component}, that summarizes the direction of maximal variance. We can call this first principal component $C_1$. 

If we stopped with $C_1$, we would be explaining a good amount of variance in the data in most cases, though not the total amount by definition. Variance, especially in high dimensional contexts, can be complex and proceed in many directions. Thus, once $C_1$ is found, PCA proceeds to search for the next direction in the data along which the second-most amount of variation exists. We can call this second principal component $C_2$, which now summarizes the unique, \textit{remaining} variance after accounting for $C_1$. This process is continued until we have explained the \textit{full} data space, such that $C_* = p$. Sticking with the number of components the size of the dimensionality of the full data space is equivalent to saying we have summarized the entire amount of variance in original setting. At this point, it would not make any sense to continue with PCA, as we would be back in the high dimensional setting, where $C_* = p$. Instead, our task is to home in on some subset of the components, $C < p$, to accomplish our goal of \textit{simplifying} the high dimensional, complex data space. 

As the algorithm searches for and places new prototype components, $C$, to summarize unique variance in the full data space, \textit{uniqueness} is defined by a requirement of each component being orthogonal to all others. This means we can only define a new principal component if it is explaining completely unique, previously unexplained variation in the data.

\begin{figure}[h!]
	\centering
	\includegraphics[scale = 0.38]{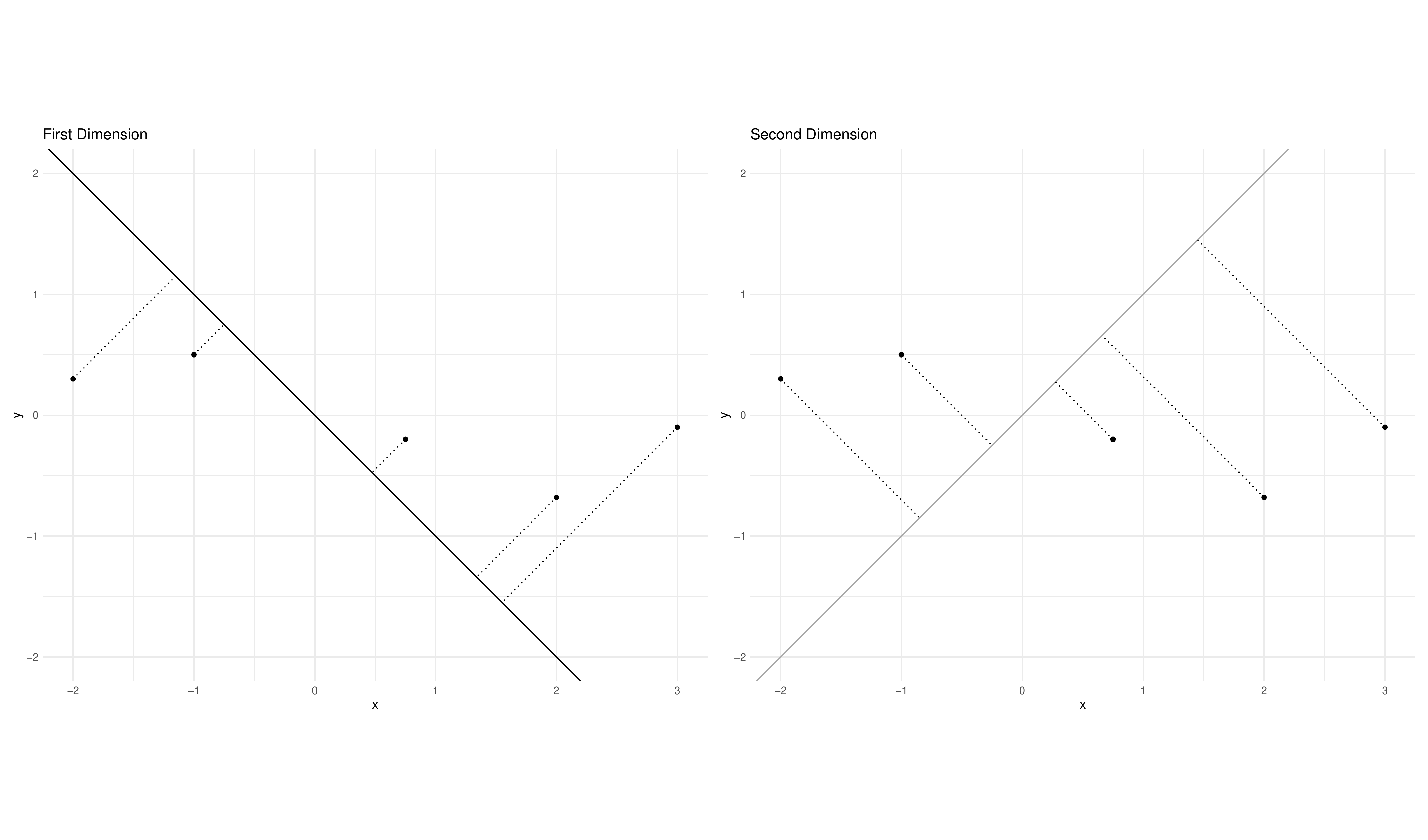}
	\caption{Projecting Points onto Components}
	\label{figure:twocomps}
\end{figure}

Importantly, as we are operation in some $d$-dimensional space, placement of components is defined both by feature values and observation placements in the high dimensional setting. That is, part of the component placement is constrained by the projection of individual points onto the component. In so doing, we are left with component \textit{scores} across each component. Returning to our simple example in Figure \ref{figure:five}, we can see what this process looks like for two components in Figure \ref{figure:twocomps}.  

The left plot in Figure \ref{figure:twocomps} is showing the first principal component (first dimension), which is summarizing the maximal variance in the data, which in the case is a pattern that extends diagonally from the upper left of the plot to the lower right of the plot. Individual points are projected onto the component, giving these points new measured values in the first dimension of the new lower dimensional space. Think of these component scores the same as the measured values for any feature like self-reported political ideology. Then, the right plot in Figure \ref{figure:twocomps} shows the second component, which is orthogonal to the first. Similarly, points are projected onto the component, and these scores are the new ``measured'' values for the second dimension. If a researcher stopped at this point, then $C1$ and $C2$ could be used as input features for some predictive modeling task. 

\begin{figure}[h!]
	\centering
	\includegraphics[scale = 0.5]{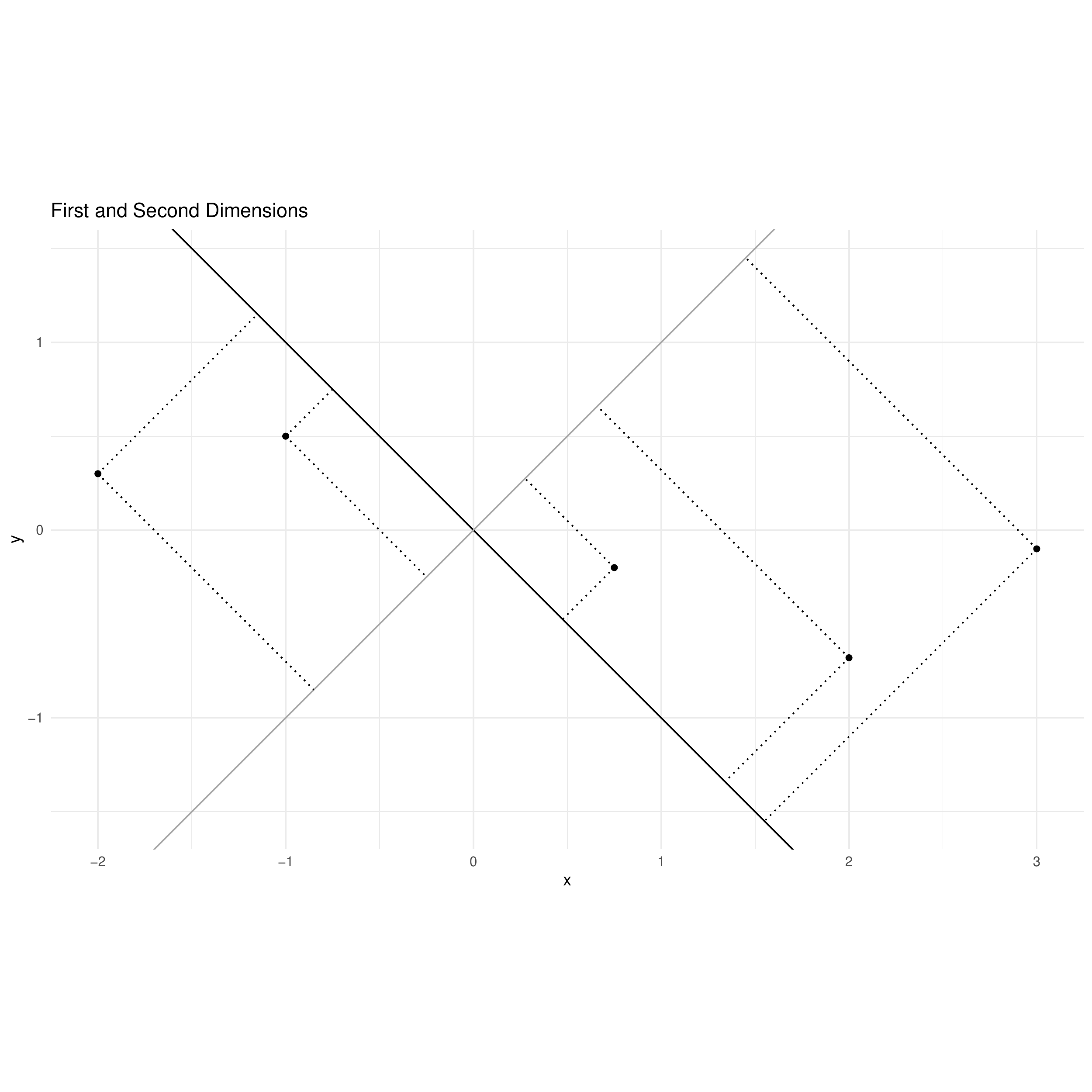}
	\caption{Full PCA Solution for Simulated Points}
	\label{figure:fullcomps}
\end{figure}

Thus, upon finding these components, we would get the full solution shown in Figure \ref{figure:fullcomps}. Figure \ref{figure:fullcomps} more clearly shows the \textit{uniqueness} aspect of the PCA solution, where each component is orthogonal to the other. 

\subsection{Formalizing PCA}

With the previous substantive discussion in mind, we can formalize these ideas with equations. Recall I mentioned in the previous subsection that PCA is a special case of linear regression, just without estimated parameters or an intercept. We can see that in the base construction of PCA for the first component, $C_1$,\footnote{I use $\beta$ to emphasize the link to linear regression, given the wide spread familiarity with regression. The weights are essentially operating in very similar ways as an estimated $\beta$ coefficient in linear regression.} 

\begin{equation}
C_1 = \beta_{11}X_1 + \beta_{21}X_2 +\ \dots\ + \beta_{p1}X_{p}
\label{eq:pca1}
\end{equation}

where, $\beta_{11}$ is a weight for the first component on the raw feature $p \in \{1, \dots, P\}$. From this form, we can see the unique contribution of each feature to the calculation of each component, $C_*$. These weights are typically called \textit{loadings}, which captures the idea of each raw feature, $p$, loading differently onto each calculated component, $C_*$. The relation to the simple linear regression should be clear from Equation \ref{eq:pca1}, where the component, which recall is also a summary line that passes through the data and the origin, is a function of up to $p$ unique contributions. 

Importantly, with PCA and all dimension reduction techniques covered in this Element, standardization of the features is critical, given unique raw feature values can take over variance of other features if these features are on different scales. For example, if there were two features weight in pounds and income in U.S. dollars, the vastly different scale of the features would result in an imbalance of contribution to the calculation of the component solely on the basis of differences across their scales, rather than unique variance in relation to all other features. To account for this issue, standardization, which is simply defined by dividing each feature by it's standard deviation, guards against this threat. The result is all features are allowed to be directly compared in a scale-free way. 

Then, for each subsequent component, we simply update the index notation (e.g., $\beta_{11} = $ feature 1, component 1), and find the next component, $C_*$, orthogonal to and thus uncorrelated with the preceding components. Such a strategy allows us to continue to pick up unique, left-over variance with our solution,

\begin{equation}
C_* = \beta_{1*}X_1 + \beta_{2*}X_2 +\ \dots\ + \beta_{p*}X_{p}
\label{eq:pca2}
\end{equation}

\citet{james2013introduction} offer a simple framing of the problem of finding optimal $\beta$ values for each component, $C_*$, to maximize the \textit{sample variance} across the observations, $i \in \{1, \dots, n\}$, and features, $p \in \{1, \dots, P\}$,

\begin{equation}
\frac{1}{n}\sum_{i=1}^{n}\left(\sum_{p=1}^{P}\beta_{p1}X_{ip}\right)^{2}. 
\label{eq:pcaopt}
\end{equation}

Viewed as an optimization problem, finding optimal $\beta$ values in the loading vector, $\beta_* = \{\beta_{1*}, \dots, \beta_{p*}\}$, can be solved using many techniques such as singular value decomposition (SVD). Overly simplified and in words, SVD is generally comprised of three steps. First, project the observations on the component, and store the coordinates. Second, calculate the distance from each point to the origin, which always has cartesian coordinates, $(0,0)$. Third, square each of these values and add them together. This series of steps gives the eigenvalue (EV) for each principal component. Values are squared, because $\sqrt{EV}$ is equal to the \textit{singular value}, which is involved in the decomposition of $\mathbf{X}$. For a more thorough treatment of SVD, see chapter 14.5 in \citet{friedman2001elements}.

A final step in a PCA solution is to decide on the number of components to retain, which is the step referenced a few times to this point deciding ``how much information to throw out and how much to keep.'' Though there is no formal rule for determining this, there are a number of descriptive techniques that can help. But before getting to these, a final definition that is integral to understanding PCA is the proportion of variance explained (PVE). The PVE is a normalized (to equal 1 across all summed components) value that gives an indication of each component's contribution to the full PCA solution. Again, though no clear rule exists for evaluating these as it relates to determining the optimal number of components to retain, it is reasonable to suggest a total PVE of around 75-80\% is a fair base line, as this could include either a single component that is doing the bulk of the explanation, or it could include several components that are contributing to a simpler version of the high dimensional space.

\subsection{Applying PCA to the ANES Data}

With a clear idea of why PCA is useful, what PCA is doing, and how PCA works to reduce dimensionality, we conclude this section with an application of PCA using the 2019 cleaned ANES data. Once applied, we will discuss the output and the several options for evaluation and interpretation. 

There is a remarkably small amount of code needed to fit a PCA model in R. We will be using the \texttt{prcomp()} function from base R, given the long-standing status of PCA in applied statistics and statistical computing. Some other packages have PCA functions such as the \texttt{FactoMineR} or \texttt{adea4} packages. Yet, in practice it is much more common to use \texttt{prcomp()} to fit a PCA model given it's computational efficiency and simplicity. As such, this is where we start as well. 

We will work with a new package for excellent, simple plotting options, which is built upon the tidyverse's \texttt{ggplot2} package. With the data loaded, we fit the PCA model on all feeling thermometers in the cleaned dataset, withholding the party feature, \texttt{democrat} in the 36th column of the data matrix, to use for visualization of PCA scores.

When we summarize the model output, the ``Importance of Components'' is returned. The PVE, which is the second row of values, tells us the proportion of variance explained by each calculated component. Recall, the PVE tells us how much of the unique variance is explained by each of the PCs. As expected, the PVE is decreasing as we move from left to right as the requirement for defining uniqueness in PCA is defined by subsequent components being uncorrelated and orthogonal to all preceding components. Thus, we are left with progressively less variance to be explained as we continue to find and calculate components. Related, note that the solutions returns 35 components (denoted by $PC1$, $PC2$, and so on). This is the case, because as previously mentioned, when we explain \textit{all} of the variance in a data space, we are back in the high dimensional setting, which by definition is fully explaining itself on the basis of the inclusion of all features. The task of PCA, then, is to hone in on a \textit{reduced} version of the full space on the basis of explained variance. The PVE, naturally, helps us out with this task. 

Next, and related, we can see the cumulative PVE values (CPVE). These values are the inverse of the PVE, where they can be progressively summed, and will eventually equal 1, once summed across all components. For example, we see a clear jump with no component ($CPVE = 0$) to the $PC1$, which has a $CPVE = 0.3599$. Then, when we proceed to $PC2$, we get an increase in PVE of $0.1657$, as CPVE is at 0.5256 when accounting for PC2, minus the PVE 0.3599 based only on PC1, gives an increase of 0.1657 moving from PC1 to PC2. Summing PVE for each subsequent PC, we get a CPVE of 1.000 at $PC35$, meaning we have now explained the full data space, such that $C = P$.

Finally, the model output returns the standard deviation (first row). Recall, in statistics the standard deviation is a measure of spread and is defined as the square root of the variance. And recall also, we previously noted that in PCA the variance is defined by the eigenvalues across the components. And finally, recall that we said the singular value is simply the square root of the eigenvalues. Thus, we interpret the standard deviation from PCA output as the square root of the eigenvalues computed for each principal component. The decrease in standard deviation values as we move from the first to the final principal component, thus, make sense, as we are left with progressively less variance to explain once we reach $C = P$.

We can also call the loadings, which are feature contributions to each principal component, by calling \texttt{pca\_fit\$rotation}. The output is omitted due to its size. Yet, a more effective method for evaluating PCA output is visually. 

\begin{figure}[h!]
	\centering
	\includegraphics[scale = 0.387]{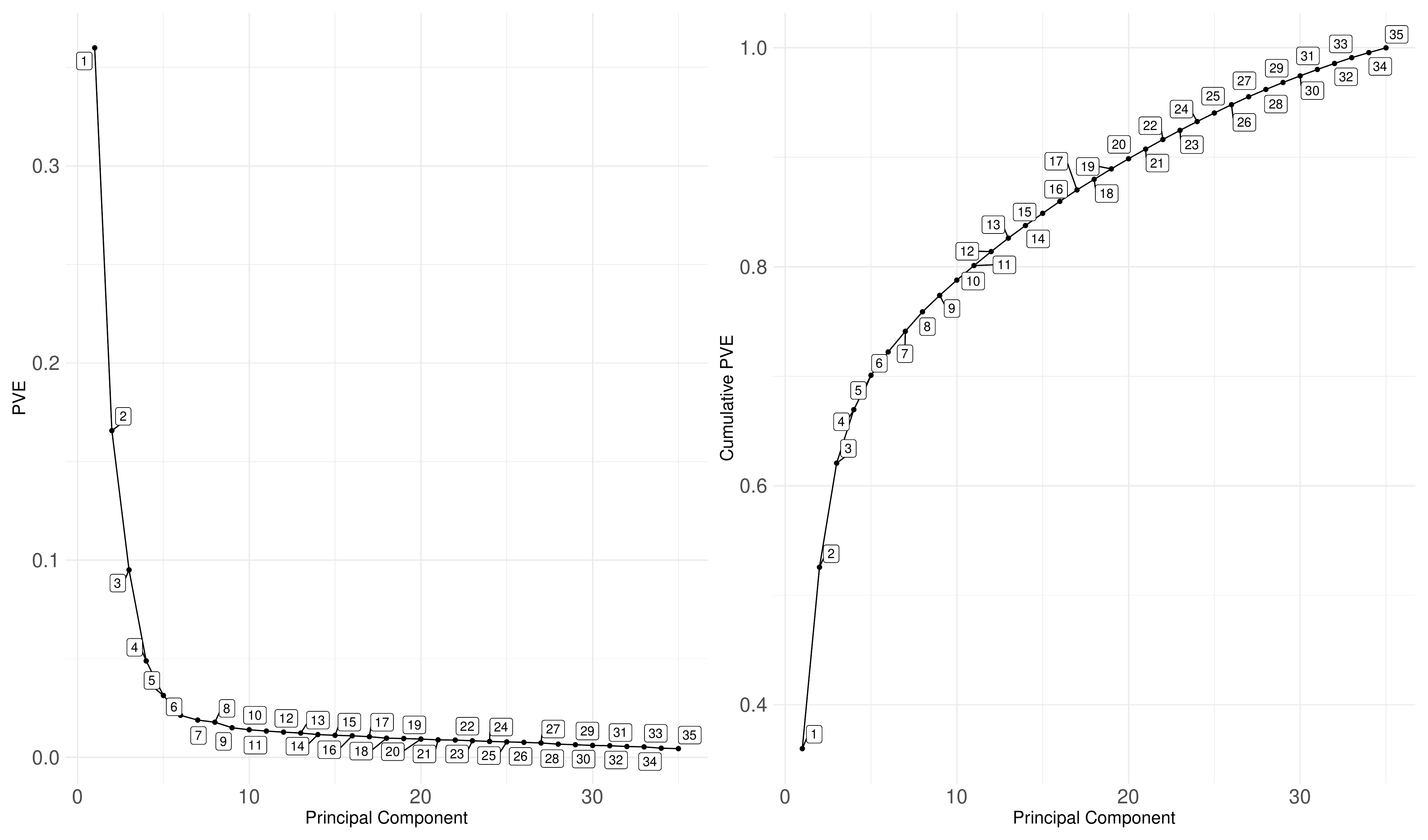}
	\caption{PVE and CPVE for the PCA Solution}
	\label{figure:pvecpve}
\end{figure}

To do so, we start by visually evaluating the structure of the space, which builds on the previous numeric exploration of the PCA output. We will manually calculate the PVE and CPVE, and create two \texttt{ggplots} for each respectively. These plotted values over each component are included in Figure \ref{figure:pvecpve}. Running the respective code will first make the calculations and store the values accordingly. Then, the subsequent code will produce both plots side-by-side, with labels according to the component number, ranging from 1 to 35 and created using the \texttt{ggrepel} package.

Indeed, in Figure \ref{figure:pvecpve}, we get visual corroboration with the numeric output, that the first few principal component explain the majority of the variance. Even the first two components explain over half of the data, suggesting a large amount of correlation in the full feature space. 

\begin{figure}[h!]
	\centering
	\includegraphics[scale = 0.38]{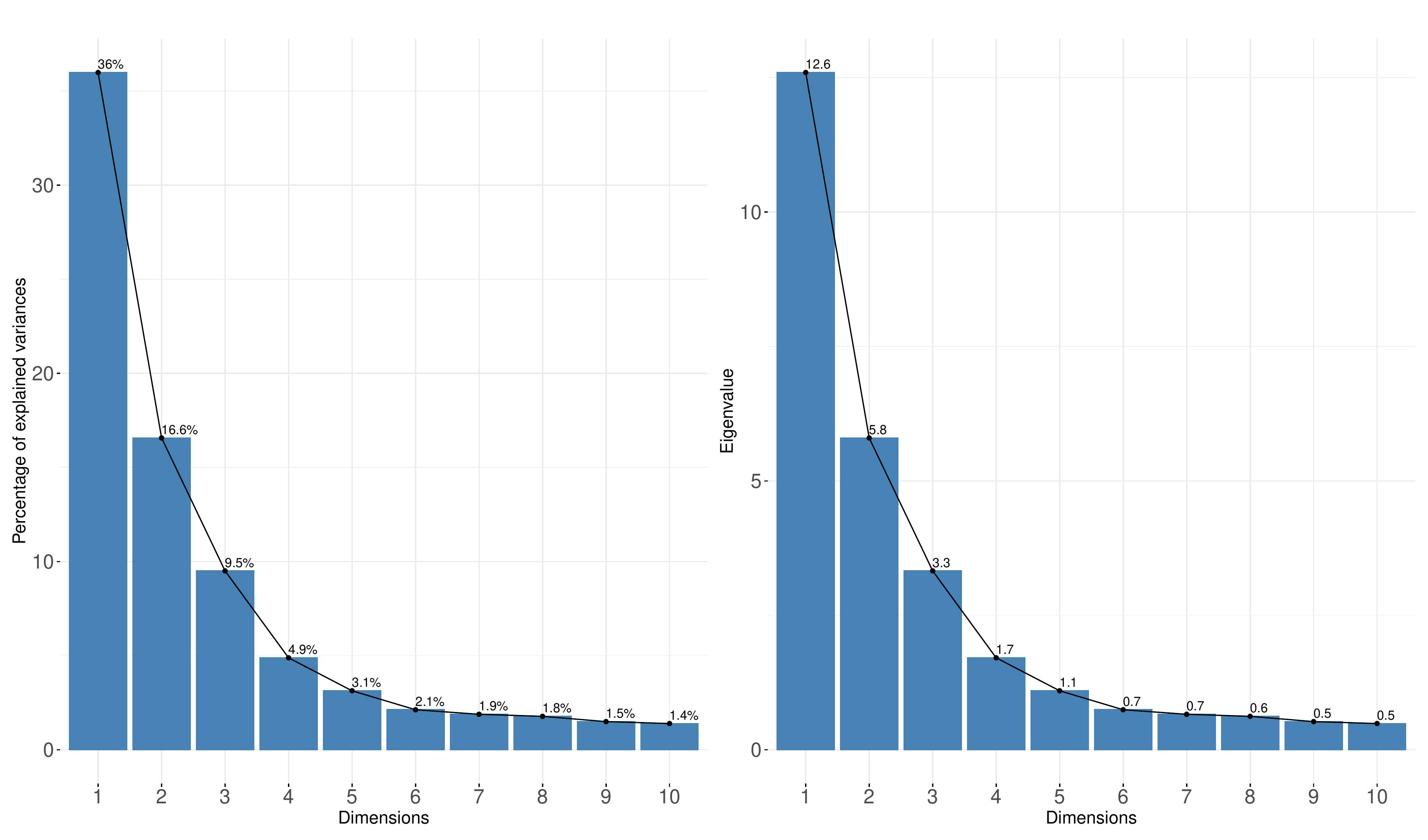}
	\caption{Scree Plots of Variance Explained (Percentages \& Eigenvalues)}
	\label{figure:screes}
\end{figure}

Another popular approach is to use a scree plot to evaluate the dimensionality of a space. For this approach, the \texttt{factoextra} package includes options for scree plots for either \textit{percentage} of variance explained, or \textit{eigenvalues} for each principal component. For either version, which give identical information as each are capturing differences in variance by component, simply call the \texttt{fviz\_screeplot()} function, and specify the proper \texttt{choice}, either ``variance'' or ``eigenvalue.'' See these results in Figure \ref{figure:screes}.

Figure \ref{figure:screes} shows the first few dimensions/components seem to be explaining the bulk of the variance.

Though we have options to explore variance explained by the PCA fit, we need to use this information to determine how many dimensions to retain in our simplified version of the data space. Some make suggestions based on total variance explained as previously discussed, and others suggest components with eigenvalues greater than 1. Still others suggest looking for the ``knee'' or ``elbow'' of the scree plot to make a get decision. If all of these approaches sound murky, that is because they are. There is no formal guidance on the optimal number of components to retain to define the lower dimensional space. The best we can do is inspect the results in several ways, as we have done to this point, and then make a decision. Thus, across all of these suggestions, I would conclude that likely 4 dimensions characterize the space well, as we see a drop off in PVE, percentage, and eigenvalues around four and five components. Given that anything greater than 4 dimensions is virtually uninterpretable by our brains as discussed earlier in the Element, four dimensions seems like a reasonable cut off. Even still, the first two dimensions still capture a large amount of variance, which will be useful for plotting component scores at the conclusion of this section. 

Before we get to scores, though, two other useful ways to visually assess a PCA model, are a biplot and a plot of the feature loadings. We will walk through both using the \texttt{factoextra} package again for clean, simple code.

\begin{figure}[h!]
	\centering
	\includegraphics[scale = 0.5]{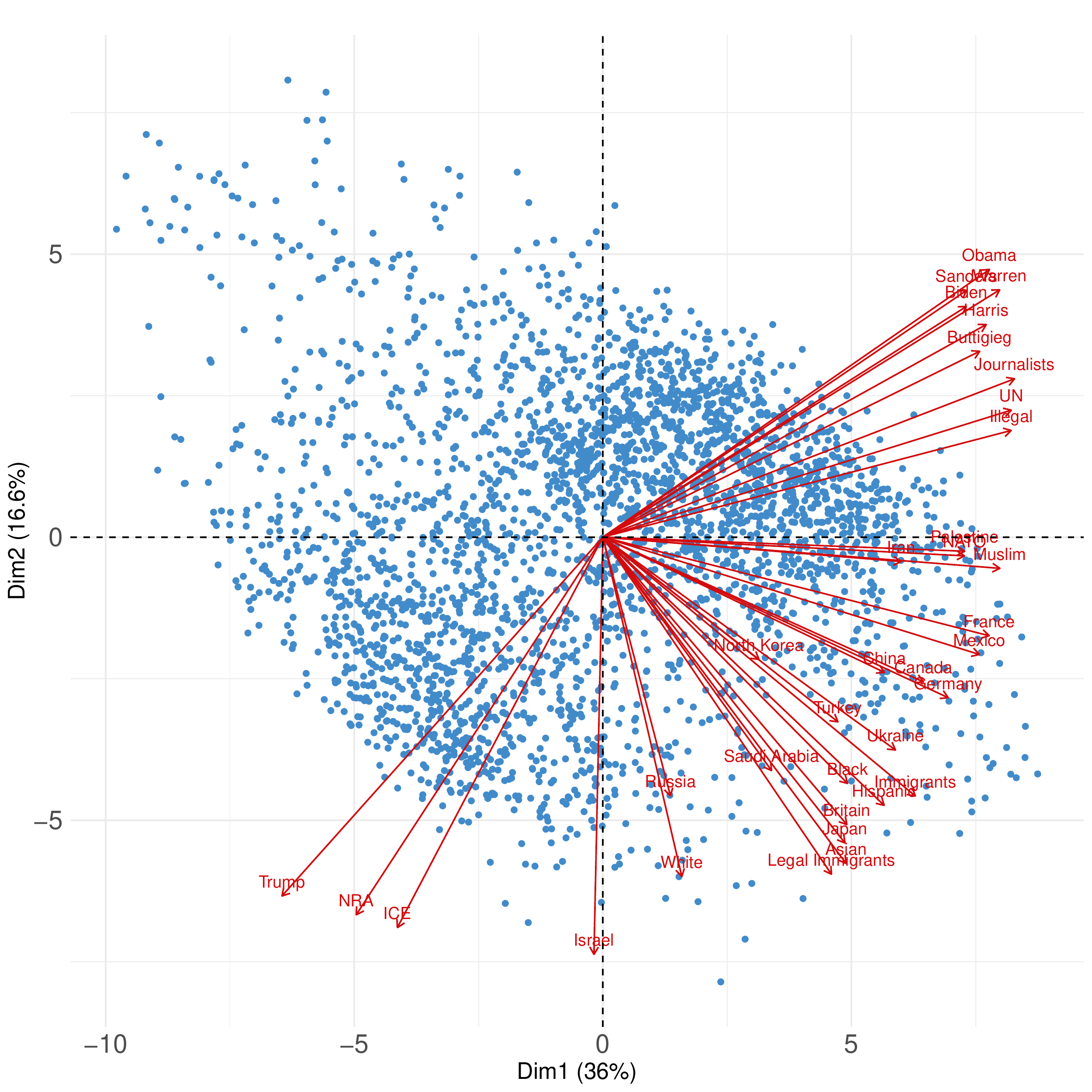}
	\caption{Biplot of PCA Fit}
	\label{figure:biplot}
\end{figure}

First, the biplot of the PCA fit in Figure \ref{figure:biplot} plots the scores in two dimensions, where dimension 1 is along the x-axis and the PVE is in parentheses, and dimension 2 is along the y-axis. The points are the component scores, which recall in two-dimensional space is the coordinates for the projection of points onto both principal components. The arrows in a biplot show the connection between features and each dimension. To create the biplot in Figure \ref{figure:biplot}, run the respective code for Section 3.

The dashed lines at $(0,0)$ in Figure \ref{figure:biplot} are for reference only. In Figure \ref{figure:biplot}, a clear pattern emerges building on the earlier base expectations regarding Trump and other concepts often associated with Trump (e.g., Trump, NRA, ICE, White, Russia, Israel) are characterizing the second component. The first component, though, is more diffusely characterized by most of the other feeling thermometers, though extremely closely by feelings toward Iran, Palestine, and Muslims. Thus, from this simple rendering of the PCA solution, we can start to get a hint of similarities across features, and how these can be more simply represented in a lower dimensional setting.  

\begin{figure}[h!]
	\centering
	\includegraphics[scale = 0.5]{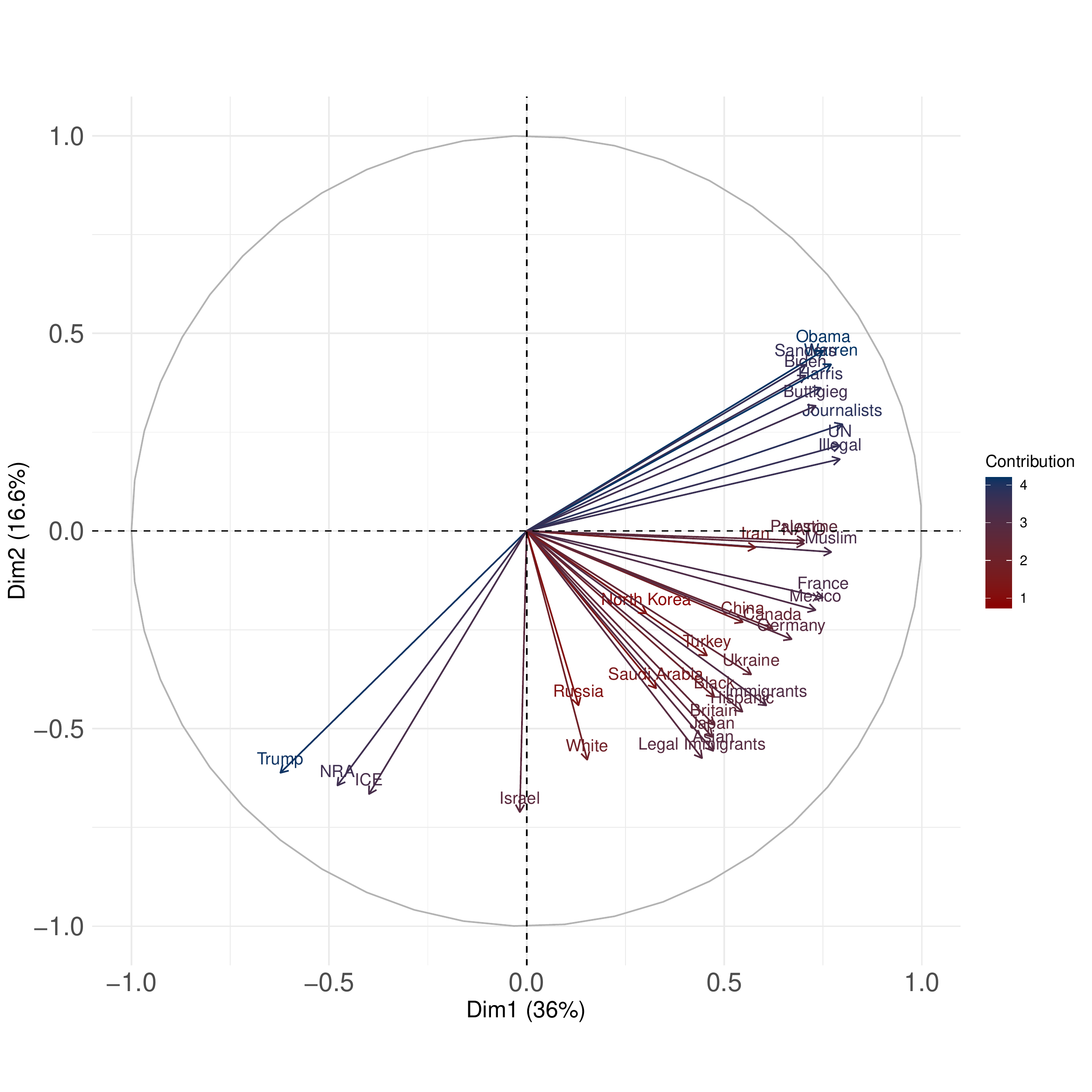}
	\caption{Feature Loadings from PCA Fit}
	\label{figure:loadings}
\end{figure}

Next, consider the other approach to visually interpreting PCA output by plotting the loadings Figure \ref{figure:loadings}.

To help us interpret the output from Figure \ref{figure:loadings}, it is useful to point out that a constraint is placed on the search for the optimal loading values in Equation \ref{eq:pcaopt}. That is, we are interested in maximizing sample variance across the data space, but subject to a normalization factor, 

\begin{equation}
\sum_{p=1}^{P}\beta_{p1}^{2} = 1.
\label{eq:normal}
\end{equation}

That is, the sum of the square loadings must add up to one. Though \citet[~376]{james2013introduction} and others emphasize this normalization constraint in the context of preventing ``arbitrarily large variance,'' another useful result from normalization of this sort is in the ease of interpretation. Restricting the size of the coefficients in such a way allows us to effectively interpret the loadings downstream as we might a correlation coefficient, especially because we also get negative and positive loadings where features load positively \textit{or} negatively onto different components. 

As such, interpreting Figure \ref{figure:loadings} showing the feature loadings on each of the first two dimensions, we start by inspecting the tip of the arrow. At the tip of the arrow, the contribution of the feature to the component's calculation is either negative if it is to the left of (below) 0.0, or positive if to the right of (above) 0.0. Shorter arrows, then, reflect less correlation with, or contribution to, the dimension(s). Longer arrows reflect greater correlation with, or contribution to, the dimension(s). For example, in Figure \ref{figure:loadings}, the \texttt{Israel} feature negatively loads onto the second component to a degree of about $-0.75$. All input features can be interpreted accordingly with a feature loadings plot. And of note, readers can double check the loading values by calling them directly from the PCA fit via \texttt{pca\_fit\$rotation}.

\begin{figure}[h!]
	\centering
	\includegraphics[scale = 0.5]{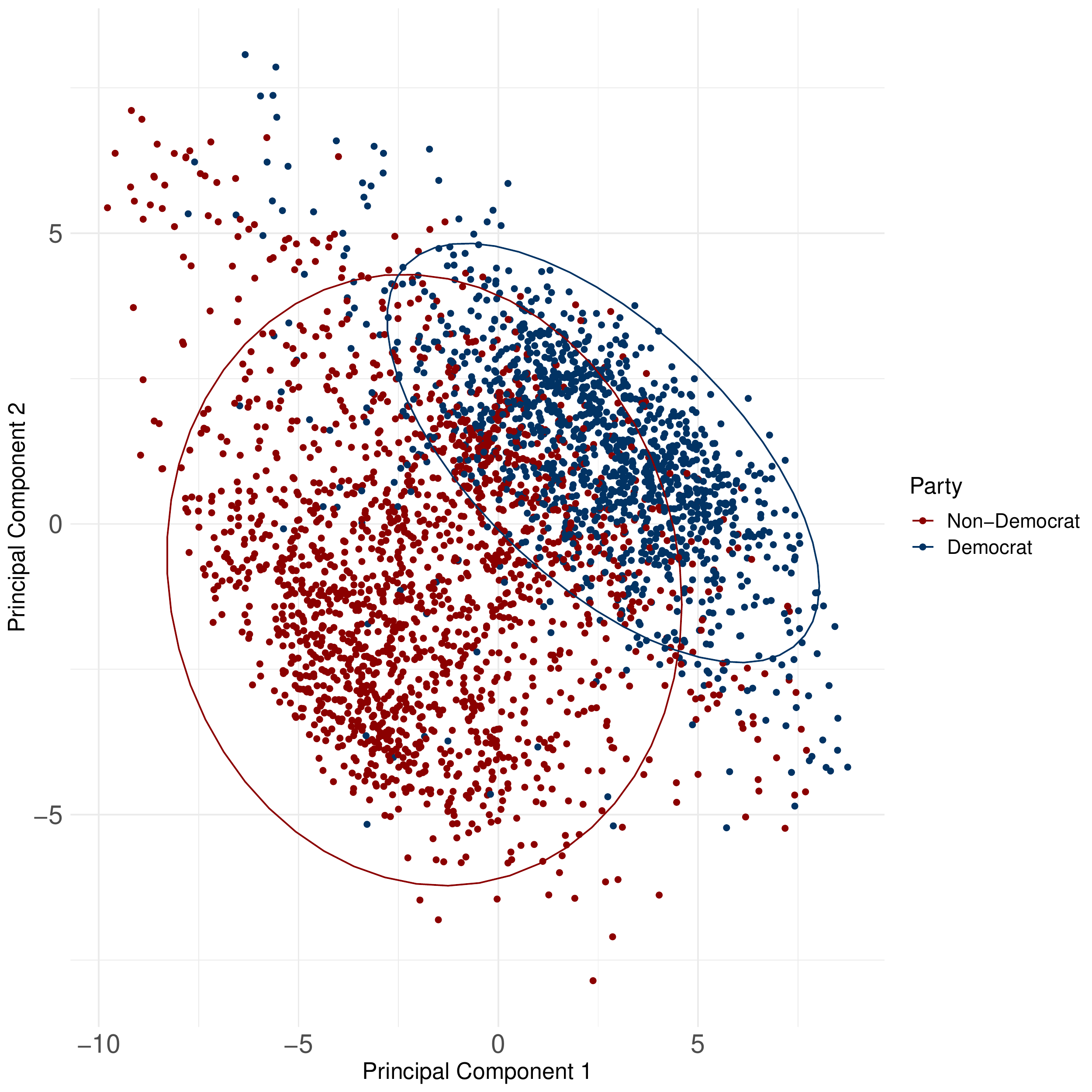}
	\caption{PCA Scores by Party Affiliation}
	\label{figure:scores}
\end{figure}

We will conclude the PCA section with a return to our guiding goal from the outset of the Element, where we are interested in exploring whether the latent structure in the feeling thermometer space varies along a partisan dimension. To do so, I generate a custom view of the solution along the first two dimensions by plotting the scores against each other and then coloring the points by party affiliation. Recall, the \texttt{pid7} party affiliation feature in the original ANES data included eight levels, where $1 = \text{Strong Democrat}$, $2 = \text{Moderate Democrat}$, $3 = \text{Lean Democrat}$, $4 = \text{Independent}$, $5 = \text{Lean Republican}$, $6 = \text{Moderate Republican}$, $7 = \text{Strong Republican}$, and $8 = \text{NA}$. To simplify the plots, I recoded this feature to be dichotomous, where $1 = \text{Democrat}$, and $0 = \text{Non-Democrat}$. The goal with this step is to avoid discarding data or information (e.g., dropping all non-Democrats or non-Republicans), while instead grouping those who identify at any level with the Democratic party ($\texttt{pid7} == 1:3$), or do not ($\texttt{pid7} == 4:7$). But ultimately, the chief benefit here is to clarify the visual patterns from the algorithmic output, which is a strategy adopted throughout the Element. As such, the PCA results with color according to party affiliation are shown in Figure \ref{figure:scores}. Ellipses are loosely drawn around each party group for descriptive value only.

In Figure \ref{figure:scores} a clear partisan pattern emerges based only on survey responses to the batter of feeling thermometers. And perhaps more strikingly, recall the first two dimensions from the PCA solution account for just over half of the PVE, 52.6\%. Thus, our ability to pick up on a clear partisan distinction based on only half of the variance in the data suggests that, while responses to these feeling thermometers are not overtly on partisan terms (i.e., ``as a Republican, rate your feelings on X''), there is a pronounced undercurrent of partisanship in these survey responses; a pattern we will reference back to throughout the remainder of the Element.

Perhaps the biggest weakness of PCA is the ambiguity surrounding the optimal number of reduced features to retain, as there is no formal guidance for making this choice. As a result, the benefits of dimension reduction via PCA become less clear. Related, in smaller dimensional contexts (e.g., 5 or 6), the value of PCA drastically diminishes as well. Recall, we can calculate up to $p$ principal components, such that the total variance can be explained by the PCA solution. When all of the variance is explained, we are back in the high dimensional context. Here again, the value of dimension reduction via PCA is substantially less clear. Thus, PCA is most useful when there is both high correlation across the full input space such that a reduced set of features contributes to the goal of simplifying and learning from data, and also when the dimensionality of the space is high.

To conclude, a great deal of space was dedicated to introducing and applying PCA, as well as diagnosing correlations across features in the original high dimensional space, because nearly every dimension reduction that has followed has built on PCA. Regardless of the scope of any post-PCA methodological innovation, PCA's approach to making high dimensional spaces more manageable and interpretable has not only impacting decades of development of subsequent dimension reduction techniques but has contributed to the rise of an entire subfield dedicated to dimension reduction. This subfield and its interpretational value in high dimensional contexts has never been more important than in the current era of massive data production at an incredible rate. 

\subsection{Suggestions for Further Reading} 

\citet{friedman2001elements} present excellent detail on PCA estimation in Chapter 14.5, along with discussion of extensions. \citet{zou2006sparse} offer an extension of PCA in high sparsity contexts, whereas \citet{scholkopf1997kernel} derive a \textit{nonlinear} version of PCA.

\clearpage

\section{Locally Linear Embedding} 

PCA is very valuable as a starting place for reducing complexity of some higher dimensional data space and learning an orthogonal representation of that data space based on maximizing variance across the input features. The result is a clearing of the clutter introduced by multicollinearity across the features. Yet, as \citet{goodfellow2016deep} point out, we must move beyond a simple linear combination of features if we want to more efficiently handle complex, high dimensional spaces. In short, we need PCA to \textit{understand} dimension reduction, while launching to construct more complex algorithms to help us \textit{handle} more complex data spaces.

Locally linear embedding (LLE), which has been around for about 20 years, is a linear dimension reduction technique like PCA \citep{roweis2000nonlinear}. Yet, LLE's focus on local structure and then global projection allows it to handle and accurately summarize more complex and \textit{nonlinear} data spaces. As PCA is interested in fitting multiple summary lines to data in orthogonal directions of greatest variance, LLE is interested in learning the \textit{shape} of the underlying data structure. This shape is called a \textit{manifold}, and LLE proceeds by linearly combining weighted features (like in PCA), but in a fundamentally different way as we will soon see. 

As such, I introduce several new concepts and themes to deepen coverage of dimension reduction: nonlinear dimension reduction, manifold learning, local vs. global approximation and representation. These naturally flow from the PCA approach, and also set up the more complex approaches covered in the following sections (e.g., we have to know what a manifold is if we want to use LLE, UMAP, or t-SNE).

\subsection{Manifolds and Complex Structure}

To effectively reduce the dimensionality of data, we need algorithms that, first, appreciate this reality of underlying data structure, and then, are flexible enough to capture it and then project it locally in a simpler subspace as we have noted to this point. A different way to think about data complexity is in terms of a manifold. A manifold is a $d$-dimensional geometric shape that is treated as locally Euclidean. Assuming a manifold underlies data, then, means that regardless of the contours of the manifold (that is, whether or not it is highly nonlinear), all observations are assumed to be exist somewhere along it, and can often be treated locally as linear. This unlocks the potential of applying many methods to learn the shape and contours of the manifold in a higher dimension, which can then be approximated and projected onto a simpler subspace. Though our goal remains the same, the key difference by introducing manifolds is that we are no longer interested in directions of variation across the features (i.e., data summaries), but rather we are interested in learning the full nuance of the structure along which the data are distributed, and then recreating that learned structure in a lower dimensional setting. 

Manifold learning is a central task in machine learning, and especially unsupervised learning. The notion of a latent manifold has been around for well over a hundred years, e.g., \citet{riemann1873hypotheses}. Indeed, Riemann's ideas have influenced not only the field of differential geometry but have given rise to a subfield known as ``Riemannian geometry." At a basic level, Riemannian manifold learning is premised on the idea that a latent, simpler manifold characterizes the input space, but in some high dimensional ambient space. \textit{Ambient} in this context means the possible space an observation could occupy is massive. Yet, that observation only occupies a single point in space. For example, members of Congress represent districts, cities, and states. So, at a strictly theoretical level, a legislator could be from the 35th district representing San Francisco, Louisiana. There is a district number, a city, and a state. The combinations are non-sensical in practice, of course, but this combination is technically possible in ambient space. And thus, the ambient space is comprised of a massive amount of \textit{possible} places to occupy in the high dimensional setting (35 in our running ANES example, where this silly example is only a 3-dimensional case). Yet, as each observation only occupies a single position across all dimensions in the higher dimensional, ambient space, the assumption is that all input data are distributed in this space and can be represented in a lower-dimensional way. The task of dimension reduction in these terms, then, is comprised of two parts: first, learn the contours and shape of the latent manifold (including distances between observations that exist in the common space), and then second, recreate the true structure in a lower dimensional, understandable way. In so doing, for these types of dimension reduction tasks, we are moving from some massive high dimensional ambient space with nearly infinite configurations across all dimensions, to a still-high dimensional, but more real version of the data (full, raw input space, $\mathbf{X}$), to ultimately a low dimensional subspace of two or possibly three dimensions. 

\begin{figure}[h!]
	\centering
	\parbox{7cm}{
		\includegraphics[width=7.65cm]{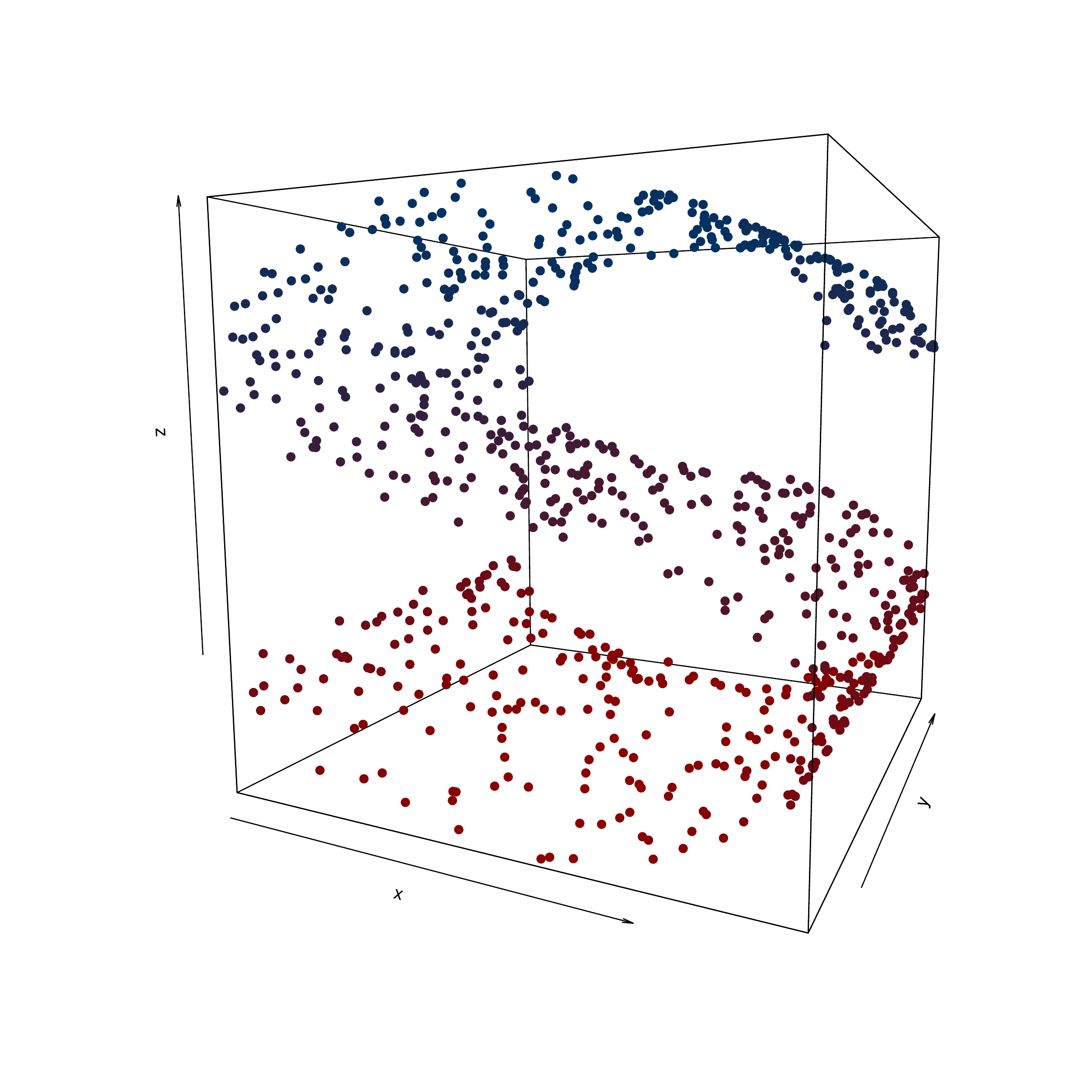}
		\caption{3D S-Curve}
		\label{figure:s}
	}
	\qquad
	\begin{minipage}{7cm}
		\includegraphics[width=7.65cm]{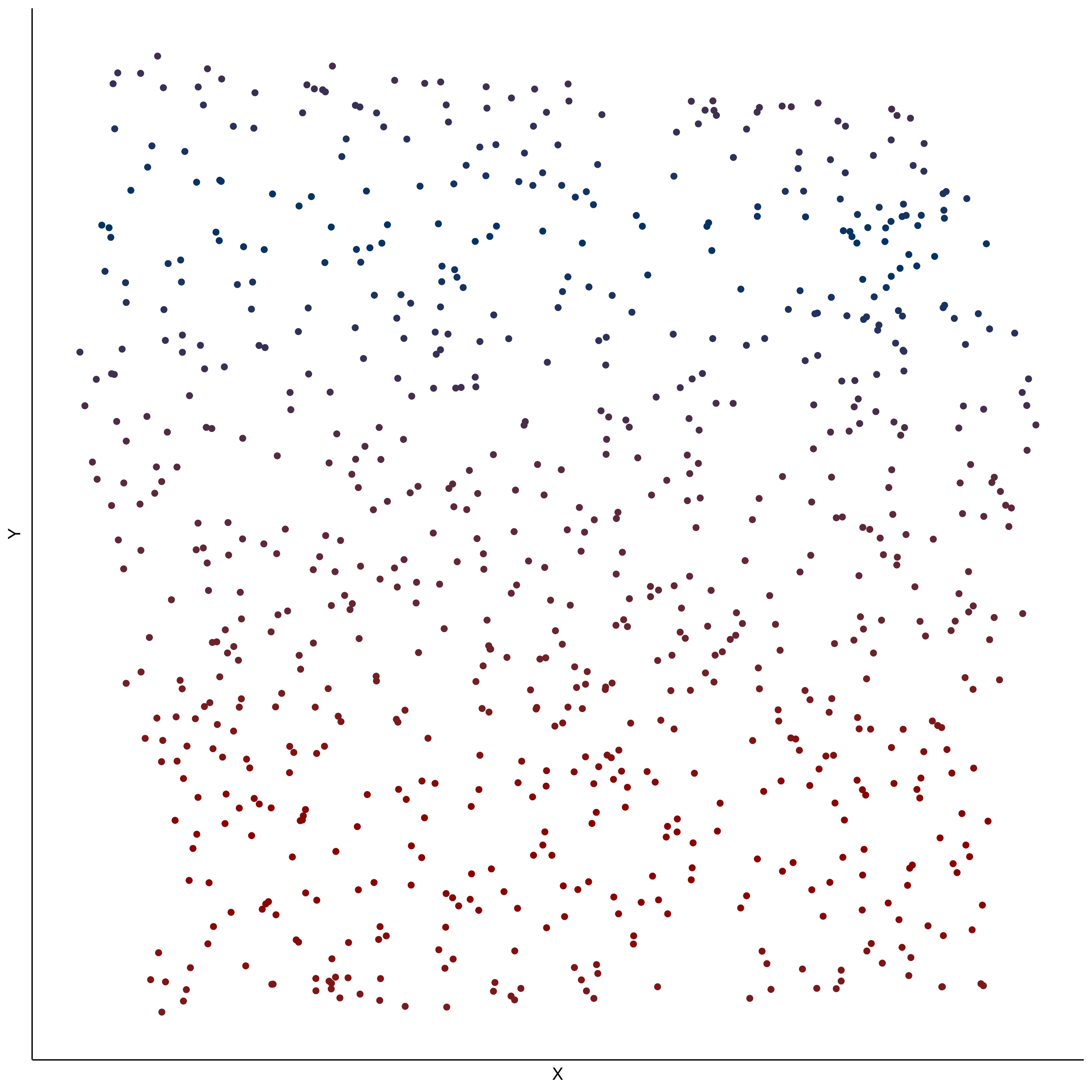}
		\caption{2D S-Curve Projection}
		\label{figure:sproj}
	\end{minipage}
\end{figure}

We can make this come alive a bit by demonstrating LLE's ability to unravel a three-dimensional ``S-curve'' of data, and project it to a two-dimensional scatterplot. We will use colors in the following illustration to help with understanding the location of points in the higher and lower dimensional versions. Of note, this application is possible and requires little code, when loading a few packages that contain the S-curve data. Start with the three-dimensional S-curve data in Figure \ref{figure:s}. To learn this structure and project it onto a lower dimensional subspace, we would be interested in a two-dimensional scatterplot of the data, but with the original structure remaining intact. This projection is shown in Figure \ref{figure:sproj}. 

In Figure \ref{figure:s} we can see the relatively complex S-curved structure. If we wanted to flatten this curve so to speak, and plot it in two dimensions on a scatterplot, a way to do so would be to first learn the distances between each point across all three dimensions. Recording these dimensions as new coordinates, we can then plot these coordinates in a two-dimensional sub (new) space. This is what LLE is doing at a high level, and indeed, LLE was used to flatten the S-curve in Figure \ref{figure:s} and then project the lower dimensional result in Figure \ref{figure:sproj}. 

Critical to understanding dimension reduction from a manifold setting perspective is the balance between local and global structure. Local structure refers to the pointwise approximations of a space. Combined across all observations in the data space, we then get a global picture of the true structure of the data, \textit{based on the local, pointwise approximations}. But importantly, as with a photograph, the projection is not a perfect mirror of the original high dimensional input space. There are errors that are made in attempting to take a 35-dimensional set of coordinates and project it down to a two-dimensional set. Yet, as with PCA, which was also based on ``making mistakes'' by throwing data away for the benefit of focusing on the majority of the variance across the features, we are willingly ``throwing data away'' in an implementation of LLE in an effort to learn the contours of the latent manifold. Once learned, we are then able to represent it in a way that is more intuitive and understandable.

\subsection{Formalizing LLE}

LLE attempts to reconstruct a data space based on a series of $n$ pointwise local calculations of similarity. \textit{Local} means proceeding across the input space on a point-by-point basis, and calculating an approximation of each observation $i \in \{1, \dots, n\}$, on the basis of a weighted, linear combination of a set of nearest neighbor observations. This approach to locally learning the data space as it is distributed along the manifold allows for the full space to be reconstructed, which results in the lower dimensional projection. Importantly, LLE is a \textit{linear} approach to \textit{nonlinear} dimension reduction. That is, the process of deriving the weighted local representations of each point, which are later combined to represent the global structure, are linearly combined (i.e., added together). Yet, the shape and structure of the manifold along which the data are distributed can be (and often is) nonlinear and complex, especially as the dimensionality of the space increases. The goal of LLE, then, is to find the weights in a local fashion, combine (multiply) them by the raw feature values, and then add them together to give an approximation of the single point. This process is repeated for all observations, up to $n$ times, resulting in the learned, lower dimensional representation of the high dimensional input space. 

To formalize LLE, start with the high dimensional input space, $\mathbf{X}$, which recall is an $N \times P$ data matrix, consisting of elements $i \in \{1, \dots, n\}$ across column features, $p \in \{1, \dots, P\}$. The output space, which given our focus in this Element, is not a predicted value like in a supervised task, but is another matrix, which we can call $\mathbf{Y}$ for simplicity. We let $\mathbf{Y}$ be an $n \times d$ matrix, consisting of values $i \in \{1, \dots, n\}$ across each dimension, $d \in \{1, \dots, D\}$. Importantly, as with PCA and the other techniques covered later, we will typically let $D = 2$, such that we are interesting in moving from the higher dimensional space, e.g., in our case $D = 35$, to a lower dimensional subspace, e.g., $D = 2$. This is a formal way of describing any two dimensional plot like the earlier PCA scores plot in Figure \ref{figure:scores}.

Recall we are interested in linearly combining \textit{weighted} with original feature values to produce an approximation of each point, ultimately done $n$ times. These weights, which are at the heart of LLE and will be stored in $\boldsymbol{\beta}$, are based on distances between a candidate observation, $i$, and a small surrounding neighborhood of observations, $j$, of size, $k$. Thus, the weights are subject to a normalizing factor for each observation, $i$, 

\begin{equation}
\sum_{j=1}^{k} \boldsymbol{\beta}_{ij} = 1.
\label{eq:lleweights}
\end{equation}

Regarding distance, most applications of LLE and other dimension reduction algorithms either use Euclidean distance or Manhattan (``city block'') distance. Euclidean distance, $d_{e}$, is found by taking the square root of the squared difference between observations $i$ and $j \forall i \neq j$. For example, in a single dimension, which is easily scalable to higher dimensions,

\begin{equation}
d_{e}(i,j) = \sqrt{(i-j)^2}.
\label{eq:euc}
\end{equation}

Manhattan distance, $d_{m}$, measures the distance between two points by taking the absolute value of the difference between two observations, $i$ and $j \forall i \neq j$. Again, in one dimension we have,

\begin{equation}
d_{m}(i,j) = |i-j|.
\label{eq:man}
\end{equation}

For our purposes, unless otherwise noted, whenever a measure of distance is required in the techniques covered in the Element, Euclidean distance will be used. 

Imagine a small neighborhood of observations surrounding each candidate observation, $i$. We are interested in calculating the distance between $i$ and each surrounding observation, $j$. We then multiply the distance, which itself is the ``weight,'' by the raw feature value for each feature across all points in the small neighborhood. Importantly, allowed by the constraint in Equation \ref{eq:lleweights}, we can more readily interpret the weights for each point, $j$, relative to $i$, such that higher values indicate the observation is closer to and thus looks more like the candidate observations, $i$. In this way, across $n$ points, the local structure is learned in a point-by-point fashion and is then able to be represented in a lower dimensional version of the space, as we are left with a set of $n \times d$ coordinates (stored in $\mathbf{Y}$) by which we can simply produce the $d$-dimensional plot. LLE accomplishes this by capturing and retaining all of the \textit{original} information and structure across the full data space (which again, is distributed along the latent manifold). By plotting $\mathbf{Y}$, we get the simpler, lower dimensional representation of the input space that has retained the original, high dimensional structure. This is how LLE can produce Figure \ref{figure:sproj} given only the complex input data in Figure \ref{figure:s}. 

Of note, LLE, as with many of the other techniques we will cover like autoencoders, is all about reconstruction. That is, there are many possible combinations of neighbors that \textit{could} surround $i$, and be used to approximate that point. Yet, the optimal LLE solution is defined by the configuration of all local approximations that minimize the reconstruction error. As reconstruction is based on the approximated value of $i$, based on the weighted sum of the neighborhood of $j$ of size $k$, this suggests we need to minimize the mistakes we make in making the pointwise calculations. Put differently, reconstruction error in the context of LLE is identical in structure to error in a simple linear regression setting. As in regression and LLE, our goal is to find the solution that minimizes error. Error in both LLE and regression settings is defined as information loss. By approximating the relationship between the inputs, $\mathbf{X}$ and the output, $\mathbf{Y}$, we are necessarily losing information for the sake of a well fitting, but parsimonious explanation of the data. To demonstrate this point, consider the similarity in the loss functions for regression in Equations \ref{eq:reg} compared to LLE in Equation \ref{eq:lle}. 

For regression problems, the most common approach to define loss is the mean squared error, 

\begin{equation}
MSE = \frac{1}{N} \sum_{i = 1}^{N}{\left(y_i - \hat{f}(x_i)\right)^2}.
\label{eq:reg}
\end{equation}

The goal in regression is to identify a model that generates the smallest possible MSE. That is, we want to minimize the error (mistakes in prediction) when we decide to approximate some relationship between in the inputs and output. 

The goal is the same in LLE. By substituting $y_i$ in Equation \ref{eq:reg} with the candidate observation, $x_i$, we are treating the candidate observation as the ``output'' for a single step in the algorithm. With a slight update to the notation, we get a measure of reconstruction error for a fit of LLE,

\begin{equation}
RSS = \frac{1}{N} \sum_{i = 1}^{N}{\left(x_i - \sum_{\forall j \in k \neq i} \beta_{ij}x_j\right)^2}.
\label{eq:lle}
\end{equation}

By using the set of weights found from minimizing RSS in Equation \ref{eq:lle}, we are able to find the set of two-dimensional coordinates, which ensures a minimal global reconstruction error, based on use of the optimal set of weights.

The optimal LLE solution, then, is found when RSS is smallest. By locally approximating each point in such a way, we can find the set of \textit{new} observations, $\mathbf{Y}$, which are derived on the basis of the weighted contributions of each feature, $p$, each point $j \neq i$, across all observations, $n$, distributed along the manifold. The result, when plotted, should look similar to all original candidate observations around which each neighboring observation was based, only now in two dimensions rather than 35. As such, LLE is able to capture global structure on the basis of local, pointwise approximation. 

\subsection{Applying LLE to the ANES Data}

As with the section on PCA, we will transition to apply LLE to the ANES 2019 pilot study data. We will be using the same input features covering feeling thermometers on a battery of issues, people, countries, and institutions. Prior to exploring the data, we first load the appropriate packages and scale the data to ensure all data are effectively unitless, and thus comparable on a common scale. 

\begin{figure}[h!]
	\centering
	\includegraphics[scale = 0.5]{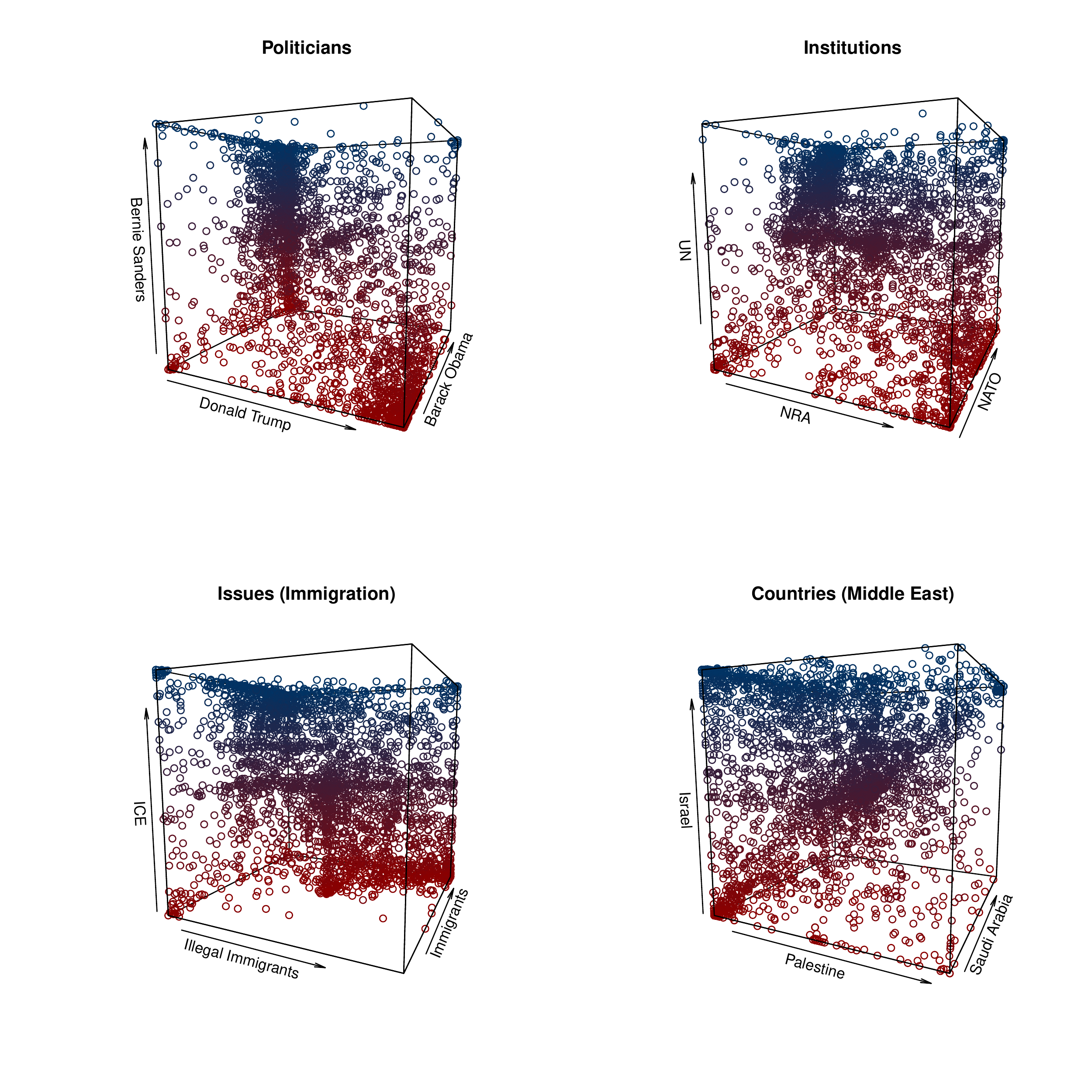}
	\caption{3D Distributions Across 12 Features}
	\label{figure:3d}
\end{figure}

With the data loaded, we will explore the data in a similar 3D way as in Figure \ref{figure:s}. We will explore three features at a time that are substantively related to each other to begin to get a sense (albeit an incomplete one) of the shape of the manifold. This is necessarily ``incomplete'' because a 35-dimensional plot would be substantively useless. See the results in Figure \ref{figure:3d}. 

The four plots in Figure \ref{figure:3d} show the distributions of three features unique to four substantive topics: Politicians (Bernie Sanders, Donald Trump, Barack Obama), Institutions (the UN, NATO, the NRA), the issue of immigration (ICE, illegal immigrants, immigrants in general), and middle eastern countries (Israel, Palestine, and Saudi Arabia). These plots in Figure \ref{figure:3d} reveal some useful patterns to inform our LLE application. For example, respondents tend to have extreme feelings toward the politicians with clear groupings of respondents at the bounds of the feeling thermometers on all three axes. 

We need to search for the optimal neighborhood size, $k$, prior to fitting the model. The main \texttt{lle} package includes a useful function, \texttt{calc\_k()}. This function runs LLE for each supplied value of $k$, which in the case below was a real-valued, non-negative integer from 1 to 20. The ``optimal'' value is defined at the value of $k$ for which the coefficient, $\rho$, is highest. $\rho$ is simply the correlation between approximations in the high dimensional setting compared to the low dimensional projection. Higher correlation means better representation. The function \texttt{calc\_k()} allows for plotting values of $1-\rho^2$, such that we are interested in the value of $k$ for which this quantity is lowest.

Using parallel computing on two different computers, it took 9.2 minutes run on 7 cores, and 10.9 minutes to run on 3 cores. Within the function call, we first pass the scaled data object (minus ``party affiliation''), followed by specifying the number of dimensions for the projection, $m = 2$. I then set \texttt{parallel = TRUE} and specified the number of \texttt{cpus} to use for the parallel process. Parallel programming of this sort is not necessary, but it significantly speeds up the code, especially for larger data applications. 

\begin{figure}[h!]
	\centering
	\includegraphics[scale = 0.5]{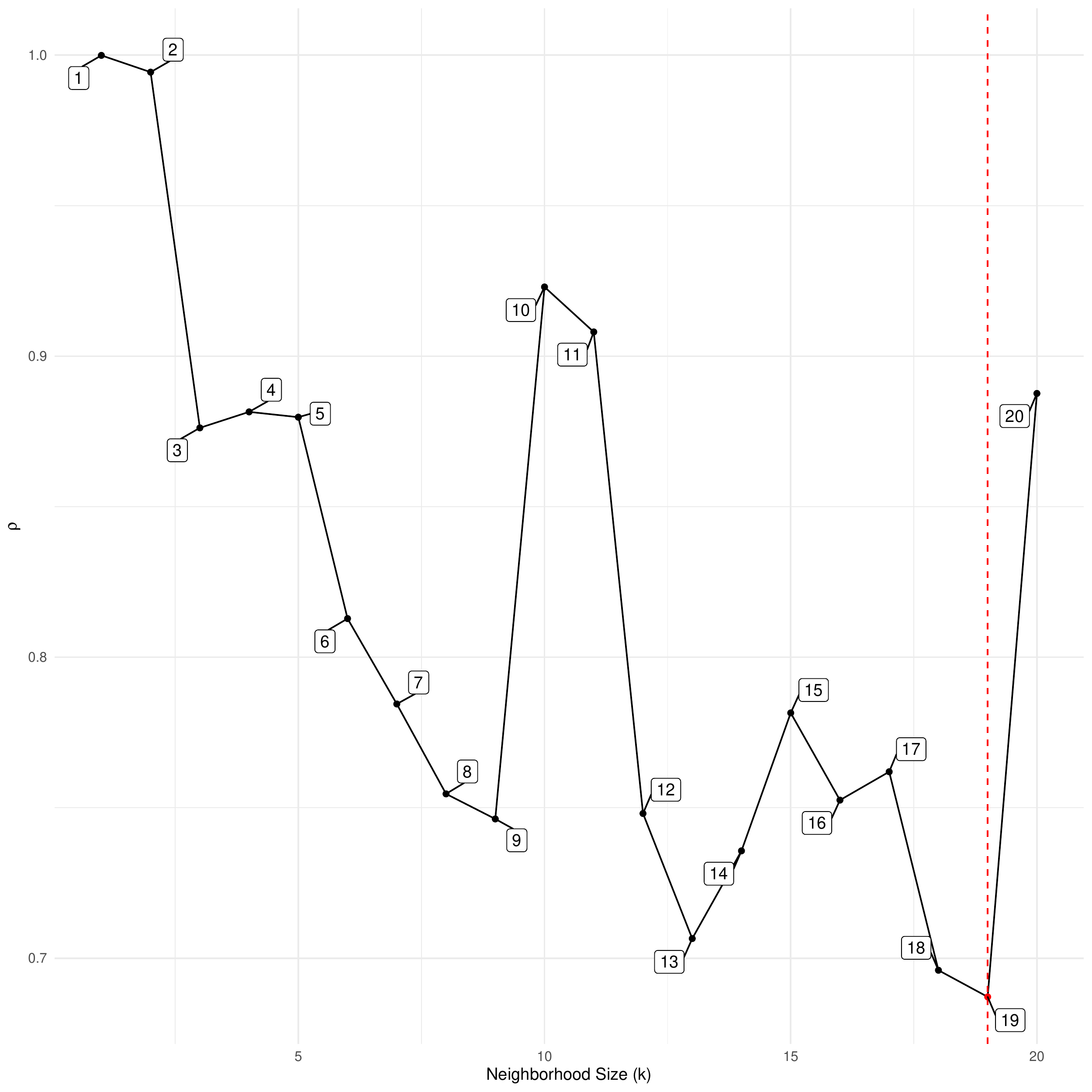}
	\caption{Optimal $k$ from Minimal Value for $1-\rho^2$}
	\label{figure:optk}
\end{figure}

\begin{figure}[h!]
	\centering
	\includegraphics[scale = 0.5]{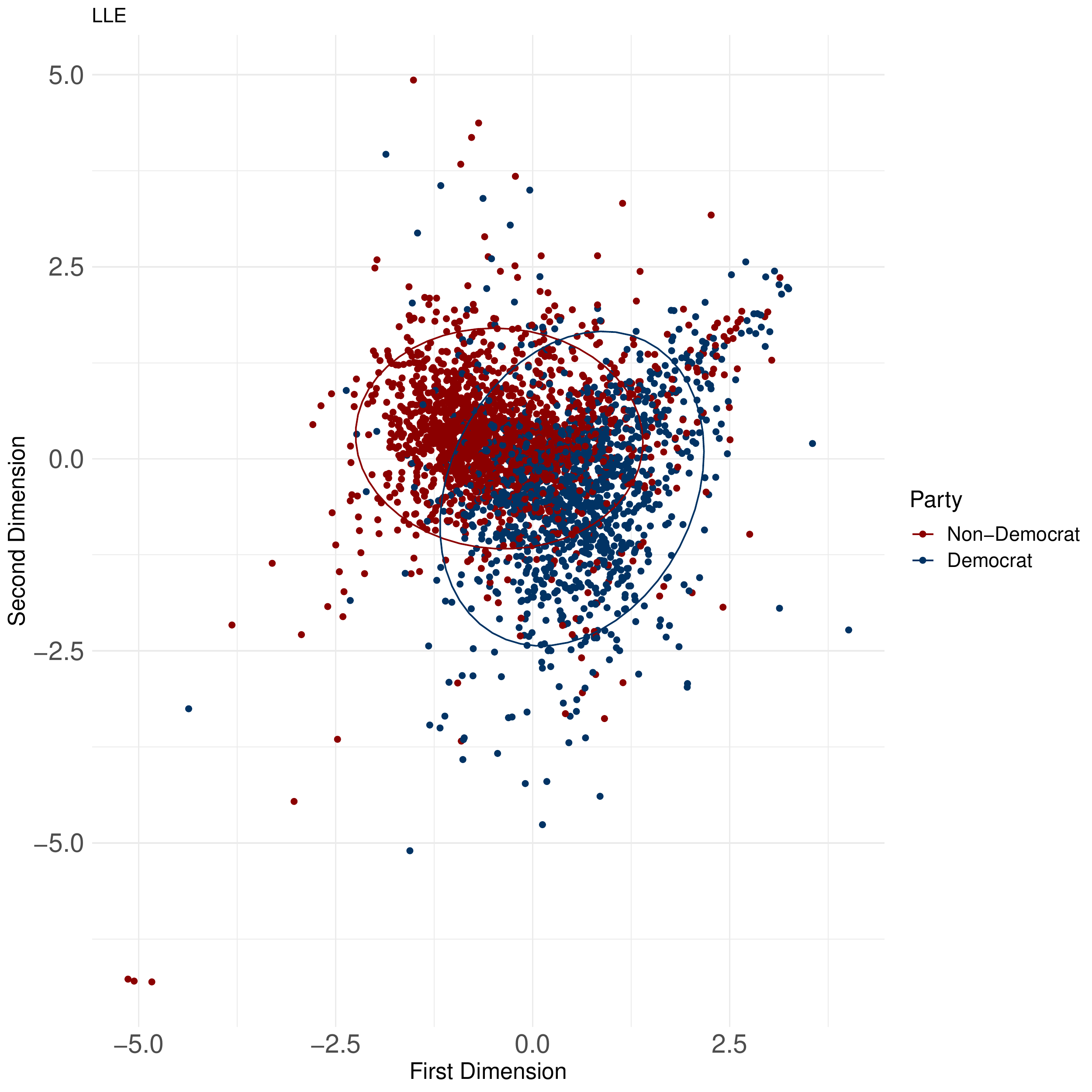}
	\caption{LLE Projection Results}
	\label{figure:lle}
\end{figure}

In addition to the code producing Figure \ref{figure:optk}, I include a brief line to manually locate the optimal value of $k$. Both the figure and the manual search reveal the optimal neighborhood size to give the best representation of the high dimensional space is 19. We proceed accordingly, setting $k = 19$. We follow the run of the algorithm with a visualization of the coordinates that mirrors the PCA approach. See the results in Figure \ref{figure:lle}.

\begin{figure}[h!]
	\centering
	\includegraphics[scale = 0.38]{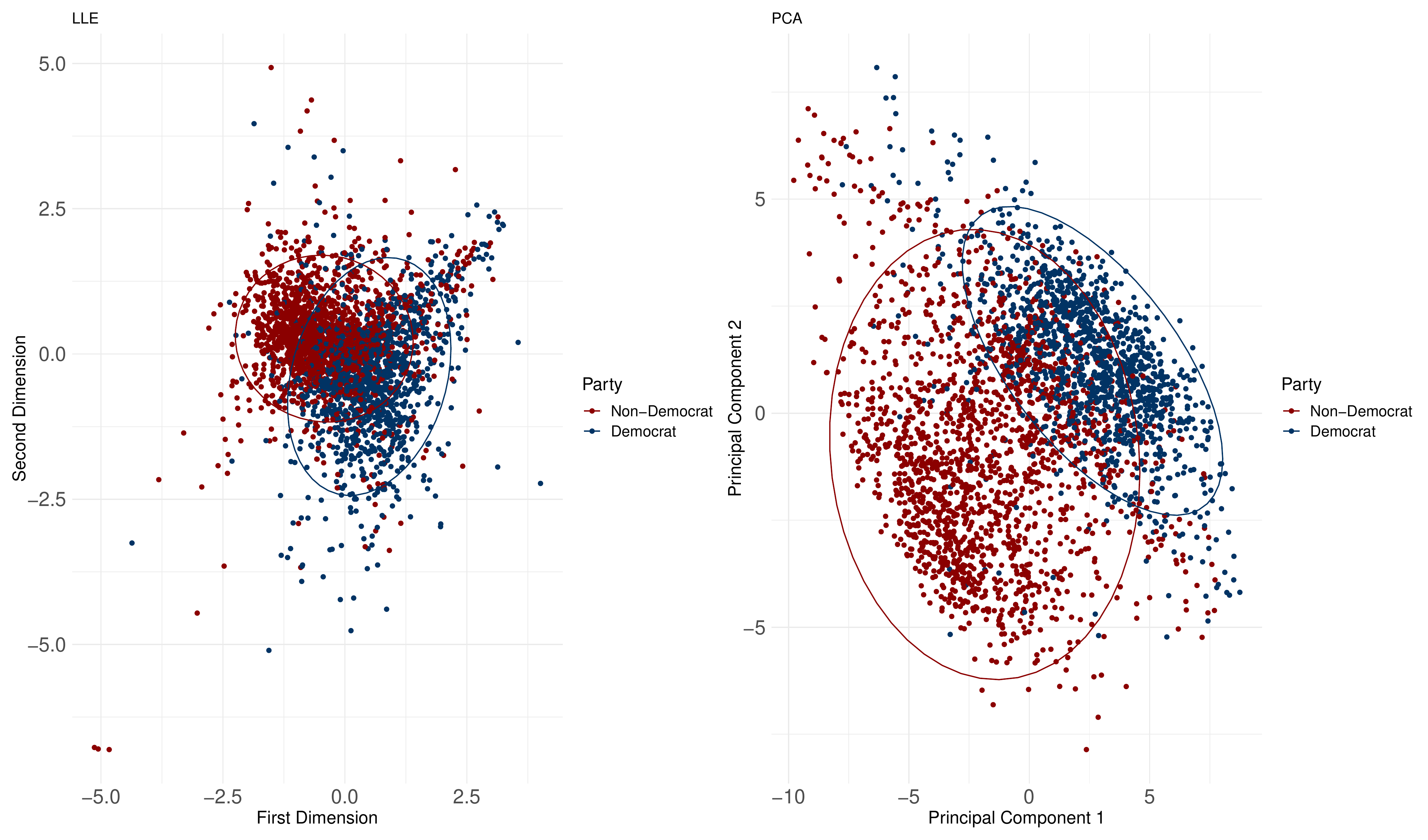}
	\caption{Comparing LLE to PCA}
	\label{figure:pcalle}
\end{figure}

From Figure \ref{figure:lle}, there indeed seems to be natural partisan differences across the first two dimensions in the projection space. Recall, the feature for party affiliation was not included in the LLE fit, allowing us to conditionally color points accordingly. Though differences on a party dimension seem to exist, the LLE results can be better contextualized by directly comparing with scores from a PCA fit. To do so and directly compare patterns, see Figure \ref{figure:pcalle}. 

First, in Figure \ref{figure:pcalle}, we can see similar distinction between Democrats (blue) and non-Democrats (red). Both algorithms are picking up on the partisan split in the feeling thermometer space. Interestingly, though, PCA results in a more diffuse projection, whereas the LLE results place respondents in a more compact space. As the LLE algorithm is capturing local structure and projecting the nuance of the distribution of points along the manifold, compared to PCA, which is plotting the most unique variance in the data along the first two dimensions, the LLE results are likely more accurately picking up the shape of the underlying structure of the data. Whereas the PCA solution is helpful to show the distribution of variance across the input space in the first two dimensions, the LLE solution is likely giving a more realistic look at the overall shape of the structure. This suggests, then, that the true distance between Democrats and Republicans, though distinctly separate, is not as diffuse as the PCA solution seems to suggest. The contours of this separation are addressed throughout the coverage of the other techniques in the Element. 

To recap, the goal of LLE is to create a simpler representation of the complex original space based on reconstruction error (and an attempt to minimize it). As such, the idea of reducing the original space with an eye toward recreating it on the basis of the simplified version is central to another approach to nonlinear dimension reduction called autoencoders. Autoencoders, which are built on a neural network-based architecture, are addressed later in this Element. 

Importantly, LLE helps address the limitations of PCA by focusing on recovery of the latent structure through both local \textit{and} global exploration of the data, rather than on simply maximizing shared variance as in PCA. Yet, a weakness of LLE is the linear combination of the weighted features, which can limit LLE's ability to recover global structure, especially in non-convex contexts for example. While LLE can handle nonlinear data better than PCA, it cannot do so as efficiently as some of the more recent, graph-based (t-SNE or UMAP) or neural network-based (SOM or autoencoders) techniques covered in the following sections. 

\subsection{A Risk-Averse Workflow}

To this point, we have covered PCA and LLE, along with several initial checks for multicollinearity across the feature space. As such, this final subsection offers a very biased opinion on what a risk-averse workflow for a dimension reduction project might look like given weakness of and connections between PCA and LLE.

First, adopt a demeanor of extreme caution when pursuing and interpreting output from dimension reduction algorithms and models. As these are unsupervised, it is impossible to \textit{conclude} that the patterns uncovered are indeed a good and reliable reflection of the high dimensional concept you are trying to capture. With that, though, I recommend starting with PCA as it is widely used, statistically well-established, and interpretation is quite intuitive. Upon generating the PCA solution, plot the scores as these are essentially the new measurements for the newly calculated features/components. These are the features that would be fed to a supervised or other model downstream, depending on the goals of the project. And to help deepen practical understanding of the PCA solution, consider coloring the scores in the two-dimensional plot along some theoretically interesting feature as in Figure \ref{figure:scores}. Then, with the single plot in hand, proceed to a more complex algorithm like LLE and compare the patterns as in Figure \ref{figure:pcalle}. 

\begin{figure}[h!]
	\centering
	\includegraphics[scale = 0.38]{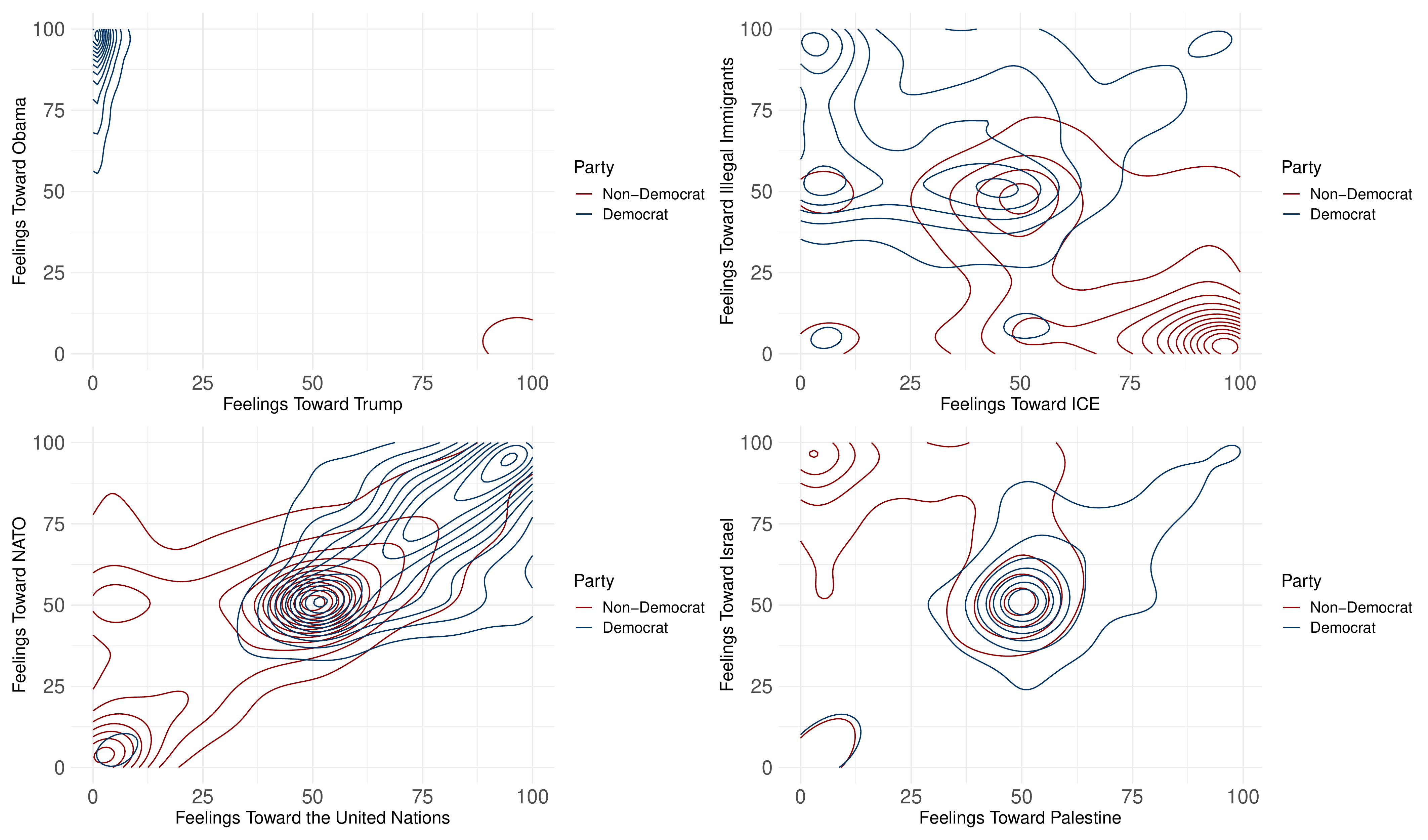}
	\caption{2D Contour Plots Across Eight Original Features}
	\label{figure:2dcontour}
\end{figure}

A final step that never hurts is to compare to the original, raw feature values. That is, plot several input features against each other and still color by party affiliation (or the key conditional feature) to explore whether natural separation in the high dimensional space is similar to the recovered, low dimensional separation. Importantly, as this is a manual, feature-by-feature approach, it is inefficient to progress across all input features, and indeed may not make sense to do for especially complex, high dimensional data sets (e.g., $p > 100$). Yet, for the sake of demonstration, we might plot a few key features against each other and color densities of observations by party to explore whether we find similar separation in the original space as we do in the transformed, reduced space as in the plots in Figure \ref{figure:pcalle}. To do so, and merely for demonstrative purposes, I plot the two-dimensional densities of observations across eight features, and color by party affiliation. They are feelings toward: Trump and Obama (upper left in Figure \ref{figure:2dcontour}), ICE and illegal immigrants (upper right in Figure \ref{figure:2dcontour}), the UN and NATO (lower left in Figure \ref{figure:2dcontour}), and Palestine and Israel (lower right in Figure \ref{figure:2dcontour}). See all of these results in Figure \ref{figure:2dcontour}.

In Figure \ref{figure:2dcontour}, we can certainly see separation in the raw, original feature space. For example, in the Obama/Trump plot in the upper left of Figure \ref{figure:2dcontour}, observations of each party are extremely far from each other as expected (i.e., feelings toward Trump are extremely polarized relative to feelings toward Obama). We see similar patterns for feelings toward ICE and illegal immigrants, and to a lesser degree in the feelings toward Palestine and Israel. Yet, perhaps most interesting is the spread of raw feelings toward NATO and the UN in the lower left plot of Figure \ref{figure:2dcontour}, which reveals a strikingly similar pattern of separation that we saw in the projection space from both PCA and LLE in Figure \ref{figure:pcalle}. There is a cluster of overlapping respondents in the middle of the plot, but large and clear clusters at the outer bounds across respondents of different parties. As a result, perhaps these features could be proxies for broad partisan divisions between respondents. 

This workflow based on comparison from several angles (e.g., across algorithms and across original raw features) helps with interpretation. Patterns from PCA, which recall are more interpretable than more complex manifold-based learners, can be directly compared and thus placed into context. If patterns from the more complex algorithm look widely different from the patterns in the simpler PCA approach, then this could be a signal that something is less clear than anticipated. And further, it could provide a useful clue to you, the researcher, that more careful thought is required to effectively learn the underlying structure of the data. If, on the other hand, similar patterns emerge, then this lends support for your expectations that, first, some non-random structure does indeed appear to characterize these data, and second, you have homed in on its contours and shape. That is, you have deepened an understanding of your data as you set out to do in this unsupervised, exploratory type work. Indeed, this is the general approach I take in this Element. I will continually compare substantive patterns with conditional coloring in my plots as a sanity check, of sorts, to ensure the patterns I am uncovering are likely non-random, and real.  

Ultimately, though, as referenced frequently to this point and elsewhere in this Element, we are engaging in unsupervised machine learning, such that \textit{no} ground truth is guiding the process. As a result, we are forced to proceed with our best judgement, which is by definition always ambiguous to at least some degree. The best we can do is compare across several approaches to a common problem using common data, and critically assess the emergent patterns to give insight into our data and motivating problem. 

\subsection{Suggestions for Further Reading} 

Building on the original LLE derivation \citep{roweis2000nonlinear}, \citet{saul2003think} present an explicit machine learning view of LLE, which is useful to understand more complex extensions including the robust version of LLE offered in  \citet{chang2006robust}.

\clearpage

\section{Nonlinear Dimension Reduction for Visualization} 

As we have seen to this point, we can use dimension reduction techniques to reduce the complexity of data, giving a simpler, more manageable version of the high dimensional input space. This is especially useful when the size and dimensionality of the data are quite large, which is less common in social science applications. Even still, as we have begun to see, the precise approach to reducing the dimensionality of a high dimensional space, can vary quite widely. We can consider a simpler version of a data space as one that reflects the most variance across the input features (PCA). Or we could consider a simpler version of a data space as one that reflect local and structural similarity, giving a lower dimension approximation or the original space (LLE).

Another way to think about how to select from among the many dimension reduction techniques is based in the \textit{goal} of the task or project. To this point, we have been implicitly interested in reducing the complexity of a data space to result in some reduced set of newly-calculated features. Though this process of \textit{feature extraction}, we proceed to plot these new features against each other and then condition the color of the observations by their party affiliation to offer an informal look at validating the quality of the dimension reduction solution. That is, \textit{did the technique separate members of different parties on the basis of a simplified version of the feeling thermometer space?} 

In line with the approach to this point of plotting results, is to more explicitly treat dimension reduction as a tool for visualization. That is, reducing the complexity of a high dimensional data space with an eye toward assessing the patterns visually, rather then an a priori focus on the new sets of calculations giving the new features as in PCA and LLE. Two extremely popular and widely used techniques for this goal of dimension reduction for visualization are: t-distributed stochastic neighbor embedding (t-SNE) and uniform manifold approximation and projection (UMAP).\footnote{Importantly, there exist other approaches to dimension reduction for visualization such as multidimensional scaling and ISOMAP. Though valuable, these are not covered in this Element due to limitations in space. For excellent coverage of multidimensional scaling in the social sciences, see \citet{armstrong2014analyzing}.} In this section, we will cover both t-SNE and the more recent UMAP approaches to dimension reduction focusing especially on the visual results of the algorithms to help us move closer to out goal of understanding whether latent partisan trends exist in the higher dimensional feeling thermometer space. 

\subsection{From Linear to Nonlinear Dimension Reduction}

Though the task focused on in this section is only slightly distinct from the previous sections and though all approaches have included heavy visual components, a key shift in covering t-SNE and UMAP is in how the algorithm treats the data to find the optimal lower dimensional projection. Recall, PCA and LLE were both \textit{linear} approaches to dimension reduction, finding the lower dimensional data space based on adding together a linear combinations of weighted raw feature values. With t-SNE and UMAP, though, we rely on a \textit{nonlinear} combination of raw input features to give a very different treatment of the data. We are still working with the same high dimensional data (35 to be precise), but we are now treating and processing it a bit differently than we have to this point.

Both t-SNE and UMAP rely on neighbor-based smoothing to give local versions of the high dimensional data. That is, we are interested no in working locally and in a neighbor-based fashion (as with LLE), but based only on distances within the neighborhoods, rather than using those distances to give a weighted version of the raw input features. Then, the manifold is attempted to be recreated based on these local distances. The distances are calculated such that observations far away are also pushed far away in the projected version of the full space, whereas observations that are close to the candidate observations in the high dimensional setting are brought very close to the candidate observation in the lower dimensional project. These repulsive and attractive forces are baked into the algorithm, which result in a lower dimensional version of the high dimensional space with spatial similarities \textit{and} differences being exaggerated. Spatially similar observations are grouped tightly together, whereas spatially different observations in the higher dimensional setting are reproduced to be pushed far from each other. Importantly, t-SNE is a probabilistic approach to this problem whereas UMAP is a graph-based approach that relies only on smoothed distances metrics. This innovation in UMAP overcomes the core limitations in t-SNE, which is addressed in the remainder of this section. 

The result from these approaches to dimension reduction is a (usually) two-dimensional plot of the lower dimensional version of the data space that looks, at first glance, quite odd given these exaggerated differences between observations on the basis of spatial similarity and difference. Yet, with the ability to tune the key hyperparameters in each algorithm to control the global versus local behavior of the algorithm (i.e., the amount of exaggeration), these algorithms can give incredibly important insight into the underlying structure of high dimensional data.

\subsection{t-SNE}

t-SNE is an algorithm that was developed in 2008 in an effort to make clear the underlying structure in data in a visually digestible way. With an expressed focus on developing and using the algorithm for visualization, \citet{maaten2008visualizing} center their approach on comparing probability distributions. The comparison between the original high dimensional version of the data is compared to the new lower dimensional representation using the common distribution-comparison metric, Kullback-Leibler (KL) divergence. When values are small, this means the distributions are extremely similar, and when KL-divergence is high, then distributions are different from each other. But first, what are these distributions being compared? 

The first step of t-SNE is to measure the distances between each candidate observation, $i$, and the surrounding neighbors, $j$ in the neighborhood the size of $k$. These distances are then placed under a normal distribution in order to convert them to probabilities, where higher probabilities mean an observation, $j$, is close to observation $i$. To make these more interpretable, $d(\cdot) \forall j \in k$ for each $i$ are scaled to sum to $1$ and can be stored in $\mathbf{Z}$, where $d(\cdot)$ is a measure of distance between some $j$ and the candidate $i$, 

\begin{equation}
\sum_{j=1 \forall j \neq i}^{k} \mathbf{Z}_{ij} = 1.
\label{eq:tsnedist}
\end{equation}

The set of coordinates capturing the probabilities between each observation in the high dimensional setting is stored in $x_h$, which belongs to the full probability space $\mathcal{X}$. 

Then, the next step is to create a new, \textit{randomly}-distributed low dimensional version of the data set in (usually) two dimensions. That is, the algorithm randomly plots the data in a lower dimensional setting, where $d = 2$. Then, the process of placing a distribution around each observation, $i$, and the calculating probabilities of observations lying close to each other is repeated. Yet, this time, the t-distribution is used to account for greater uncertainty given the force of information loss by placing the similarities into a lower dimensional setting, i.e., moving from $d = 35 \rightarrow d = 2$. With a new set of probabilities between all observations and each other in hand, the moving of observations continues at each iteration of the algorithm, continuing to move observations closer together that have increasingly probability of being close to each other, \textit{and} also moving farther apart from observations that have a lower probability of being similar to each other. We are left with a lower dimensional version of the data space $x_l \in \mathcal{X}$. 

Finally, we come to the distribution comparison. At each iteration (the movement of data points to place similar observations closer and dissimilar observations farther apart), the algorithm compares the matrix of probabilities between all observations in the lower dimensional space, $x_l$, to the original matrix of probabilities in the original, higher dimensional space, $x_h$. The measure of cost that captures differences between these versions is the KL-divergence metric, which is sometimes called relative cross-entropy,

\begin{equation}
\sum_{x \in \mathcal{X}} x_h \text{log}(\frac{x_h}{x_l}),
\label{eq:kl}
\end{equation}

where each probability distribution $x_*$ in the full probability data space, $\mathcal{X}$, is compared by logged, relative difference between the high dimensional version, $x_h$ and the low dimensional version, $x_l$. 

The idea here is that as probabilities in the lower dimensional setting look more like probabilities across observations in the high dimensional setting, then the KL metric decreases, suggesting less dissimilarity in the data space. When the metric is low and unchanging, the algorithm stops, suggesting we have arrived at a lower dimensional version of the high dimensional data, that has a structure that mirrors the original complex structure. And to reiterate, we are no longer working with weighted versions of the original input features. Rather, we are now attempting to capture the probabilities that observations lie close to some in space and far from others in the same space. Basing these measures of similarity and differences on measures of spatial similarity, instead of linearly adding individual feature contributions together, we are bypassing the additive component of LLE and PCA's approach to dimension reduction. This is what is meant by ``nonlinear dimension reduction'' in the case of t-SNE.

Thus, the goal of t-SNE is to reproduce the probabilistic (via spatial) similarities across all points in the high dimensional context (again, based on local probabilistic similarities), in a lower dimensional (usually 2) setting, such that clearly similar groups of observations are tightly connected and are distant from clearly different groups of observations. 

There are multiple hyperparameters that need to be tuned, and thus globally applied to the model to give an optimal solution. The most important hyperparameter is \textit{perplexity}. Perplexity controls the tradeoff between focus on reproducing global structure versus local structure. Ideally the goal is to effectively balance between these, thereby finding a good lower dimensional representation of the more complex higher dimensional input space. Perplexity controls this tradeoff by varying the size of the distribution placed around each point. The higher the perplexity value, and thus the wider the distribution around each point, the more global the solution will look, as the distribution will allow point potentially far off in space to be given a higher similarity/probability of being close to the center of the distribution, which is where the candidate observation, $i$ is. So, by increasing the size of perplexity and thus the size of the pointwise distributions, we are allowing for higher a greater probability that distant points are treated as more similar. Inversely, then, lower perplexity gives much narrower distributions around each point, resulting in a lower probability that distant points will be treated as similar to the candidate observation, $i$, at the center of the distribution. Thus, lower perplexity scores result in recovery of more local structure. The question becomes, how do we home in on the optimal perplexity value, given the potentially drastic change in the lower dimensional representation we get as a result? There are two options here: manually vary the value and inspect distributions or conduct a grid search across multiple parameter values to inspect the evolution of structure over the range of the hyperparameter. 

\begin{figure}[h!]
	\centering
	\includegraphics[scale = 0.38]{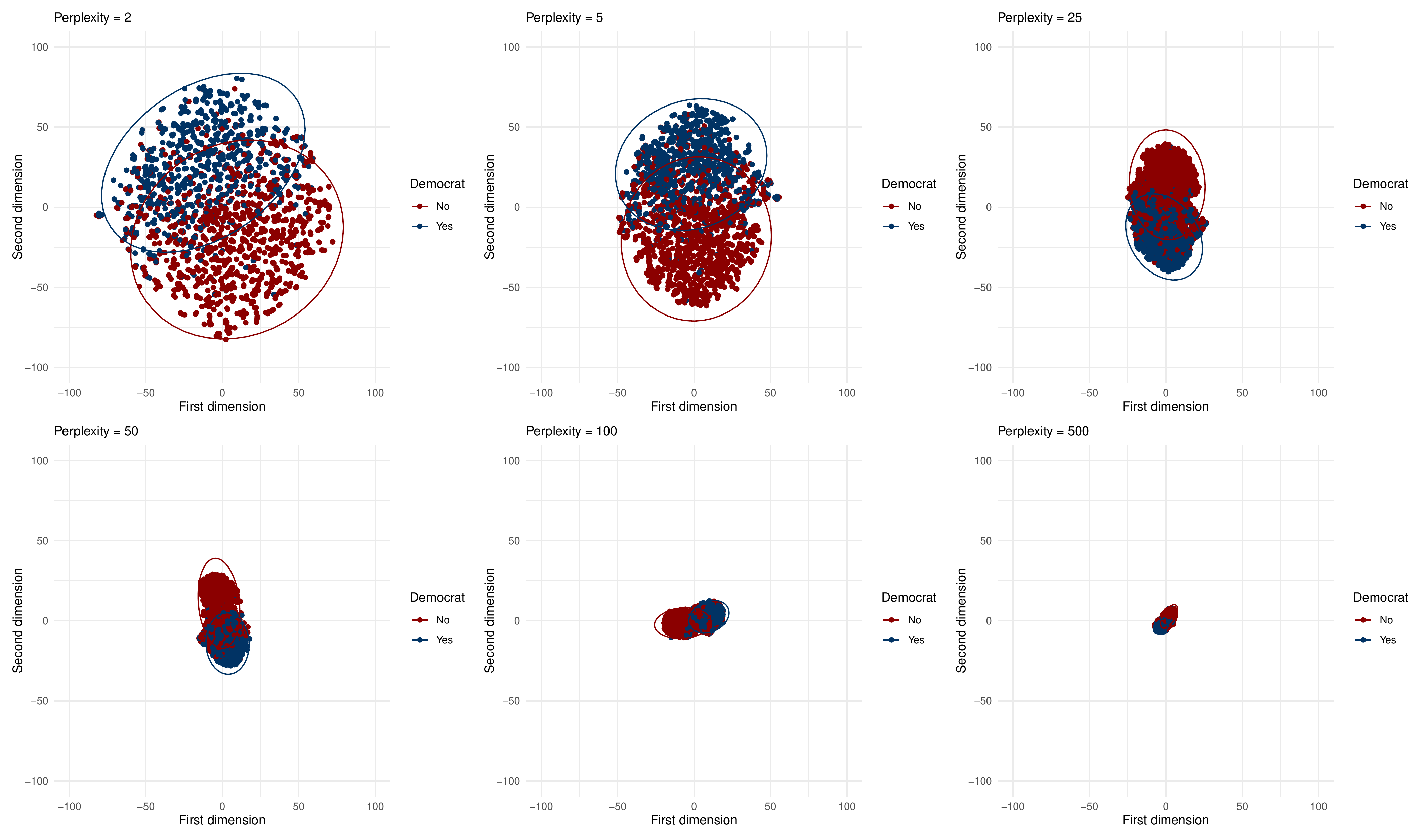}
	\caption{Manually Varying Perplexity for t-SNE}
	\label{figure:tsne1}
\end{figure}

Consider the first approach of manually varying perplexity. For the application of t-SNE using the ANES data, we will rely on the \texttt{Rtsne} package. As before, we will be working with the ANES clean data object and several tidyverse tools (e.g., \texttt{ggplot2}, \texttt{dplyr}). Upon manually varying perplexity, we can \texttt{patchwork} the six plots together to result in Figure \ref{figure:tsne1}.

Note the clear change in the picture of structure we get in Figure \ref{figure:tsne1}, when perplexity is varied manually. As expected, moving from small values of perplexity we pick up on more local nuance in the structure, compared to high perplexity values, where the global structure is the focus. In the case of the latter, e.g., perplexity $= 500$ as the extreme version in the lower right plot, the space is extremely dense, suggesting ``structural similarity,'' essentially defined by the fact that the respondents were simply responding to the same survey. In other words, such extremely high perplexity yields a less-then-useful look at this space, where we are unable to detect any degree of separation in the projection space. 

Regardless of the different values of perplexity, a clear pattern is the separation between Democrats and Non-Democrats in this projection space. Though the separation naturally lessens when we tune the algorithm to focus more on global structure, the parties are never fully mixed together. This suggests that there is indeed a latent partisan structure to this feeling thermometer space. 

The \texttt{Rtsne} package is built on a subtly updated version of t-SNE using the Barnes-Hut approximation. Though details of this extension are beyond the scope of current purposes, I point interested readers to the details in \citet{van2014accelerating}. In brief, this implementation of t-SNE includes an additional hyperparameter, $\theta$ to capture the tradeoff between \textit{speed} of fitting of the algorithm with the \textit{accuracy} of the solution. Therefore, to more efficiently search across both of these hyperparameters, the code below briefly demonstrates that which such a grid search might look. Importantly, though limited by space, the following code can be extended in numerous ways, e.g., comparing across other hyperparameters such as \textit{$\eta$}, which captures the learning rate or size of steps to take between iterations. But sticking with perplexity and $\theta$, for which zero gives the original t-SNE implementation and thus the most accurate, but slowest solution, whereas $\theta = 1$ gives the fastest, but less accurate solution. Of note, as we will see in the follow subsection, UMAP is a much faster algorithm being run in a fraction of the time as t-SNE. Indeed, the t-SNE grid search took about 19 minutes to run on 4 cores. 

\begin{figure}[h!]
	\centering
	\includegraphics[scale = 0.38]{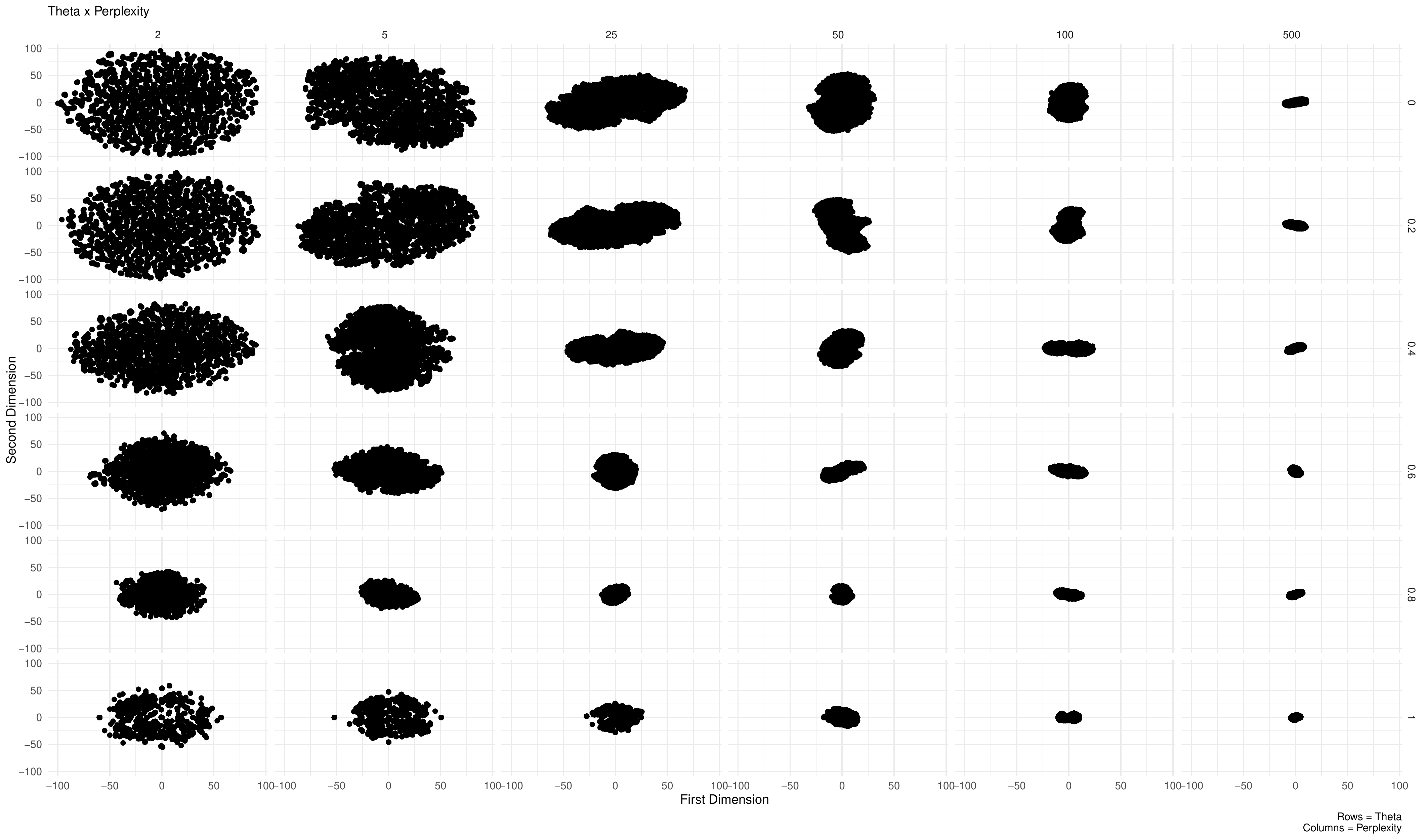}
	\caption{t-SNE Grid Search Across $\theta$ and Perplexity}
	\label{figure:tsne2}
\end{figure}

From Figure \ref{figure:tsne2}, there is a clear shift in the structure across various hyperparameter values. For example, as previously seen in Figure \ref{figure:tsne1}, higher perplexity values reveal a more global version of  the lower dimensional projection, whereas smaller perplexity values in the left-hand plots of Figure \ref{figure:tsne2} pick up on more local, diffuse structure. Yet, interestingly, not much nuance appears for either lower or higher $\theta$ values, across the range of perplexity. The clear shift in structure is seen in the columns as perplexity is varied. The goal of this demonstration is offering a more efficient approach to varying multiple hyperparameters in a single chunk of code, instead of manually varying values and checking patterns after each. Especially for a slow algorithm like t-SNE, the manual approach is much less efficient.  

\subsection{UMAP}

At this point, we come to a close relative of t-SNE: UMAP \citep{mcinnes2018umap}. UMAP is a very recent approach to nonlinear dimension reduction and is also primarily used for high dimensional problems and visualization. In brief, UMAP is also interested in proceeding locally to learn the structure in the original high dimensional space and then project it in a lower dimensional subspace. Yet, a key difference between UMAP and t-SNE is the ability to reproduce a projection solution. Whereas t-SNE is interested in assuming similarities equate to probabilities based on the overlaying of t-distributions in the lower dimensional projection process, UMAP avoids the notion of probabilistic similarities altogether and instead relies on smooth geometric distances. More firmly rooted in mathematical theory and specifically manifold learning, UMAP seeks to first learn the shape and structure of the high dimensional manifold, and then project it locally. Though the goal was similar to the goal in t-SNE, UMAP updates the manifold learning process by searching two regions in the higher dimensional space: between points, and then around regions of points. 

UMAP essentially conducts two searches in order to first learn the high dimensional structure and then be able to project it locally. First, the algorithm searches for local connection in a pointwise fashion across the full data space. As observations exist within this search region, they are connected to each other. Then, the second search is in the ambient space, previously referenced in the LLE section. This search involves taking the nerve of the now-learned higher dimensional manifold. Practically, this means the algorithm places a fuzzy secondary search region around each point, such that densely populated regions have a smaller sized neighborhood (defined by the radius of the region), compared to less densely populated regions along the manifold, which have a larger radius around them to ensure the manifold is connected, at least locally. This assumption of at least local connection is a core assumption in UMAP, and one that is potentially problematic, as perhaps not all observations truly belong to the same manifold. Still, the current implementation of UMAP defends this mathematically in \citet{mcinnes2018umap}, so we continue assuming it is true. The goal with these search regions, though, is to balance between local and global structure. Formally, the clearest look at UMAP's update of t-SNE is in the cost function. UMAP's cost function updates Equation \ref{eq:kl},

\begin{equation}
\sum_{x \in \mathcal{X}} x_h \text{log}(\frac{x_h}{x_l}) (1-x_h) \text{log}(\frac{1-x_h}{1-x_l}). 
\label{eq:umap}
\end{equation}

Now, the first term, $x_h \text{log}(\frac{x_h}{x_l})$ which was also in the calculation of cost in t-SNE, measures the local structure, the addition of the second term, $(1-x_h) \text{log}(\frac{1-x_h}{1-x_l})$, in concerned with the learning the full shape of the manifold. 

In addition to updating the cost function, we also update the definitions, where we are no longer interested in calculating conditional probabilities for $x_h$ and $x_l$. But instead, we are interested in calculating $n$ pointwise similarity scores based on smoothed spatial proximity to each other \cite{mcinnes2018umap}, e.g.,

\begin{equation}
x_{j | i} = exp[(-d(i,j)-\rho_i)/\sigma_i],
\end{equation}

where $\rho$ and $\sigma$ are hyperparameters that control the smoothing process. Thus, as with t-SNE, there are several hyperparameters to tune, which apply globally to the model. For example, others include $k$, which is the size of the neighborhood in the first search region, epochs, which is the number of times the algorithm sees the data, and also distance, which is a distance metric (e.g., Euclidean). 

To demonstrate the tradeoff in learned high dimensional structure and its impact on the local projection of that structure, consider the following demonstration holding the neighborhood size small and fixed at 5, and letting the times the algorithm sees the data (\textit{epochs}) vary to be relatively few times at 20 to many at 500. The intuition is the true structure is more likely to be learned when the number of epochs is high. The code for this demonstration relies on the \texttt{umap} package to implement UMAP with the ANES data. Indeed, there are other packages to fit and diagnose a UMAP model, one of which is covered at the end of this Section.

\begin{figure}[h!]
	\centering
	\includegraphics[scale = 0.38]{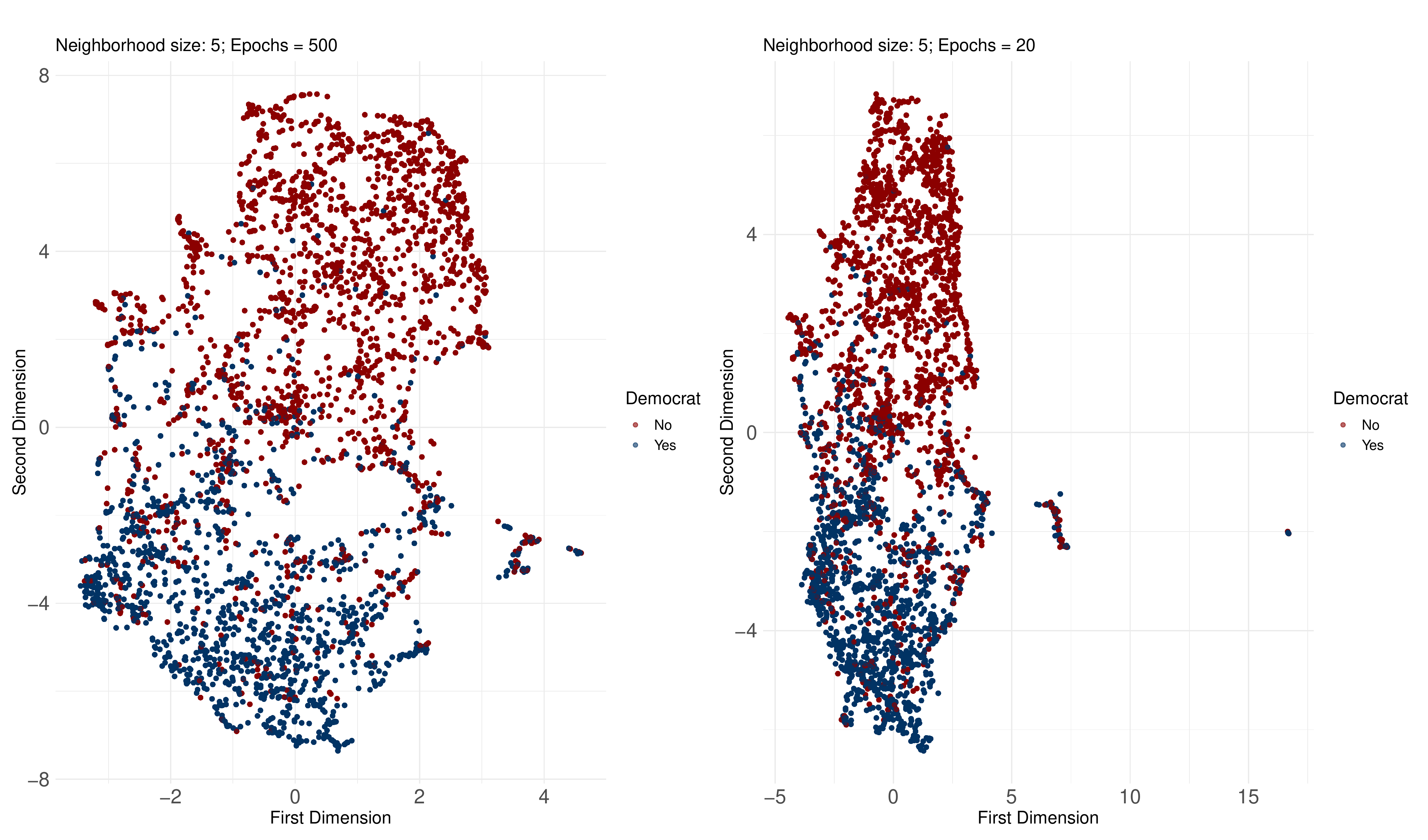}
	\caption{Comparing UMAP Fits Across Epochs: 500 vs. 20, with $k = 5$}
	\label{figure:umap1}
\end{figure}

\begin{figure}[h!]
	\centering
	\includegraphics[scale = 0.38]{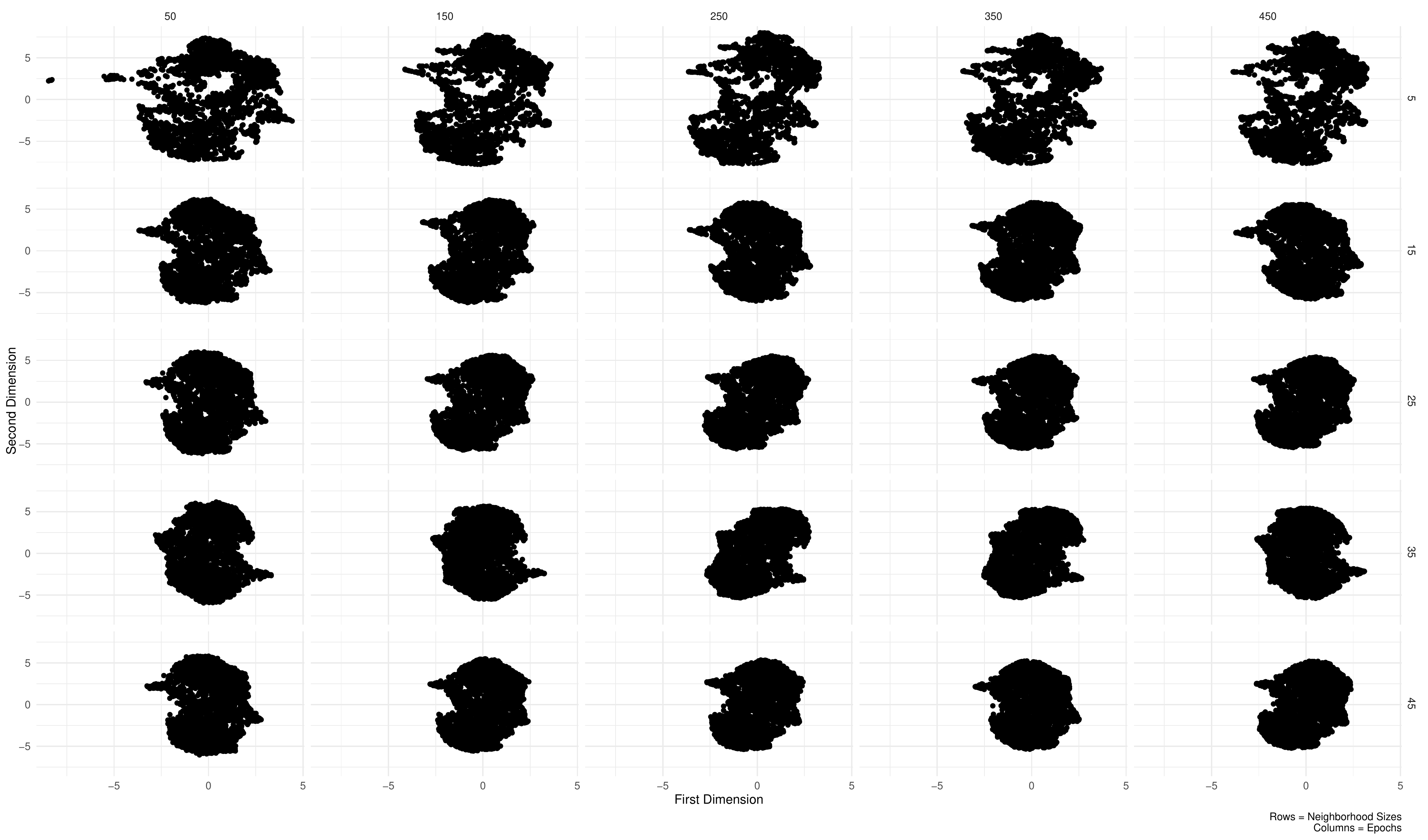}
	\caption{UMAP Grid Search Across $k$ and Epochs}
	\label{figure:umap2}
\end{figure}

From Figure \ref{figure:umap1}, we are again able to see clear segmenting on a partisan basis. But for the purposes of comparing the number of epochs, note the strange ``islands'' of observations to the right of the main cloud of data points in both plots, but especially the right plot in Figure \ref{figure:umap1}. Likely, the reason for this strange pattern is that that algorithm was not given enough time to learn the true structure of the manifold, such that it had a harder time of representing it in lower dimensional space. To verify this, and similar to the grid search exercise with t-SNE, we can update the code to conduct a grid search across several values of $k$ and the number of epochs. I essentially update the t-SNE grid search code, and store results after several fits of UMAP. Then, results are plotted in Figure \ref{figure:umap2}, with values of $k$ and columns showing varying number of epochs. 

\clearpage

Of note, the UMAP grid search took only about 5.5 minutes to complete, which is about 3.5 times faster than the t-SNE algorithm on the same data. This marked increase in computational efficiency is one of the biggest improvements over t-SNE. 

The results in Figure \ref{figure:umap2} confirm our expectations that the strange island of observations off to the left originally in Figure \ref{figure:umap1} are indeed due to insufficient time for the algorithm to learn the true structure of the latent manifold. We see those points slowly migrate over to the full cloud of observations as the number of epochs increases (i.e., moving from top to bottom) across Figure \ref{figure:umap2}. Given these patterns, though no formal rule exists for selecting optimal hyperparameter values, I recommend adjusting values until patterns stabilize. Too much tuning and we are in threat of overfitting to the training/sampled data. Yet, too little tuning, and we run the risk of learning or picking up on noise in the data. The trick is to find the sweet spot across hyperparameter values, such as when the islands in Figure \ref{figure:umap1} stop shifting as the number of epochs increases for example. Thus, the grid search is immensely helpful in this regard by providing several visualizations in a single plot to be able to observe shifts back-to-back across different hyperparameter values.

In sum, the results across \textit{all} techniques substantively suggest that members of all parties, while not extremely separated in the projection space, are indeed consistently distinct from each other, with minimal overlap between Democratic respondents compared to Non-Democrats.

\subsection{A Tidy Tangent on UMAP via \texttt{uwot::umap()} and \texttt{embed}}

For the sake of thoroughness, consider a very different, but tidy approach to UMAP. To do so, I use the \texttt{tidymodels} package, which requires use of two other packages: \texttt{uwot} and \texttt{embed}, instead of the formerlly used \texttt{umap package}. Here, I replicate the initial UMAP fit in the left plot of Figure \ref{figure:umap1}, where $k = 5$ and the number of epochs $= 500$. 

The main value, and thus justification of this tidy tangent, is to demonstrate that different packages and approaches can be used for the same task. This may come in handy when certain programming grammar is more comfortable to certain programmers, compared to others. In this case, I use the tidyverse approach, syntax, and packages (with the exception of \texttt{uwot}). As noted at the outset of the Element, tidy programming is built on the principle of adaptable code meant first and foremost to be readable by humans. Related is the ability to pipe/stack multiple functions, even plotting functions, which results in a single chunk of code to prepare data, fit a model, and visualize the results. Such an approach is ultimately simpler and much more efficient. 

\begin{figure}[h!]
	\centering
	\includegraphics[scale = 0.5]{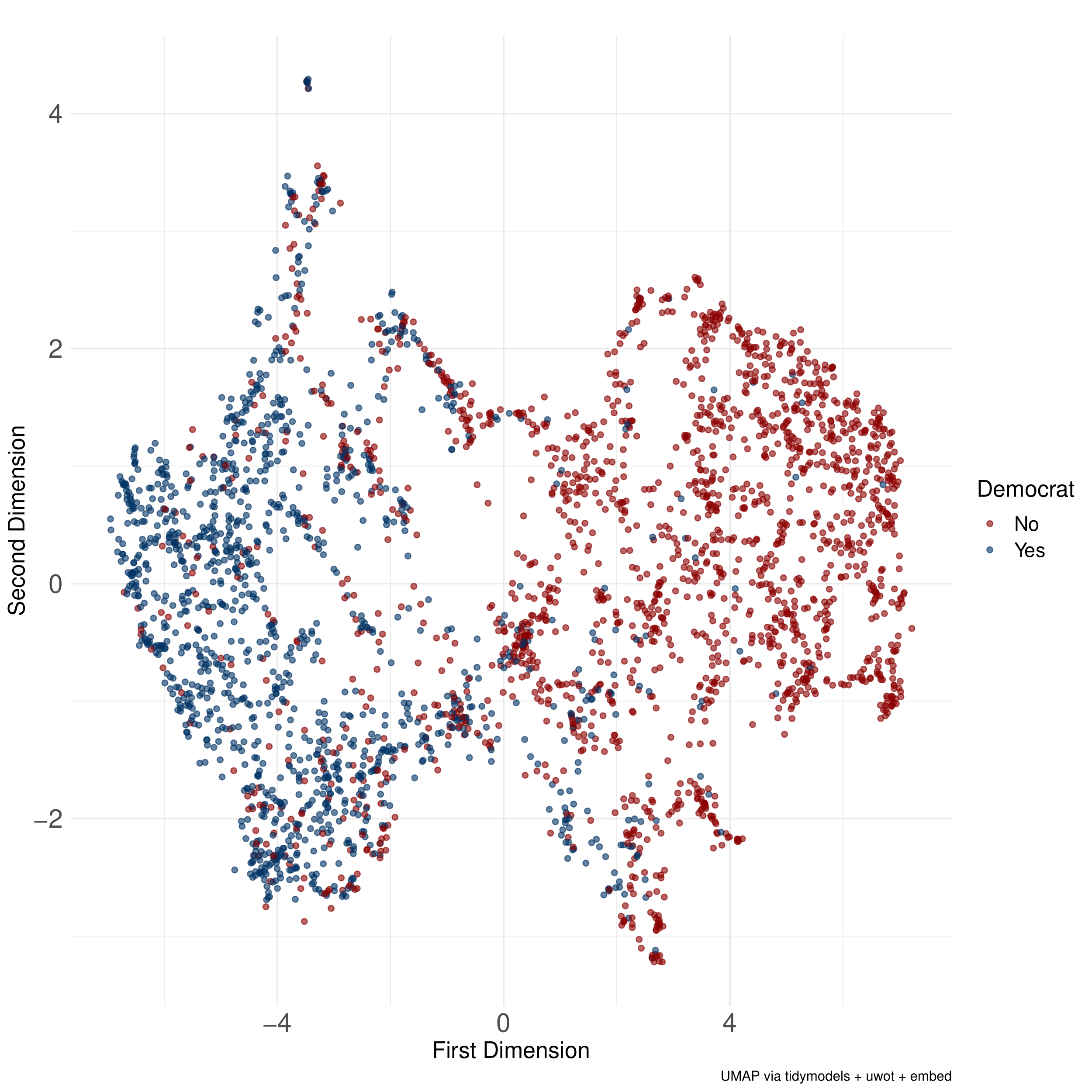}
	\caption{A Tidy UMAP Solution}
	\label{figure:tidyumap}
\end{figure}

A second order benefit of this tangent is to demonstrate that, though some packages differ in programming philosophy, the results are stable across different constructions. I encourage readers to inspect package documentation for \texttt{umap} and \texttt{uwot} (for the tidy implementation) packages to understand how they differ. With that, consider the following code that uses similar syntax as in the feature engineering code in Section 2 of the Element, to set up the embedding process via UMAP. Then, I pipe the projection to a \texttt{ggplot} and get a virtually identical rendering of the projection subspace in Figure \ref{figure:tidyumap}.

The take away from Figure \ref{figure:tidyumap} is that, though placement and scales are different compared to the \texttt{umap} package approach, the algorithm still learns the underlying manifold along which parties are clearly separated. Substantively, given UMAP's nonparametric and unsupervised architecture, this offers an additional point of validation of results across several techniques that these ANES preference data are relatively stable and reasonably well-separated along a partisan dimension. A possible take away could be that the many dimensions used in these types of surveys might be redundant, as we are able to pick up on a lot of information just by looking to party affiliation of the respondent. Of course, a generalization of this sort is not possible from these data alone, but the possibility exists and deserves a closer look.

\subsection{Suggestions for Further Reading} 

\citet{wattenberg2016use} offer a widely used, practical guide to t-SNE, which gives a nice complement to the coverage of t-SNE in this Element. Further, \citet{waggoner2021pandemic} presents an application of UMAP in American Congressional policymaking related to understanding COVID-19 policy structure. \citet{ordun2020exploratory} directly compare UMAP to topic models and digraphs using COVID-19 Twitter data. These two applications offer a ``real-world'' view of these methods and their value in practice.

\clearpage

\section{Neural Network-Based Approaches} 

In this final section, we shift gears to a very different approach to dimension reduction at a conceptual level, though technically we are still interested in some weighted representation and transformation of input features to give the low dimensional representation of the high dimensional data space. The conceptual shift is in how to think about the data first, and then how to conduct the dimension reduction exercise. That is, we now consider dimension reduction as a neural map. Within the neural network-based approaches, which are still unsupervised, there is further nuance, mostly centering on the number of hidden layers. If we have zero hidden layers, for example, we are in the world of self-organizing maps. But if we have 10 hidden layers, we are in the world of deep autoencoders. But before we get too far down the path, we will start with a high-level look at a neural network in order to place unsupervised dimension reduction in a neural network-based context. 

\subsection{A Basic Neural Network Architecture}

Neural networks are widely used computational modeling tools, which can give extremely accurate predictions. Though mostly used for supervised learning tasks, their architecture (stacking layers for processing tasks) can be adapted to an unsupervised framework for dimension reduction. But to understand their construction in such a light, an understanding of the basic architecture is required. As a caveat, this section is only interested in presenting the very basic architecture. Indeed, many complications and nuanced extensions exist. The goal at present, rather, is to introduce unfamiliar readers to the structure of a neural network and how it processes data to give predictions in order to understand them in in the dimension reduction context. For a more exhaustive treatment of neural networks and in a deep learning framework, see \citet{goodfellow2016deep}. 

The simplest neural network is comprised of three layers: input, hidden, output. The input layers is the raw data matrix we have seen several times to this point, $\mathbf{X}$, which is simply an $N \times P$ matrix of raw input features. The output layer is the response feature/outcome/dependent variable. This is where predictions of the output are produced. The hidden layer is sandwiched between the input and output layer and is where the processing of data occurs. Before getting to processing of data, it's important to note that a neural network is connected by weights and biases.\footnote{The term \textit{bias} here is not the same as in supervised prediction tasks referring to mistakes.} A weight is the value attached to the edges in the network connected the nodes that comprise each layer, and the biases are another type of weights that each node is assigned. Upon initialization, all biases are randomly assigned, amounting to a less-than-intelligent brain, akin to making a random guess about some outcome before seeing and thinking about any data or patterns therein. The task of the neural network, then, is to pass raw data through the network one or a few observations (called ``batches'') at a time. The network processes the data in the hidden layer on the forward pass through the network, and then renders a ``decision.'' This prediction is compared to the labeled outcome feature values and the error is recorded. The data are then passed backward through the network through a process called \textit{backpropagation}. During backpropagation, weights and biases are iteratively updated in an incremental way, in reaction to the error recorded after the forward pass (or ``forward propagation) of the data. Then, new batches are passed through the network, but this time using the new weights and biases learned during the previous iteration. Error is recorded after forward propagation, and then during backpropagation weights and biases are updated once more. This process is repeated until the network has seen all of the data once. Seeing the full data one time is called an \textit{epoch}.\footnote{We encountered this term in the previous section on UMAP. The definition is the same: the number of times the algorithm sees the full data set} So, there can be multiple iterations in a single epoch. We then have the choice of continuing to learn until we reach some threshold, e.g., reaching a prespecified number of epochs (via tuning a hyperparameter) or once some acceptable level of error is reached (similar to the stopping criterion in k-means clustering). Importantly, there are \textit{many} decisions to make when training a neural network (e.g., how large should the updating steps be in backpropagation, where smaller steps mean more precise learning but potential failing to converge and larger steps mean quicker convergence, but potential a poorly performing model that fails to generalize well. This decision is controlled by the learning rate hyperparameter, often referred to as $\eta$). There are many flavors of neural networks, such as a fully connected artificial feedforward neural network (this is the basis type previous described) or a convolutional neural network (most often used for image recognition). And further, a key decision which may place you in an entirely different subfield is how many hidden layers should you include to process and learn from the data? As the hidden layers increase, the \textit{depth} of the network also increases, eventually placing you in the deep learning world. In short, the point is neural networks can get extremely complex really quickly.

\subsection{Self-Organizing Maps}

Self-organizing maps (SOM) were first introduced in \citet{kohonen1982self}. Based originally in topological data analysis, and closely related to clustering techniques, SOM have not always been thought of as a special case of a neural network (or equivalently, constrained k-means clustering), as neural networks have only recently become known by such terminology. In brief, SOM can be thought of as a special case of a neural network, where we are interested in mapping an input layer to the output layer. But the special case is defined by the lack of a hidden layer to process the data in the SOM. Rather, the input layer is directly connected to the output layer through a series of weighted edges, which are iteratively updated as the structure is learned from the input layer. Let's pull this apart a bit more. 

As with a basic neural network, the input layer consists of the raw input features. The output layer in a SOM is a fixed lattice or grid with nodes set by the researcher a priori. For example, a SOM might be mapped onto a $10 \times 10$ lattice, where there are 100 total nodes in the output layer. The goal is to place observations into nodes, such that observations closer to each other in high dimensional space are represented closer together in the lower dimension map. For example, there should be spatially similar observations placed together in a single node or in a very close, neighboring node. The SOM algorithm is interested, then, in learning the patterns of similarity across observations in the high dimensional space and mapping them based on these similarities in a smooth, lower dimensional subspace, which is the fixed lattice. We will come back to this more in a bit. 

Central to organization of the map is \textit{learning}. Learning in SOM is based on the idea of competitive learning, where these fixed nodes are, in effect, competing with each other to represent some portion of the data in the two-dimensional output layer/lattice. This type of learning is in contrast to more classic approach to learning based on error-minimization as we have seen with many of the techniques to this point. Importantly, it's useful to point out similarity to several other techniques covered and mentioned. First, SOM is neural network-based in that we are placing data in layers in an attempt to map one layer onto another layer. But SOM is also constrained k-means as mentioned above, where similar observations are clustering together. Yet, further still, by projecting the high dimensional input layer onto a two-dimensional, fixed lattice for the purpose of mapping, we are also engaging in dimension reduction. Thus, SOM can be thought of as a combination of neural networks, clustering, and dimension reduction wrapped into a single technique. This will be useful later in the section at different points. 

Put simply, a SOM learns and gives a solution based on structuring the nodes in the output layers into \textit{clusters of nodes}, where clusters are formed based on spatial proximity and similarities across observations. As with a neural network, the SOM algorithm is initialized with random weights connecting the input layer to the output. Then, the weighted raw feature values are projected onto the lower dimensional lattice, and are in the truest sense, mapped onto this space. Then, through a series of iterations, the weights are updated based on spatial similarity. The result then is some \textit{nodes} are similar to each other, relative to other nodes that are more different and spatially distant. As learning is competitive, and as competition amongst nodes to represent some large chunk of the data space is what drives the algorithm, the ``winning'' node that represents the bulk of the observations is called the best matching unit (BMU). The BMU defines the center of the neighborhood of a cluster of similar observations. There can be multiple BMUs in a single grid, suggesting multiple clusters within a common data space. For our example using the ANES data and searching for latent partisan separation in the projection space, we might expect roughly two, possibly three clusters surrounding two or three BMUs in the output layer. We will come back to this in the application. 

\subsubsection{Defining the Steps of the Algorithm}

The configuration of the topology of a SOM output layer is defined by iterating across three main steps. The first step is \textit{competition}, where nodes neurons compete to represent the input space as previously mentioned. The smallest distance between the input values and connection weights gives the BMU. The second step is \textit{cooperation}, where the winning node becomes the center of the neighborhood of nodes in the output layer. This step of cooperation is critical to SOM learning and representation being treated as a neural network, based on something called Hebb's Rule, which most simply is, \textit{neurons that fire together are wired together} \citep{hebb1949organization}. Hebb's Rule, which postulates learning in the human brain based on synaptic connections between external stimuli and information processing in the neuron, results in \textit{smooth learning} within some neighborhood of neurons. That is, neurons in the human brain do not learn or are stimulated in isolation, but rather do so in small neighborhoods. There may be a single, central neuron that is most stimulated by the raw input (e.g., the BMU in SOM), but there is also a neighborhood of surrounding neurons that reflect the same stimulation, only to smaller degrees. This notion of smooth learning within neurons is modeled in the second step of the SOM algorithm. Where the BMU is the representative, winning node representing the most similarity of all other nodes for some part of the data, the surrounding nodes are also representing the same similarity, by virtue of their proximity to the BMU, only to a lesser degree. This process is controlled by a (usually Gaussian) decay function, e.g.,

\begin{equation}
d_{ij} = e^{(\frac{-d_{ij}^2}{2\sigma^2})},
\end{equation}

where $d_{ij}$, between the winning neuron, $i$, and the neighbor unit, $j$. $\sigma$ defines the size of the neighborhood. Of note, $\sigma$ is a hyperparameter set by the researcher, which is often referred to as $R$ for ``radius,'' given that we are defining a neighborhood or region around the BMU. We start with a large amount of learning at the outset to pick up on clear patterns, and slowly refine the learning as the iterations increase. That is, the learning rate and amount of cooperation with neighborhoods of nodes exponentially decays as the patterns of clusters become clearer, e.g., $e^{\frac{-M}{V}}$, where $M$ is the number of iterations, and $V$ is a constant.

The third and final step is \textit{learning}, where nodes in the neighborhood of the BMU participate and update together, such that the higher the weights, the greater the chance the BMU continues to be the BMU in subsequent iterations. Put differently, based on the previous two steps of competition and cooperation, the patterns in the data, and the local groups existing across the lattice become increasingly clearer as the algorithm sees the data multiple times. The seeing of the data and updating is indeed \textit{learning}. Recall, as we are interested in spatial similarities, which defines a well-fitting solution, the scoring function is calculated as the distance between measured values from the input layer to their projected positions in the output layer.

To recap, the first step of competition means that for each output neuron, $j$, we compute the distance between each weight vector and input vector, $d(\boldsymbol{W}, \boldsymbol{X})$. We find the BMU, $i$, as the node that minimizes this distance over all output nodes. The next step of cooperation is defined by identifying all nodes in the output layer, $j$, within the neighborhood of $i$, defined by the neighborhood size, $R$ or $\sigma$. Finally, in the third step, for all nodes in the neighborhood of $i$, we update the connection weights, $w_* \in \boldsymbol{W}$, accordingly, 

\begin{equation}
w_{new} = w_{old} + \eta(\boldsymbol{X} - w_{old}),
\end{equation}

where $\eta$ is the learning rate, $0 < \eta < 1$. The algorithms stops when some criterion is met, which is usually when either the weights stabilize, or cluster configurations are unchanged. The result is a lower dimensional (that is, 2D) representation of the high dimensional input space.   

\subsubsection{Applying SOM to the ANES Data}

To begin the application of SOM using the ANES data, we need to scale that data, as we are working with weighted versions of raw input features as with PCA and other similar techniques in this Element. Upon scaling, we set up the output layer, which recall is a two-dimensional lattice. For our case we will set it up to be $10 \times 10$, and rectangular (instead of hexagonal). For most of the application, we will use the \texttt{kohonen} package. For some of the visual and data organization features, we will use core tidyverse packages, e.g., \texttt{ggplot2} and \texttt{dplyr}. 

With the grid constructed, and neighborhoods within the grid as Gaussian for the second \textit{cooperation} step previous discussed, we are ready to fit the algorithm. Upon passing the input data and the output grid we previously defined in \texttt{search\_grid}, there are a few additional hyperparameters we need to set prior to running: \texttt{alpha} (learning rate, which exponentially decays as previous discussed, from 0.1 to 0.001), \texttt{radius} (neighborhood size around neurons), \texttt{rlen} (number of iterations), and \texttt{dist.fcts} (distance metric). These hyperparameters apply globally to the model, and will constrain the training process. The total training process took about 13 seconds to locally run.

\begin{figure}[h!]
	\centering
	\includegraphics[scale = 0.5]{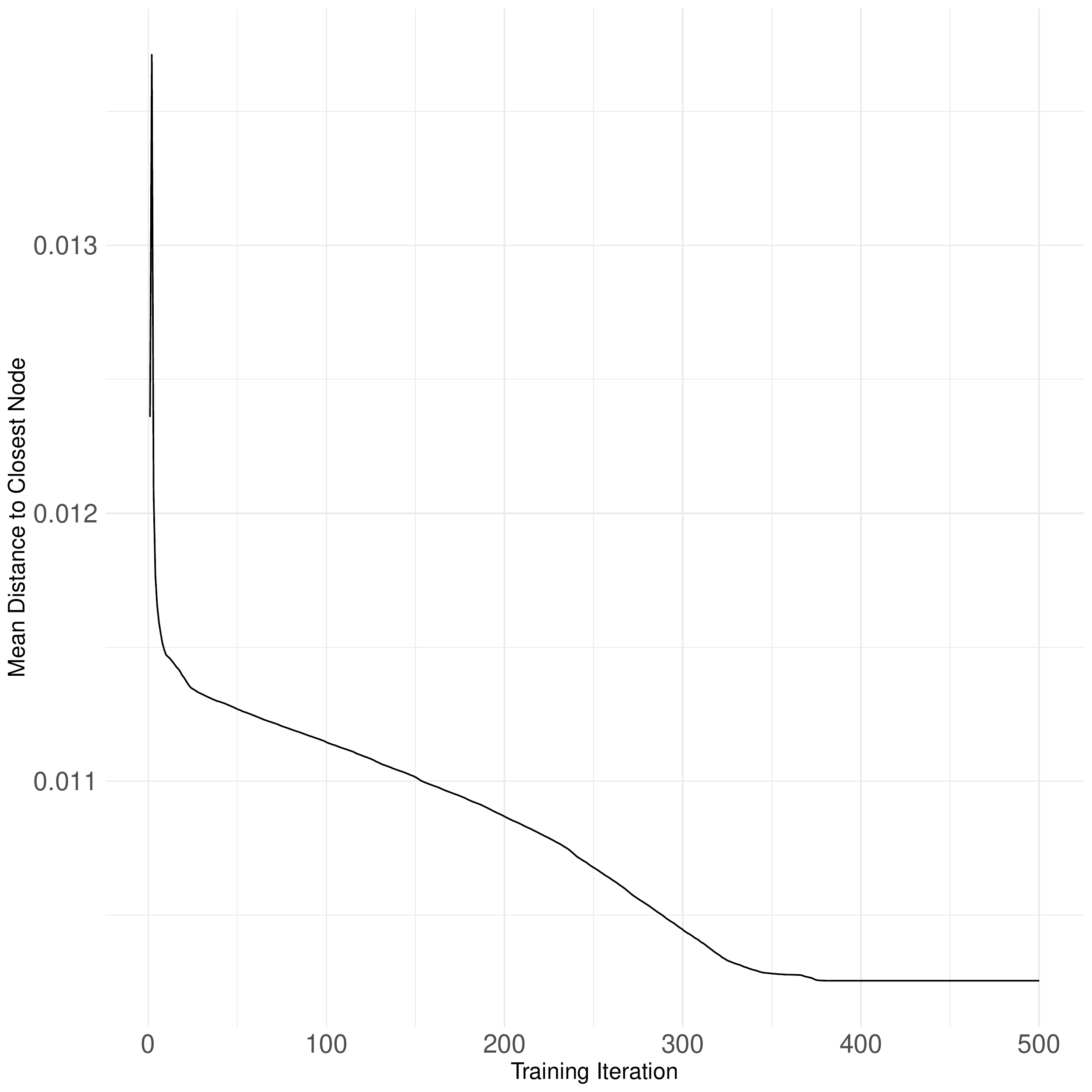}
	\caption{Training Progress for the SOM Fit}
	\label{figure:som1}
\end{figure}

Once fit, there are a variety of approaches to inspecting output from a SOM. A good place to start is to visualize the training progress. That is, \textit{how long did it take over the iterations, \texttt{rlen}, for the weights to stabilize, and thus the error to flatten out?} We can inspect this by calling the value, \texttt{\$changes} from our \texttt{som\_fit} object, shown in Figure \ref{figure:som1}. 

In Figure \ref{figure:som1} we can see that the algorithm stabilized around 350 iterations or so, of 500 that we set. Importantly, several choices can impact the performance of the algorithm. For example, the size of the grid, where smaller grids offer less space to map the data to, giving the possibility of a poor solution unable to detect separation. Alternatively, setting \texttt{rlen} too small will not allow the algorithm time to sufficiently learn the latent patterns in the data, and thus insufficient time to stabilize and find an optimal solution. These and all other choices need to be made carefully. 

We can plot SOM results in many other ways such as summed distances between neighbors in the grid space (\texttt{plot(..., type $=$ "dist.neighbours")} or the mean distances between units and observations (\texttt{plot(..., type $=$ "quality")}. Though these diagnostic plots are useful, for the sake of space I point readers to the documentation (e.g., \texttt{?plot.kohonen}), and focus instead on tying results back to our substantive example of understanding latent partisan structure in feeling thermometers. To this end, I will focus on the \textit{codes} produced by the trained SOM. 

Codes, which are sometimes called weight vectors (called via \texttt{\$codes}), show the representation of each feature in each node. These, which are akin to feature loadings in PCA, give us a clear understanding of the organization of the SOM solution in the output layer. Some features will contribute to different locations on the grid, compared to other features. By inspecting the code, then, we can better understand which features are contributing to the segmentation of the space. The standard output from a SOM fit in R is something called a fan plot, where the size of each fan blade, one for each feature in each node, indicates magnitude (bigger = greater magnitude). Thus, there are individual fan plots in each node in the output from a basic call of the \texttt{plot()} function. Yet, this plot is less intuitive and does not fully describe the value of the SOM solution for helping toward our substantive goal of understanding whether separation along a partisan dimension characterizes these feeling thermometer data. Therefore, a more informative approach is to fit a clustering algorithm to the codes output from a SOM. The intuition here is to explore whether codes are naturally grouped along a substantive dimension. For our case, of course, this is along a partisan dimension. 

To do this, we require a few steps. First, for ease of plotting, store the point colors used throughout in two objects: \texttt{point\_colors} and \texttt{neuron\_colors}, with the former darker shades corresponding with the individual observations and the latter lighter shades corresponding with the nodes/neurons in the output layer (the background in the grid shown in the Figures below). Then, I fit a k-means algorithm to the codes data, searching for two clusters given the dichotomous party affiliation feature used throughout the Element. Then, derive cluster labels (1 or 2) for the clusters, and plot accordingly varying color by party affiliation. The expectation is that the majority of observations (points) should correspond to the color of the nodes found from the k-means clustering solution. If points are scattered and not clearly coordinating, then this would suggest there is not clear separation in the output layer. 

\begin{figure}[h!]
	\centering
	\includegraphics[scale = 0.5]{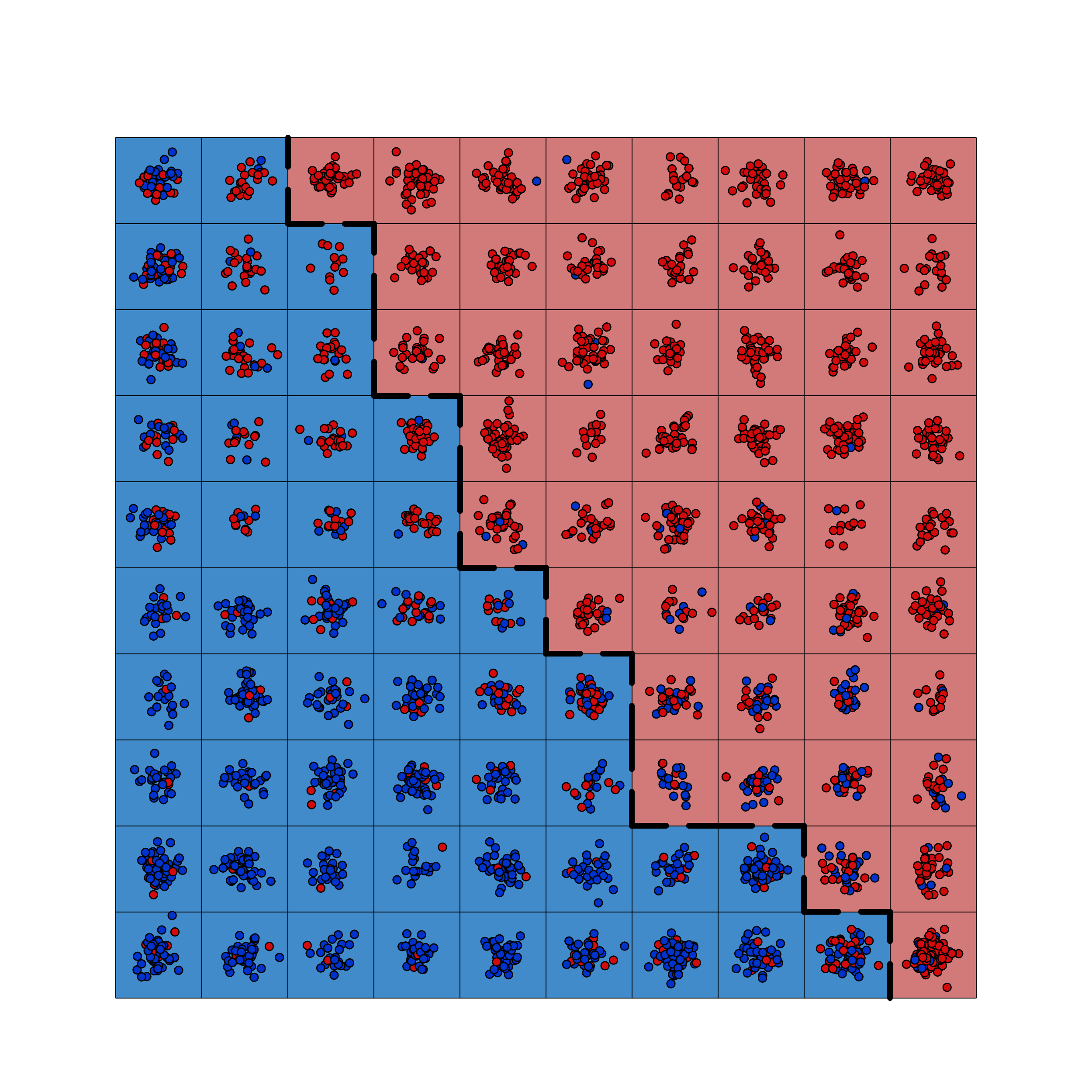}
	\caption{Searching for 2 Clusters via K-Means}
	\label{figure:som2}
\end{figure}

Indeed, Figure \ref{figure:som2} shows clear separation in the output layer along a partisan dimension. Red points (Non-Democratic) are mostly grouped on the red side of the output layer, compared to blue points (Democratic) being mostly groups on the blue side of the output layer. There is not perfect separation here but based on the results throughout this Element showing separation between parties, but still some blending across parties, these results corroborate this pattern. 

\begin{figure}[h!]
	\centering
	\includegraphics[scale = 0.5]{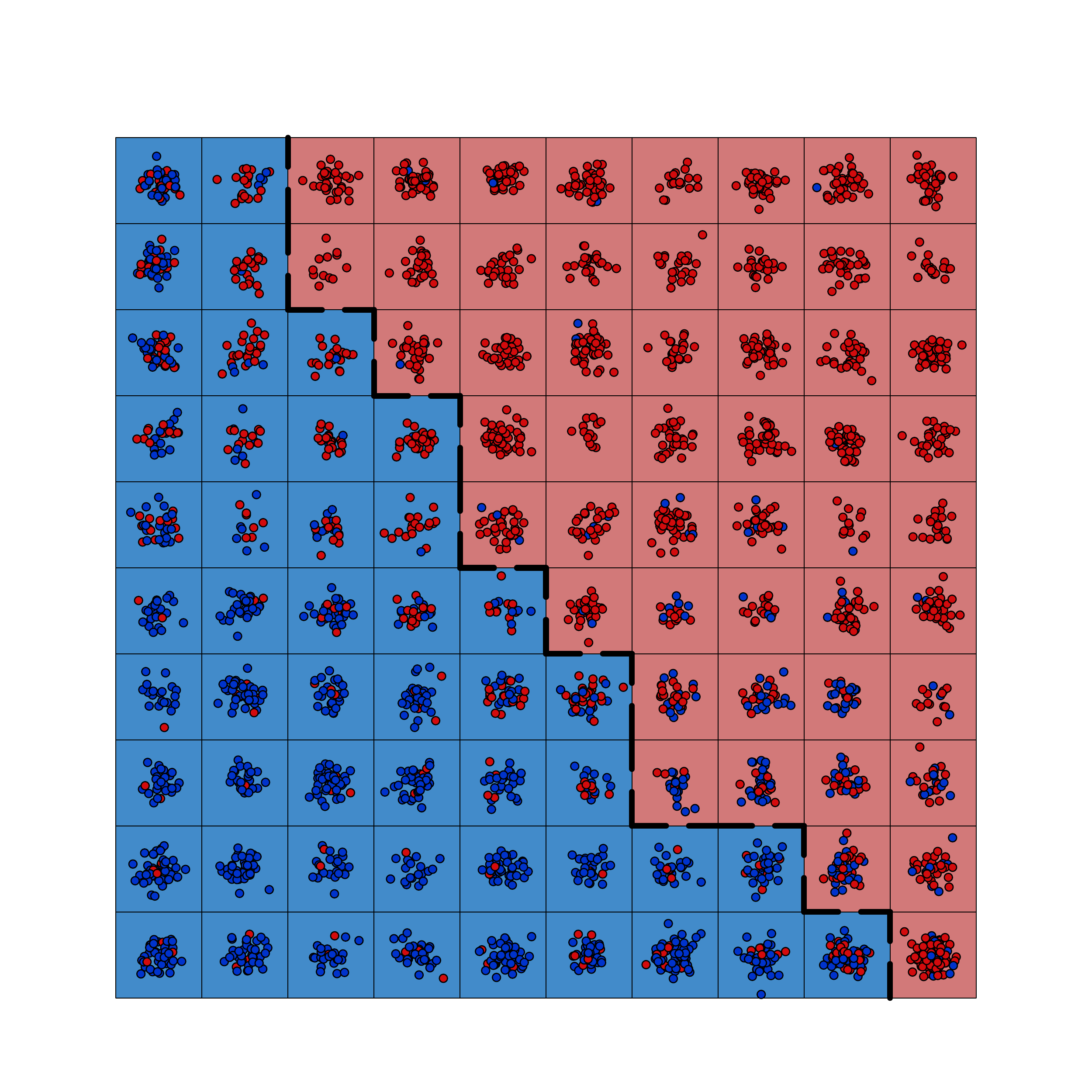}
	\caption{Searching for 2 Clusters via Fuzzy C-Means}
	\label{figure:som3}
\end{figure}

But note, k-means clustering is more of a brute force approach to clustering, where an observation is assigned to one and only one cluster and must be assigned to a cluster (rather than left unclustered). But given the blending of several observations across parties we have seen to this point, such a hard partitional approach to clustering may not be the best approach to clustering. Rather, a soft clustering approach like fuzzy c-means clustering, which is based on a majority decision from a fractional assignment of observations theoretically belonging to both clusters, just to varying degrees, may be more appropriate. Note, for readers unclear on these approaches to clustering in the context of social science problems, I point them to the recent Element in this series, \citet{waggoner2020unsupervised}, which details these two and several other algorithms for partitioning some data space in an unsupervised way. We can now update our solution, but this time using fuzzy c-means via the \texttt{fcm()} function from the \texttt{ppclust} package. See the results in Figure \ref{figure:som3}. 

The results in Figure \ref{figure:som3} look slightly more like the patterns seen elsewhere, though they are not extremely different from the patterns in Figure \ref{figure:som2} showing the k-means version. Separation for most of the observations is clear, and there are some overlapping cases in both the k-means and fuzzy c-means (FCM) versions. Even still, FCM is more theoretically motivated given the potential of observations in the middle of the two dimensions plausibly being ``assigned'' to either party. Thus, the FCM approach would be more appropriate in this case.

Substantively, codes also provide a sense of the specific features that are attracted to similar nodes, such that we are able to pick up on grouping across features in the output layer. For example, similar features (e.g., feelings toward Barack Obama and Bernie Sanders) might heavily characterize (or ``weight'') a node based on similarity, compared to other features (e.g., feelings toward Donald Trump). The result is a picture of the \textit{relationships} between both features and nodes across the full input space. 

Given the high dimensional space with which we are working, I demonstrate this view of feature-level code explorations by a sampling of a feature-by-feature basis and tie the results back into a substantive understanding and exploration of the feeling thermometer space from the ANES. Such a use of codes from our fit SOM results in a slightly nuanced view of these weight vectors compared to more common interpretation previously discussed. Substantively, we can think of weight vectors/codes like we might correlations between features and nodes. Features that are more similar will trend toward nodes in a \textit{positive} direction, suggesting similar grouping and thus similarity in latent structure across those features. This is compared to features that are more different from each other, which will trend \textit{negatively}. No relationship would result in a uniformly distributed cloud of observations with little to no slope. 

\begin{figure}[h!]
	\centering
	\parbox{7cm}{
		\includegraphics[width=7cm]{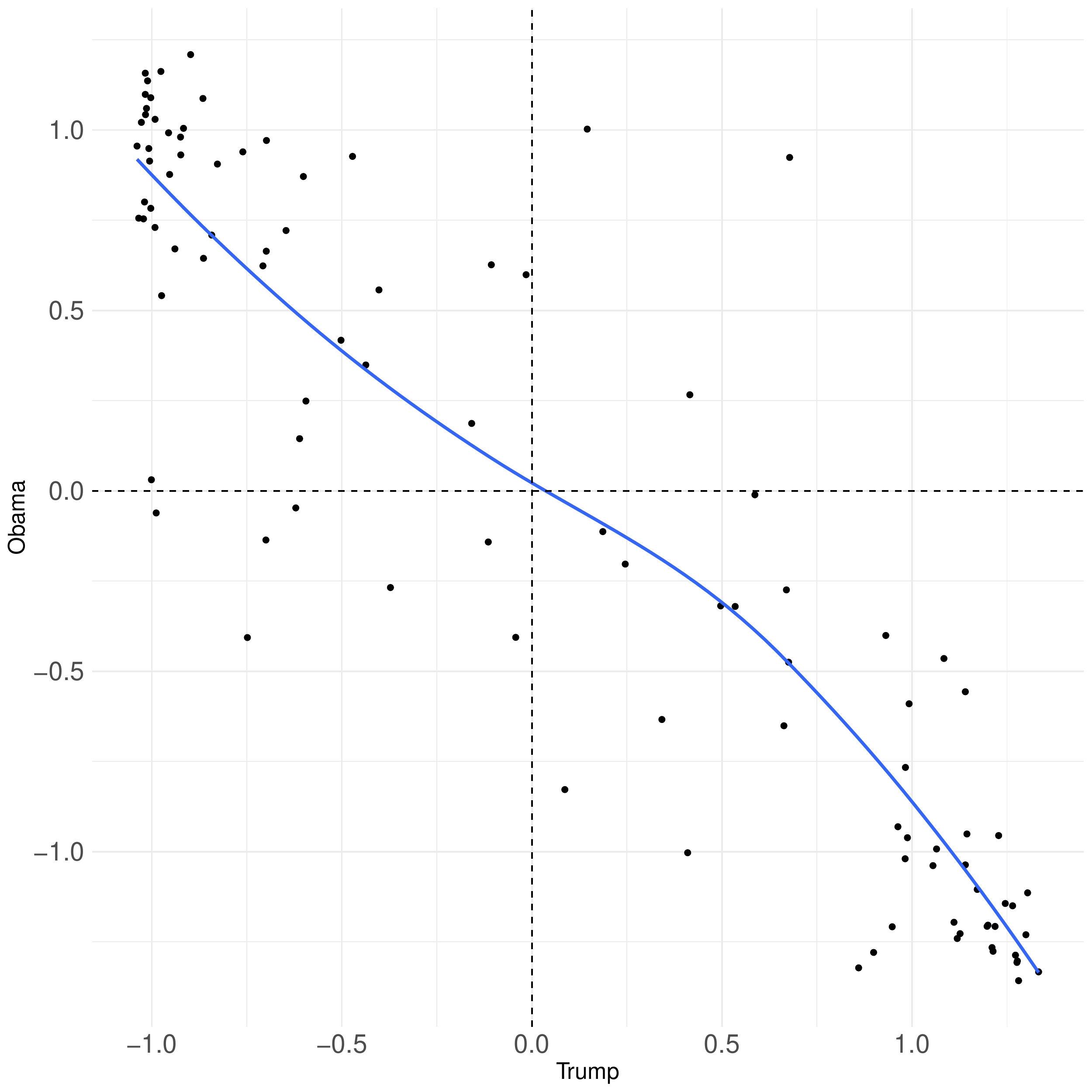}
		\caption{SOM Codes: Trump and Obama}
		\label{figure:som4}
	}
	\qquad
	\begin{minipage}{7cm}
		\includegraphics[width=7cm]{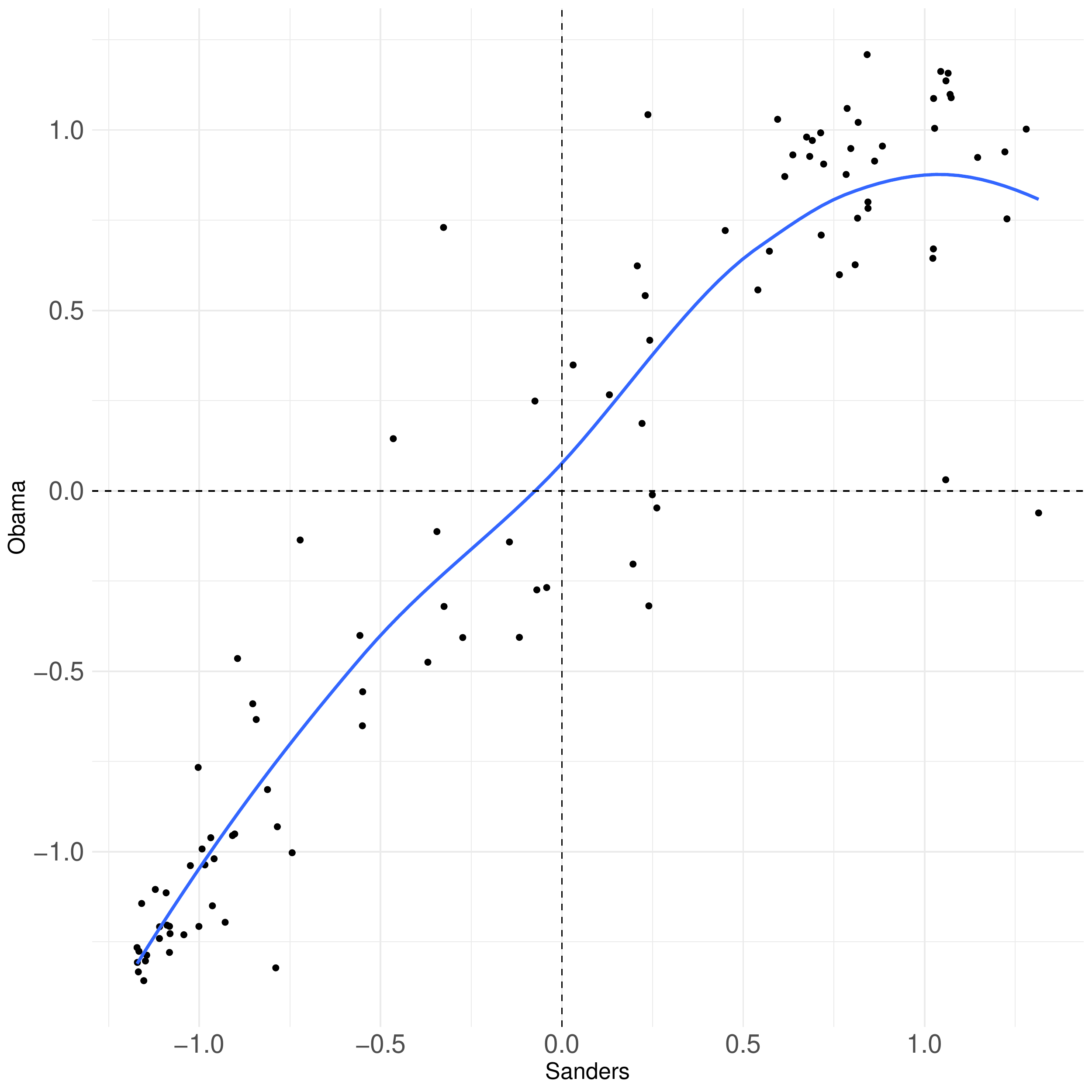}
		\caption{SOM Codes: Sanders and Obama}
		\label{figure:som5}
	\end{minipage}
\end{figure}

Consider the direction of the codes between feelings toward Trump and Obama. We might expect a \textit{negative} relationship across these features. Inversely, we can see that features that we might expect to be more similar to each other or picking up common structure to be trending in a positive direction. For example, we can see this positive trend in feelings toward Bernie Sanders and Barack Obama. See both of these examples inf Figures \ref{figure:som4} and \ref{figure:som5} with an overlaid loess smoother to aid visualization of the trend. 

\begin{figure}[h!]
	\centering
	\parbox{7cm}{
		\includegraphics[width=7cm]{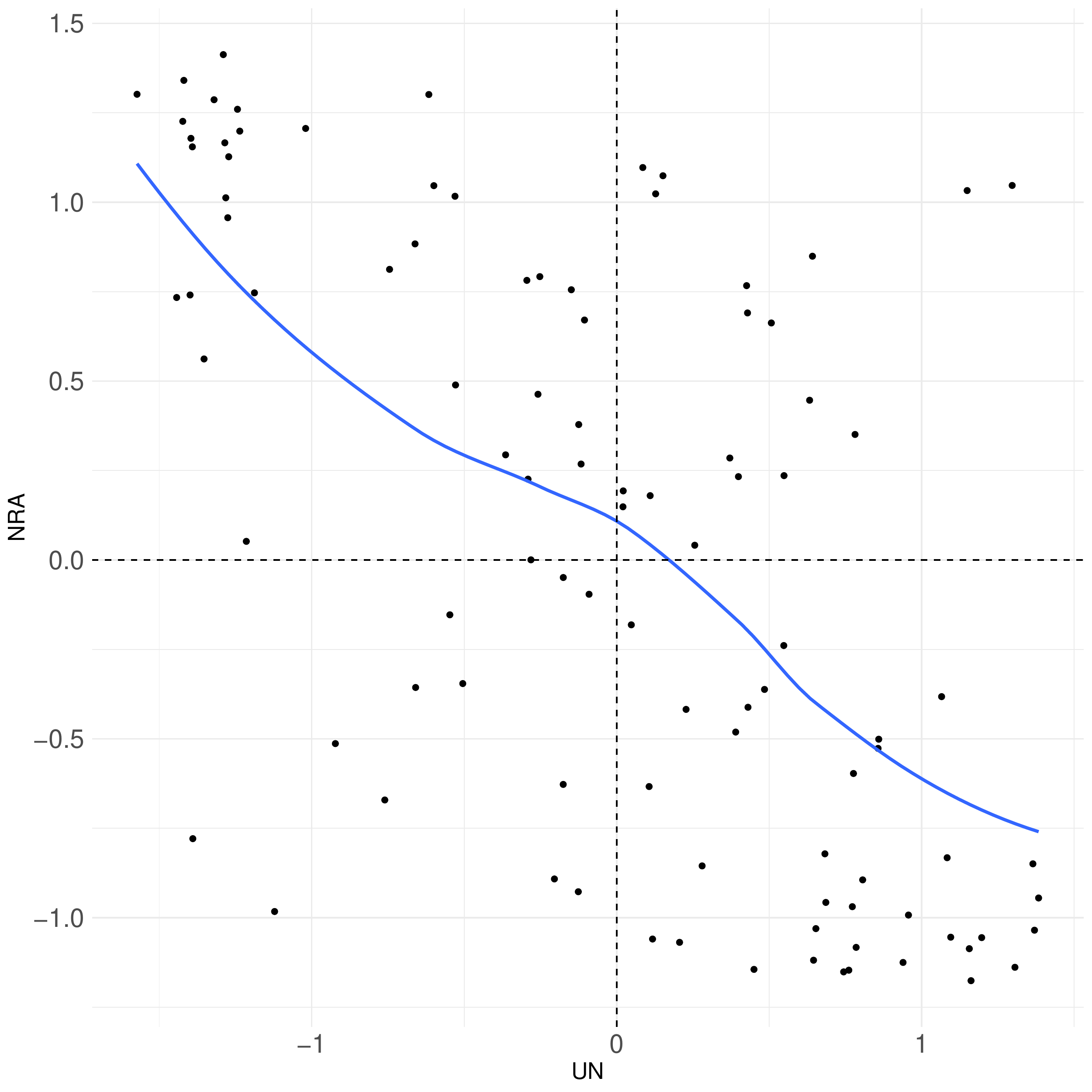}
		\caption{SOM Codes: UN and NRA}
		\label{figure:som6}
	}
	\qquad
	\begin{minipage}{7cm}
		\includegraphics[width=7cm]{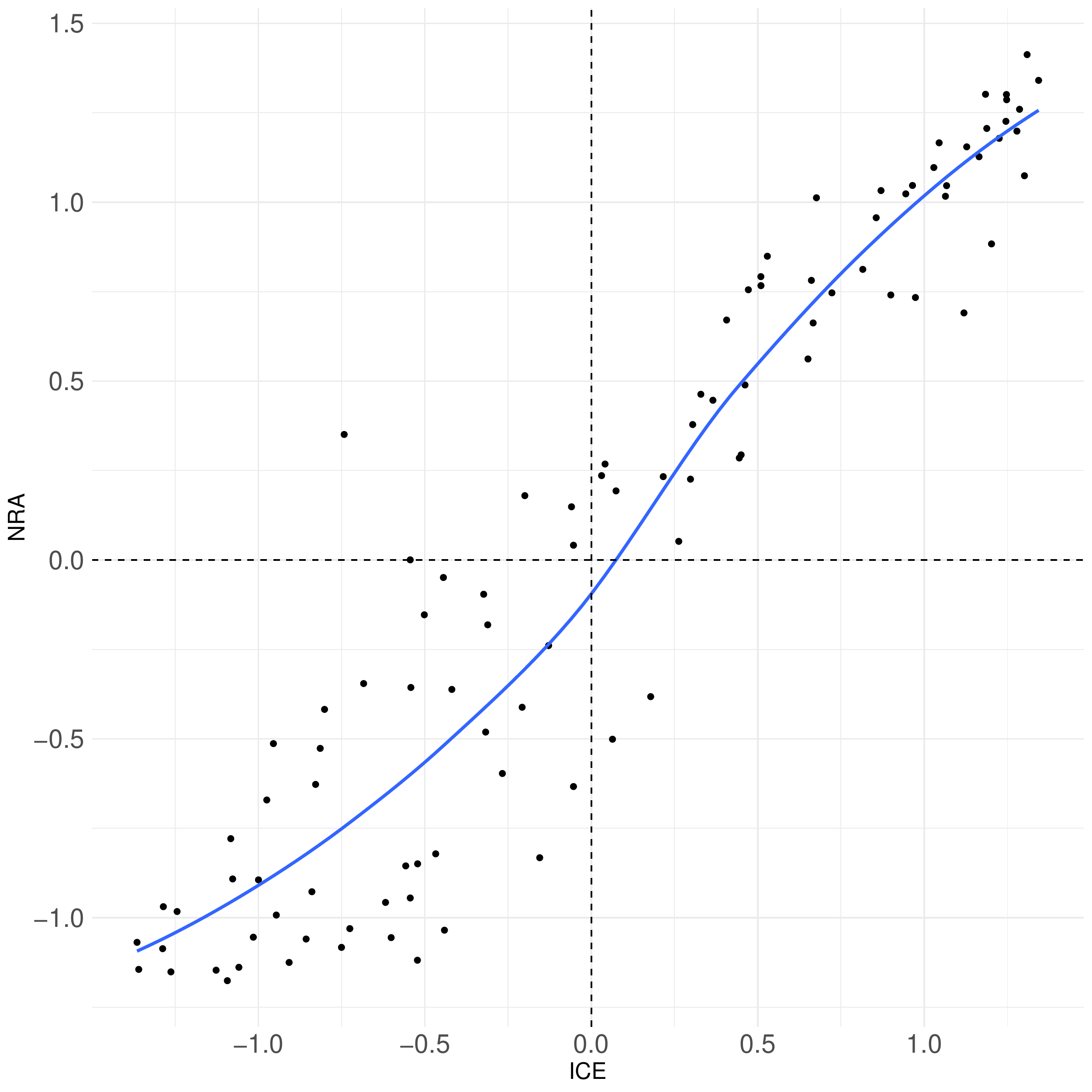}
		\caption{SOM Codes: ICE and NRA}
		\label{figure:som7}
	\end{minipage}
\end{figure}

These simpler expectations are clearly negative and positive as we might expect across political actors in Figures \ref{figure:som4} and \ref{figure:som5}, respectively. Though more overt, the same logic holds for base expectations across \textit{institutions}, as we might expect negative relationships between feelings toward the UN and the NRA, but positive between ICE and the NRA. As a final ``sanity'' check, this is exactly what we see in Figures \ref{figure:som6} and \ref{figure:som7}, suggesting the SOM is picking up this latent structure in line with substantive expectations.

In sum, the task has remained constant, which is to learn patterns in the data and then project these patterns in a lower dimensional subspace. Yet, the views of SOM as a blend of clustering, neural network, and dimension reduction, mean that simpler interpretation is not as obvious as in PCA, for example, where we clearly get a lower dimensional solution with PCA scores for the newly calculated PCs/features. Rather in SOM, we are certainly reducing the dimensionality of the input space, but we are doing so in a neural network-based way, and of equal importance to dimension reduction in a typical SOM application is the clustering of observations in the output layer. This potential downside in interpretation is outweighed by the extremely flexibility and wider value from the solution, compared to more nuanced dimension reduction techniques covered to this point. 

It is worth concluding with a word of caution when using SOM. Like t-SNE, the solutions will never be identical, as the process of weight-updating and learning is stochastic. Thus, different implementation of SOM can potentially give different solutions, which limits generalizability and applicability of this method. Further, SOM are entirely dependent on the shape and size of the output layer, which recall is a grid/lattice pre-set by the researcher. Set too small, and no structure can be revealed. Set too large, and parts of the grid will be left unpopulated by observations, potentially implying much greater separation than may truly exist in the data. Thus, selecting the ``right'' grid size is not always an apparent choice, and requires careful validation and transparent reporting. I recommend a similar approach when reporting SOM results as the \textit{Risk Averse Workflow} discussed at the conclusion of the LLE section. Still, though, if situated in a broader context of dimension reduction and compared to several other techniques, SOM can be a very powerful tool for nonparametrically reducing the complexity of a high dimensional data space. 

\subsection{Autoencoders}

We come now to the final technique: autoencoders. Autoencoders are built on a similar intuition as PCA, and several other techniques covered to this point, but all from a neural network-based framework. In brief, autoencoders taken an input data matrix, create an encoded version of that data matrix by forcing information loss (much like in PCA), and then in the final stage they attempt to reconstruct the original input space, but only on the restricted, encoded version. The goal, then, is to create a reduced version of the input space and then essentially assess the quality of that encoded version by testing its ability to inform a good reconstruction of the original high dimensional input space. 

A typical autoencoder topology, then, consists of three layers, like the basic feedforward neural network we discussed at the outset of this section: input layer, hidden layer, and an output layer. As before, the input layer is the matrix of raw data inputs including all features, which in our case is the set of feeling thermometers from the ANES. The hidden layer is also where data is processed. But in an autoencoder architecture, ``processing'' information equates to compressing the high dimensional data space to be on a lower dimensional subspace. That is, the hidden layer is of a size less than the size of the input layer, such that a bottleneck essentially forces the input layer to be at least of size $p - 1$, though in practice the hidden layer is much smaller than the input layer. This process of data compression in the hidden layer results in a new, lower dimensional version of the data, as we saw from PCA, for example. But the output layer is the same size as the input layer, where the task of the autoencoder in the latter half of the network is to reconstruct the original input space on the basis of the compressed version emerging from the hidden layer. 

Thus, there are two steps to a basic implementation of an autoencoder: first, encode; second, decode. The encoding step is the data compression/bottleneck step. The second decoding step is reconstruction step. As with other neural networks, once the batch of data has made its way through the full network, error is calculated prior to backpropagation. Yet, at this point readers might be wondering: \textit{I thought this was unsupervised, so how do we conceptualize and calculate error?} To address this, think about the decoding step; it is concerned with \textit{reconstructing the input layer}. Thus, the target is a perfect replication of the input layer. So the input layer acts as our ``labeled data,'' thereby giving a ground truth to which we are able to compare the decoded version of the data space. This is not a ground truth as we might consider in a typical supervised task like classification, where the ground truth is a labeled class, $\{0, 1\}$ or $\{\textrm{no}, \textrm{yes}\}$, or $\{\textrm{Republican}, \textrm{Democrat}\}$. Rather, the ``ground truth'' for an autoencoder is simply the original input space. In this way, we are able to calculate error (comparing the decoded version to the original version), which is called \textit{reconstruction error}, while still remaining firmly in an unsupervised framework. Put simply, then, the task of the autoencoder is quite simply to learn effective patterns or structure underlying some input set of data. This is exactly the goal of unsupervised dimension reduction as we have addressed throughout the Element. The main difference with autoencoders compared to other dimension reduction techniques, then, is that it is rooted in a neural network-based approach to computational modeling (different from LLE or PCA, e.g.), but still very different from a feedforward neural network for a supervised problem (e.g., regression or classification).

To reiterate, we are interested in training the network to minimize reconstruction error. By backpropagating through the network and iteratively updating weights based on learned structure through the reconstruction error, the autoencoder is, in a very true sense, \textit{learning} from the data in an unsupervised way. 

Vastly important to understand with autoencoders is the role of the hidden layer. As previously noted, the hidden layer forces information loss at the encoding stage, and thus constrains the picture of the high dimensional data the decoder gets to see. By creating a lower dimensional representation and then forcing the reconstruction of the original high dimensional input space, the resultant bottleneck is intentionally making the decoder work to learn underlying non-random structure in the data. Indeed, if there were loss of information, then the task of the decoder would be to simply multiply the input space by 1, which is a useless task, as there would be nothing to decode. 

From an autoencoder, as with PCA, we get a set of newly constructed features. Though called principal components in PCA, these are now called either \textit{codings} or \textit{deep features} in autoencoders, depending on the depth of the network. And by depth, I mean increasing the number of hidden layers, or opportunities to process and learn from the data. The more hidden layers, usually the better the solution. However, too many hidden layers, and we are in danger of overfitting, or learning too much nuance from the training set, such that we are left with poorly generalizable solution. Still, deep autoencoders of this sort can be extremely powerful. 

A final point of clarification is the number of nodes or neurons in the hidden layer. If there are more of these than in the input layer, then this would be considered an \textit{overcomplete} autoencoder. This is compared to an \textit{undercomplete} autoencoder, where the number of nodes in the hidden layer is less than the size of the input layer, which in our case is 35. So at a minimum, as noted above, the size of the hidden layer should be $p - 1$. There are much fewer uses of an overcomplete autoencoder compared to the undercomplete version, because of the very nature of the use of these in practice. That is, if our goal is to force information loss, and thereby compress the data in the hidden layer to make the task of the decoder harder to learn better, then it follows that the size of the hidden layer should indeed be smaller to create this information loss. Alternatively, in the overcomplete version of an autoencoder, some nodes may be virtually empty or non-representative, which amounts to cluttering the encoding step. Though this could be a justifiable decision to make the task of the decoder more difficult in a different way, the loss of information is consistently considered the core of the logic of the autoencoder. Thus, I recommend undercomplete autoencoders in most applications, unless justification otherwise is extremely well-motivated. 

Extending common notation (e.g., \citet{goodfellow2016deep}), consider a brief formalization of the generic form for the encoding step,

\begin{equation}
M = f(\boldsymbol{X}),
\label{eq:encode}
\end{equation} 

where $M$ records the codings from the original input, $\boldsymbol{X}$. Then, the decoder's goal is to reconstruct the input, $\boldsymbol{X}$, based on the codings, or the ``representation layer'' from \ref{eq:encode}. We then define the decoder as,

\begin{equation}
\boldsymbol{X^\prime} = f(M),
\end{equation}

where $\boldsymbol{X^\prime}$ is the reconstruction of the inputs, based on the $M$ from \ref{eq:encode}. To compare $\boldsymbol{X}$ and $\boldsymbol{X^\prime}$, recall we must record the error from the reconstruction attempt. Thus, the goal is to minimize the reconstruction error, in an attempt to get as close as possible to the original input space, but on the basis of the codings from the representation layer, 

\begin{equation}
\mathrm{min} \mathcal{L} = f(\boldsymbol{X}, \boldsymbol{X^\prime}), 
\end{equation}

or more compactly, 

\begin{equation}
\mathcal{L}(\hat{\boldsymbol{X}}, \boldsymbol{X}).
\end{equation}

\subsubsection{Applying Autoencoders to the ANES Data}

For this final application using the 2019 ANES data, I will first walk through constructing an autoencoder using the machine learning engine, H20. But, extending beyond past applications in this Element, I will then use the extracted features (``deep features'') in a simple supervised task to demonstrate the value and ease of such an extension. This application, then, is intended to give a realistic, albeit limited, picture of a common use of an autoencoder in a social science task. 

We start with setting a few things up. In the code for Section 6, we start by initializing the H20 session, set our party affiliation feature aside for coloring as usual, but also for our supervised task at the end of the section. Also, to speed up processing time, as these types of neural networks can take a long time, especially as they and the data grow increasingly complex. Finally, I divide the data into training, testing, and validation sets. This is a common approach in machine learning research to train, test and tune learners, but not threaten the generalizability of the learner. For more, see \citet{friedman2001elements}. Now, with the data and H20 session set up, we are ready to build a shallow (non-deep) undercomplete autoencoder. 

The syntax to build an autoencoder using the H20 engine includes several tunable hyperparameters and several default values. Of note is the \texttt{autoencoder} argument, which when set to \texttt{TRUE} allows us to build an autoencoder. Also, we have specified that we want a single hidden layer with 16 nodes, per discussion above on ensuring this is an undercomplete autoencoder. Once trained on 60\% of out data, we are also telling the \texttt{h2o.deeplearning()} function that we want to ``test,'' generate predictions using the held out 20\% of the data. This is the argument that allows for calculation of reconstruction error. Once run, we need to extract the deep features/codings learned from the model. That is, we want to focus less on reconstruction at this stage, and instead extract the codings from the representation/hidden layer. This will help generate interpretable and comparable plots in line with the other techniques and approach throughout the Element. To do so, we rely on the \texttt{h2o.deepfeatures()} function to extract the deep features.

\begin{figure}[h!]
	\centering
	\includegraphics[scale = 0.4]{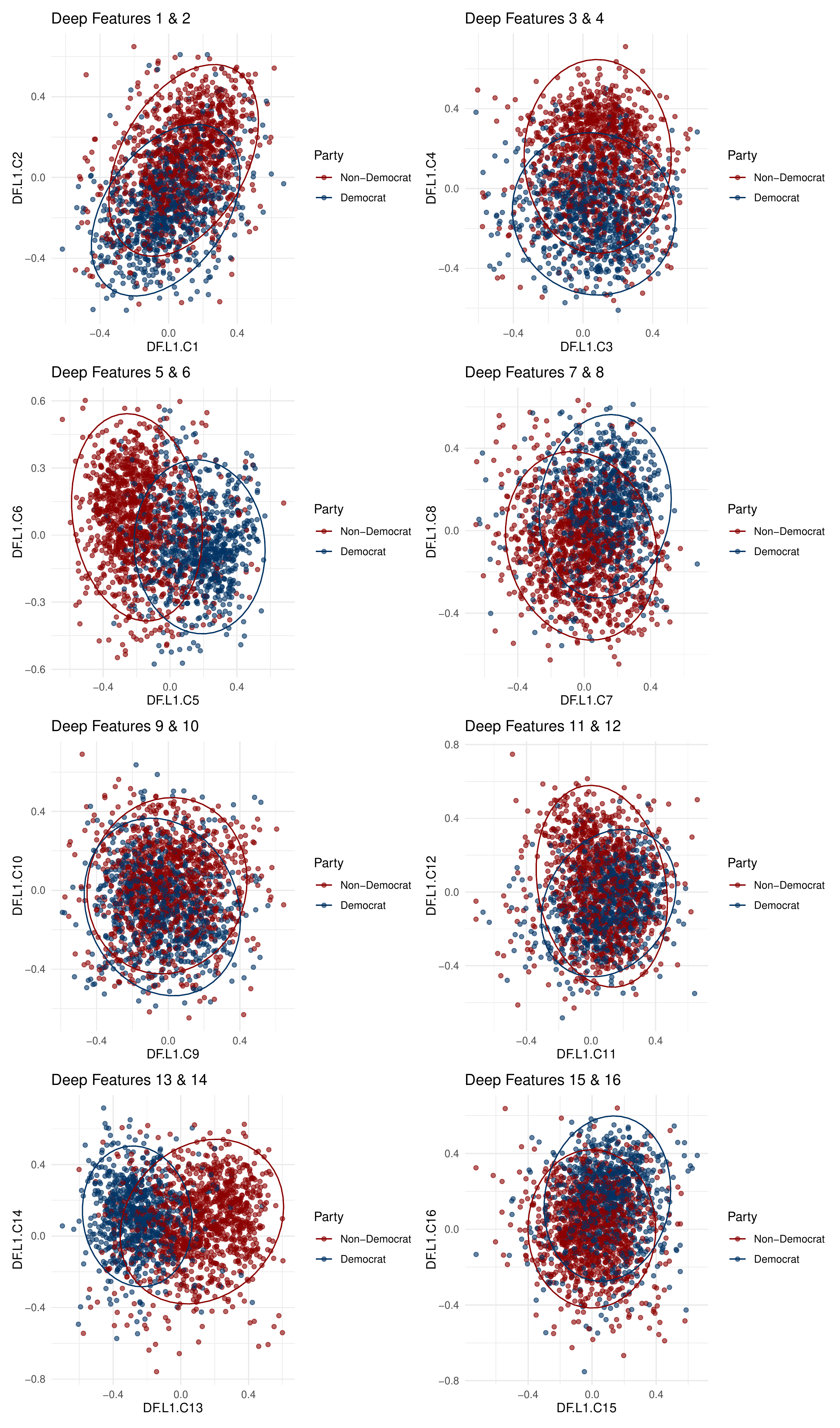}
	\caption{Deep Feature Plots}
	\label{figure:ae1}
\end{figure}

The returned output is a bit odd. The deep feature labels, e.g., DF.L1.C1 are read: ``data frame, layer number, column number.'' For our purposes, which is a simple application, we only a single data frame and a single hidden layer (i.e., \texttt{layer = 1} in the previous function call). Thus, the only value in the titles that will change is the \textit{$C_*$}. As we specified a compressed hidden layer with 16 nodes/neurons, then the output returned 16 column vectors, giving a deep feature matrix of size $1910 \times 16$, which is the size of the training set. For substantive purposes, we can plot these deep features against each other and color by party affiliation as before to get a first look as to whether they are picking up on partisan separation in the projection space. Due to limited space, the code to produce these plots is omitted, but can easily be built by repurposing some of the earlier plot code in the Element. The output is in Figure \ref{figure:ae1}. 

Surprisingly, the patterns suggest less separation than prior plots from other techniques we have seen, though separation across all deep features is clear. Democrats are indeed distinct in their feelings toward a battery of issues and actors compared to Non-Democrats. Importantly, these patterns are only from part (60\%) of the data, and could be the reason for the different patterns. I encourage readers to plot across the validation set to compare, or even change the portions of data splits in the sets. 

But to demonstrate a fuller application, we will turn now to a supervised task. That is, rather than \textit{color} points by party affiliation, we will \textit{predict} party affiliation as a function of the deep features extracted from our autoencoder. To do so, we first extract the features, and then use these as input in a deep neural network. Importantly, the model at present now is \textit{not} unsupervised as we have covered to this point in this Element. But it is supervised, where we have labeled output, \textit{party affiliation}. We are attempting to classify respondents' party affiliations based on the deep features we extracted from our autoencoder fit above. So there are two neural networks at play: the first (autoencoder) giving us our set of input features, and the second (deep feedforward neural network based on the code below, \texttt{hidden = c(8, 8)} denoting two hidden layers, each with 8 neurons) for predicting party affiliation as a function the deep features. A common and simple way to assess the quality of a binary choice model of this sort is to generate a confusion matrix showing the true (false) positive predictions (\textit{Democrat}) relative to the true (false) negative predictions (\textit{Non-Democrat}). These results are shown in Table \ref{table:nn}.

	\begin{table}[h!]
	\begin{center}
		\renewcommand{\arraystretch}{1.1}
		\caption{Confusion Matrix For Deep Neural Network Predicting Party Affiliation}
		\label{table:nn}
		\begin{tabular}{c > {\bfseries}r @{\hspace{0.7em}}c @{\hspace{0.4em}}c @{\hspace{0.7em}}l}
			& \multicolumn{4}{c}{\bfseries Observed} \\
			\multirow{10}{*}{\rotatebox{90}{\parbox{1.1cm}{\bfseries\centering Predicted}}} & & \bfseries D & \bfseries ND &
			\\
			& D$'$ & \MyBox{\centering 210\\83.7\%}{} & \MyBox{\centering 61\\16.7\%}{} &  
			\\[2.4em]
			& ND$'$ & \MyBox{\centering 41\\16.3\%}{} & \MyBox{\centering 304\\83.3\%}{} &  
			\\
		\end{tabular}
	\end{center}
\end{table}

The confusion matrix in Table \ref{table:nn} suggests our network did a very good job at predicting party affiliation as a function the deep features from the autoencoder based on the ``true'' rate reported in the main diagonal of the matrix. This also suggests that the autoencoder found deep features that are picking up on substantive, latent structure given their ability to predict party affiliation at such high rates (i.e., high true positive and true negative rates, relative to false versions of both).

\begin{figure}[h!]
	\centering
	\parbox{7cm}{
		\includegraphics[width=7cm]{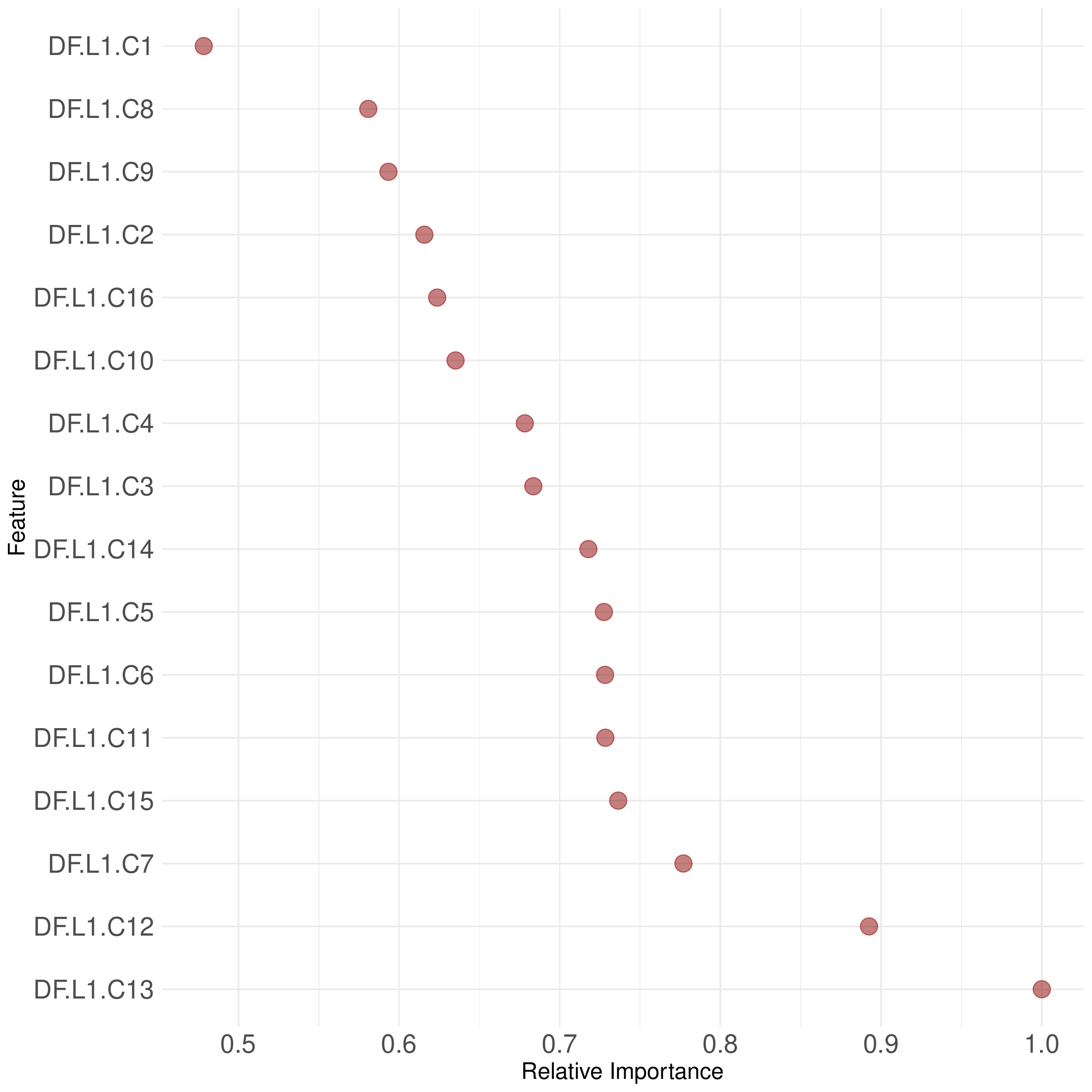}
		\caption{Relative Feature Importance}
		\label{figure:ae2}
	}
	\qquad
	\begin{minipage}{7cm}
		\includegraphics[width=7cm]{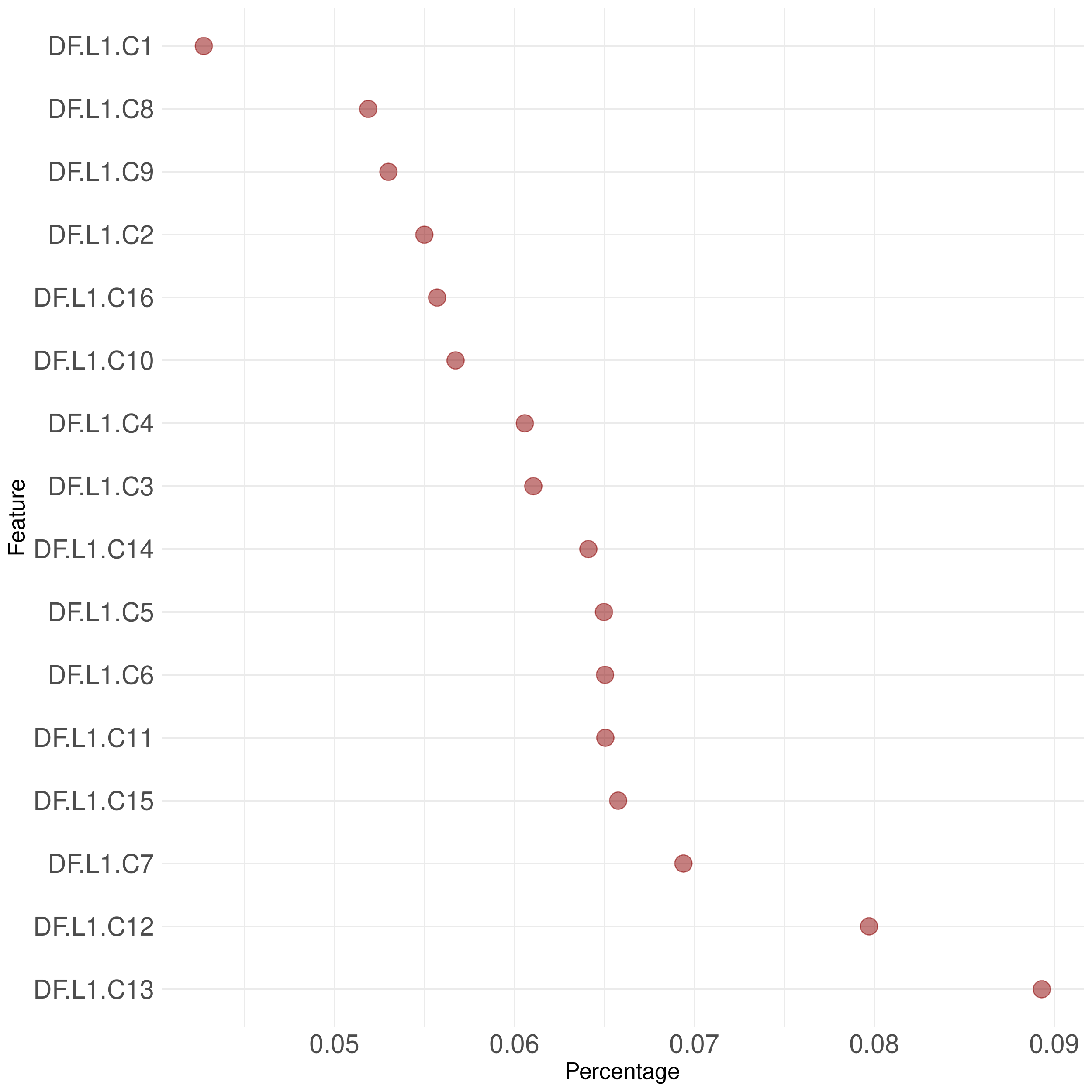}
		\caption{Raw Feature Importance}
		\label{figure:ae3}
	\end{minipage}
\end{figure}

A final approach to explore results from an autoencoder is exploring feature importance. Feature importance is often measure by a feature's contribution to overall fit of the model. For our purposes, we explore feature importance from the deep network that was used for the prediction task. Substantively, this will help by telling use which deep feature from the autoencoder was driving the prediction solution. There are two main ways to think about feature importance: relative importance (summing to 1.0 across all features) and raw percentage (contributed by a given feature). I explore this and present the results for each in Figures \ref{figure:ae2} and \ref{figure:ae3}.

Across both of these plots of feature importance in Figures \ref{figure:ae2} and \ref{figure:ae3}, deep features 12 and 13 are the most important by a wide margin. Note, this pair was not a combination of plotted features in Figure \ref{figure:ae1}. With this, as a final inspection of the results, we will now plot these against each other as they were most influential in the classification task. Thus, we might expect to see the clearest separation between party affiliation across these two deep features. I plot these against each other using both the training and validation sets below in Figure \ref{figure:ae5}, as patterns should be similar across each.

\clearpage

\begin{figure}[h!]
	\centering
	\includegraphics[scale = 0.38]{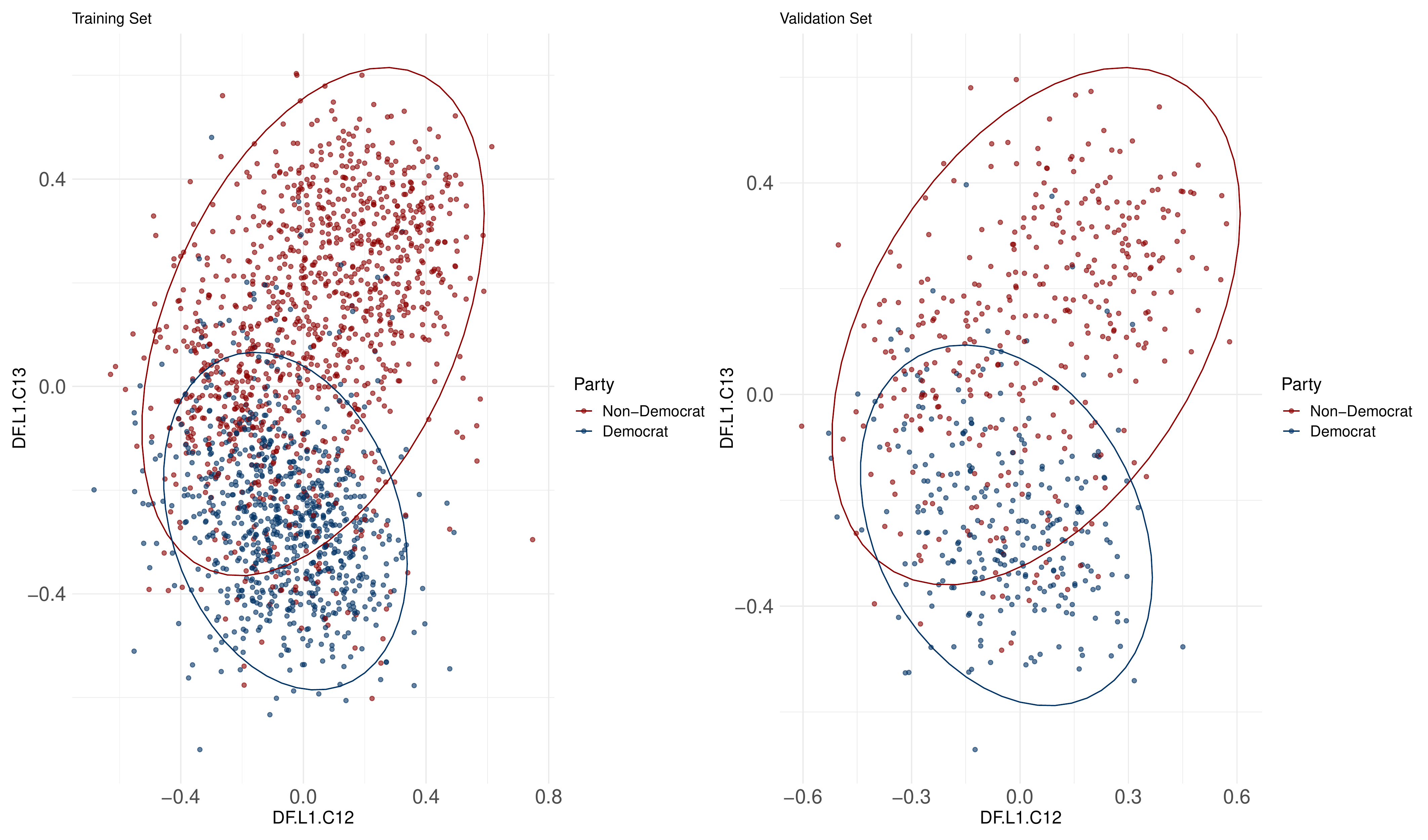}
	\caption{Deep Features 12 and 13 by Party Affiliation}
	\label{figure:ae5}
\end{figure}

Now, the patterns in both training and validation sets in Figure \ref{figure:ae5} show a pattern that looks quite similar to patterns from other techniques throughout the Element, where there is very clear separation between the parties in the projection space, but still some blending toward the middle of both dimensions. This suggests, then, that deep features 12 and 13 in this case can be thought of as the first two ``principal components'' from a PCA fit. Thus, the output of an autoencoder does not automatically sort the deep features based on any metric, whether something akin to PVE in PCA, or feature importance as we have explored here. As such, it is up to the researcher to dive into the results and effectively pull them apart and explain patterns in a substantive light.

Importantly, as with all other techniques covered in this Element, autoencoders are not without limitations. As the task of an autoencoder is to reproduce the original input space based on a simplified or ``encoded'' version of that space, the solution may very easily perform poorly on \textit{new} data. That is, by training and tuning an autoencoder to learn defining aspects of the latent structure to aid in reconstruction of the high dimensional space, it is in effect formalizing a process of overfitting to the training data. The training data in this scenario is the full input space, and the decoded output layer gives a learned version of that space based on the encoded version of the same space. For our present application working with nationally representative survey data, we might bypass this weakness by trusting that the ANES sample gives a realistic look at the American population. Thus, an ``overfit'' autoencoder would not be as much of a worry as the solutions should work relatively well on some other sample drawn from the same population using the same ANES sampling techniques. Yet, from a methodological perspective, we may be interested in a technique that gives a more generalizable solution, especially in cases leveraging worse or different data than a nationally representative American survey. For example, such an approach to guard against this threat of overfitting that is innate to all vanilla autoencoders, is a recent innovation called a \textit{denoising} autoencoder. The task of the denoising autoencoder is to first add statistical noise to the original input layer. This initial step has the effect of making the job of the decoder much harder, as it must attempt reconstruction based on a noisy version of the already-reduced input space. And the increased difficulty of decoding in this scenario should, in theory, allow for a more generalizable solution \citep[~504-505]{goodfellow2016deep}. The point here is that though sophisticated and often accurate, even complex methods like autoencoders have weaknesses and limitations that must be considered when used in a research project. At a minimum these limitations should be mentioned in the reporting of the results. 

\subsection{Suggestions for Further Reading} 

As with the previous section, the value of these methods is clearest when looking at implementation in applied settings. For example, \citet{kourtit2012smart} use self-organizing maps to explore smart cities, and \citet{li2018generating} derive a clever application of an autoencoder to generate poetry.

\clearpage

\section{Final Thoughts on Dimension Reduction}

It has been my goal in this Element to present a modern picture of dimension reduction, with application in the social sciences. The \textit{modern} part of this goal focusing on recent methodological development, though, is only possible in light of the simple, but powerful original approach to dimension reduction: principal components analysis (PCA). 

Building on PCA and the shared goal of learning from data to produce a simpler projection, LLE is based on a similar linear construction, but is much more flexible to model nonlinearities in data. Yet, t-SNE and UMAP offer very different approaches to capturing global structure and projecting it locally. The extensions of t-SNE and UMAP allow for even more flexibility in model complex structures beyond what either LLE or PCA could handle, and they do so in a fundamentally different (graph-based) way. Further still, another approach to dimension reduction is through a neural network architecture focusing on both grouping patterns in data and also minimizing reconstruction error. Both SOM and autoencoders, though differing in construction, take advantage of an iterative process for filtering data to learn and project structure. Through coverage of these techniques, this Element has given a framework for thinking about and applying dimension reduction to address a host of problems all stemming from data complexity and size. This framework guides method selection, which might be dependent on the goals of the project (e.g., feature extraction versus representation) or conceptualizations of how best to think about similarity, structure, and complexity (e.g., maximizing variance versus manifold reconstruction).

All techniques covered in this Element have been extended over the years in numerous ways, with deepened complexity in their own rite. That is, we can always go deeper and get more complex, dependent on the nature and scope of the problem at hand. For example, there are many versions of autoencoders beyond even the deep version we briefly touched on in the previous section. There are variational, sparse, and denoising autoencoders, to name a few. Each of these and the many other extensions were derived to address unique problems in extremely clever ways. As noted in the previous section, the denoising autoencoder is interested in making the decoding task harder to (hopefully) give a more reliable solution. It does so by adding some extra ``noise'' to the original input layer. The task of the decoder, after the encoding stage, is not only to now reconstruct the encoded input space, but also to separate the noise from the signal, such that the added noise should be absent from the decoded solution. Such extensions are worth mentioning to not only demonstrate the rapidly developing nature of the field, but to also encourage readers to think critically about the best technique for the problem at hand. And importantly, if a technique does not exist, then readers should feel empowered to follow similar past approaches and build on the firm foundations, such as the autoencoder, to develop the needed extension to solve some problem. To underscore this point, it is important to note that the autoencoder is itself an extension of an existing method: the restricted Boltzmann machine (RBM). The RBM, which was an extension of the Boltzmann machine, was essentially attempting to leverage a neural network framework to develop a version of PCA, where the goal was the output from the encoding stage. The RBM algorithm stops short of decoding the encoded layer, and instead outputs the reduced set of features, similar to the result we get from a PCA solution. Thus, by adding the inverse of the encoding stage to the backend of the RBM, the autoencoder was developed.

Perhaps an even broader, but still related conclusion is that we have only scratched the surface regarding coverage of techniques. Though much more ground could be covered in building a dimension reduction toolbox, instead this Element has introduced a framework for \textit{thinking about dimension reduction} in a holistic light. To this end, we have covered dimension reduction on the basis of calculating and extracting new features on the basis of learned \textit{trends} in the data (maximizing variance via PCA). We have also covered dimension reduction for the purpose of learning and representing the full latent \textit{structure} of the input space instead of just a summary in a linear way (via LLE) and in a nonlinear way, for both feature extraction and visualization (via t-SNE and UMAP). Further, we have covered dimension reduction in a way that emulates the process of neurons firing and learning within neighborhoods on the basis of raw stimuli (via SOM). And finally, we covered dimension reduction for learning the structure of a space by creating and solving a problem based on an imperfect version of the original space (via autoencoders). Though the general goal remained constant, the selection and implementation of the technique resulted in unique ways to conceptualize and solve a common problem. And through the power of visualization, we demonstrated throughout that the patterns uncovered, regardless of variance in method-specific mechanics, were largely consistent suggesting the latent structure was indeed real, rather than a function of noise or randomness in the data. If the latter were the case, then different approaches to solving the dimension reduction problem would have given different versions of the same data space. 

Related, and perhaps most importantly, coverage and presentation of the techniques in this Element was based in the recognition that method selection \textit{always} flows from a theoretical assumption of the world, if even that assumption is not formalized. That is, our conceptualization of both the input data space and also the ideal way to make that input space more understandable, is manifested in the selected method. As with supervised modeling tasks, in unsupervised dimension reduction the mere selection of a technique is itself a theoretical assertion at some level. For example, deciding whether to create a lower dimensional version of the \textit{full} space is based in the assumption that the contours of the space (e.g., UMAP) reveal a more useful version of the data than a global summary of part of the space (e.g., PCA). Or alternatively, autoencoders do not rely on the local linearity assumption of the manifold as LLE does, so the modeling process is not bound by protecting against a violation of linearity. The result is may be better performance by the algorithm, but the sacrifice may be in interpretability. LLE is much simpler to interpret than a deep autoencoder. As a result, there is an implicit tradeoff in model complexity versus interpretability in unsupervised learning as there is in supervised learning. There should, at a minimum, be an appreciation of this reality along with the fact that our unique biases influence every model we build and every choice we make in the research process. This is not a normatively ``bad'' or ``good'' thing; it is merely a reality. So, by selecting a model, we are in a sense codifying our assumptions in tandem with making a judgement on the complexity/interpretability tradeoff. This codification through model selection should be appreciated, well-motivated, and justified throughout the modeling process, whether supervised or unsupervised. As such, I urge readers to take great care in justifying selection of a technique, and follow the \textit{Risk Averse Workflow} addressed at the conclusion of the LLE section.

Along with presentation of this material in a ``modern'' context means that much work is currently being done to push the field of dimension reduction forward through development of new techniques and offering new ways to think about addressing new problems. Of note is the issue of \textit{big data}. How best to process and treat big data in the age-old service of dimension reduction is a particularly vexing issue \citep{zhang2018dimension}. Some work has been done on this front. For example, \citet{krishnan2018multi} develop a multistep dimension reduction approach to effectively deal with data in batches. Related, dimension reduction is not beholden to numeric data. A host of techniques, some entirely different, are used to reduce the dimensionality and complexity of \textit{text} data. For example, vector space models like word2vec can handle massive data of sizes in the billions \citep{mikolov2013efficient}. The idea behind word2vec and other vector space embedding models is to reduce dimensionality by learning and reducing complexity based on semantic similarity across a text corpus. 

In sum, regardless of the type of data, the size of the data, or the techniques used, dimension reduction offers researchers a powerful set of tools for making complex data spaces more manageable, interpretable, and simpler. This Element has shown a few of some of the most widely used, modern approaches for doing so, along with tips on implementation of the methods in R. Taken with my earlier Element on clustering \citep{waggoner2020unsupervised}, it is my hope that the value of unsupervised machine learning for uncovering non-random structure and so learning from data is clear for a variety of applications that may have been previously unclear. As a result, readers are encouraged to dive deeper into this exciting, rapidly developing world and perhaps even develop algorithms of their own.

\clearpage

\bibliography{dr}

\end{document}